\documentclass[sigconf]{acmart}
\acmSubmissionID{126}

\usepackage{booktabs} % For formal tables

\usepackage[ruled]{algorithm2e} % For algorithms
\usepackage[capitalize,noabbrev]{cleveref}
\usepackage[table]{xcolor}
\usepackage{graphicx}
\usepackage{makecell}
\usepackage{subcaption}
\usepackage{selectp}

\SetAlFnt{\small}
\SetAlCapFnt{\small}
\SetAlCapNameFnt{\small}
\SetAlCapHSkip{0pt}

\newcommand{\tracker}{MimicManipulator}
\newcommand{\generator}{MaskedManipulator}

\copyrightyear{2025}
\acmYear{2025}
\setcopyright{cc}
\setcctype{by-nc}
\acmConference[SA Conference Papers '25]{SIGGRAPH Asia 2025 Conference Papers}{December 15--18, 2025}{Hong Kong, Hong Kong}
\acmBooktitle{SIGGRAPH Asia 2025 Conference Papers (SA Conference Papers '25), December 15--18, 2025, Hong Kong, Hong Kong}\acmDOI{10.1145/3757377.3763934}
\acmISBN{979-8-4007-2137-3/2025/12}

\newcommand\edited[1]{{#1}}

\begin{document}
\fancyfoot{}
\pagestyle{plain} 
\title{\generator: Versatile Whole-Body Manipulation}

% DO NOT ENTER AUTHOR INFORMATION FOR ANONYMOUS TECHNICAL PAPER SUBMISSIONS TO SIGGRAPH 2019!
\author{Chen Tessler}
\affiliation{%
 \institution{NVIDIA}
 \country{Israel}}
\author{Yifeng Jiang}
\affiliation{%
 \institution{NVIDIA}
 \country{USA}}
\author{Erwin Coumans}
\affiliation{%
 \institution{NVIDIA}
 \country{USA}}
\author{Zhengyi Luo}
\affiliation{%
 \institution{NVIDIA}
 \country{USA}}
\author{Gal Chechik}
\affiliation{%
 \institution{NVIDIA}
 \country{Israel}}
\author{Xue Bin Peng}
\affiliation{%
 \institution{NVIDIA}
 \country{Canada}}
 \affiliation{%
 \institution{Simon Fraser University}
 \country{Canada}
 }

\renewcommand\shortauthors{Tessler, C. et al}

\begin{abstract}

    We tackle the challenges of synthesizing versatile, physically simulated human motions for full-body object manipulation. Unlike prior methods that are focused on detailed motion tracking, trajectory following, or teleoperation, our framework enables users to specify versatile high-level objectives such as target object poses or body poses. To achieve this, we introduce \generator, a generative control policy distilled from a tracking controller trained on large-scale human motion capture data. This two-stage learning process allows the system to perform complex interaction behaviors, while providing intuitive user control over both character and object motions. \generator\ produces goal-directed manipulation behaviors that expand the scope of interactive animation systems beyond task-specific solutions.
\end{abstract}




% \keywords{Whole body manipulation, physics based simulation, versatile control}

\begin{teaserfigure}
  \includegraphics[width=0.95\linewidth]{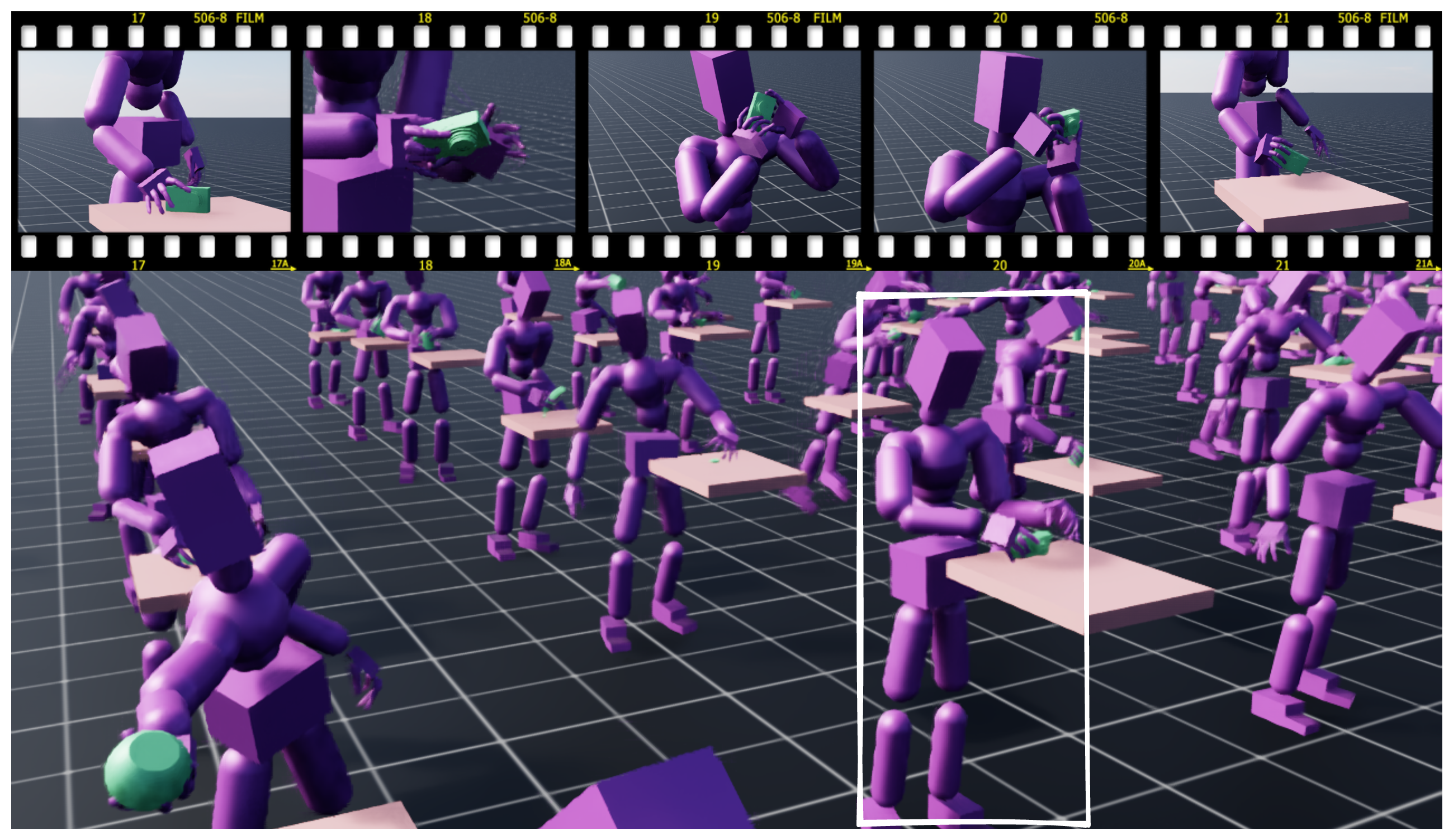}
  \caption{\generator\ enables physics-based humanoids to perform intricate, object interactions from sparse spatio-temporal goals.}
  \label{fig:teaser}
\end{teaserfigure}

\maketitle

\section{Introduction}

Generating versatile humanoid agents capable of autonomous and controllable behaviors is a shared challenge for both character animation and robotics. While recent approaches have significantly advanced locomotion capabilities for simulated \cite{tessler2024maskedmimic,wu2025uniphys} and real-world humanoids \cite{ji2024exbody2,allshire2025visual,ze2025twist}, synthesizing controllers that can skillfully interact with the environment – especially with small, dynamic objects requiring precise dexterous manipulation – remains a daunting hurdle. A key challenge lies in how to effectively translate high-level human intent into low-level motor commands. For example, walking to an object, picking it up, inspecting the object, and then placing it back down on a table.

We aim to build a versatile controller that seamlessly combines full-body locomotion with dexterous manipulation, which can also interpret such diverse command inputs. However, there exits an inherent tension: the need for broad task flexibility versus the need for exacting physical precision. On one hand, achieving versatility requires controllers capable of understanding abstract user intents, like sparsely defined target coordinates (e.g., ``reach this point'') or high-level object goals (e.g., ``place cup here''). These often underspecified goals present vast solution spaces, posing substantial challenges for control and learning techniques like reinforcement learning (RL). On the other hand, successful manipulation, particularly with common objects, demands meticulous precision. Even small deviations during physical interaction – grasping, repositioning, or placing – can lead to instability, object dropping, or complete task failure, often without recovery \cite{luo2024precise}. While prior work has tackled aspects of this challenge, existing methods typically excel at either sophisticated manipulation \cite{lin2024twisting,iyer2024open} or versatile body control (e.g., via goal-conditioning \cite{tessler2024maskedmimic}). Achieving unified versatile control that encompasses both body and object goals, and gracefully balances this flexibility-precision trade-off, remains an open problem.

We propose a unified control framework for diverse and precise full-body-manipulation tasks, responding to inputs ranging from detailed kinematic targets (e.g., tracking the hand positions via teleoperation) to sparse, high-level goals (like a desired position of an object). Our framework extends spatio-temporal goal-conditioning \edited{(MaskedMimic \cite{tessler2024maskedmimic})} to encompass both the humanoid's body parts and the manipulated object. To master the precise execution required by human-object interactions, our first stage (\tracker) leverages human motion capture (e.g., GRAB dataset \cite{taheri2020grab}). The motion capture data demonstrates successful interaction strategies, such as sequences of grasping, placing, and regrasping objects, and hand-to-hand transfers. This physics-based tracker (\tracker), trained with RL in a fully observable setting, learns the necessary actions to accurately reconstruct these intricate behaviors. Then, to enable versatile control from sparse goals, \tracker's expert interaction knowledge is distilled into our \generator\ policy. By learning from \tracker, \generator\ becomes capable of producing human-like and diverse behaviors in response to under-specified goals for both body parts and objects (e.g., ``bring the object to coordinate $(x,y,z)$''), effectively addressing the multi-solution challenge of versatile control.

The central contributions of this work are:
\begin{enumerate}
    \item \edited{We extend the versatile full-body controller, MaskedMimic \cite{tessler2024maskedmimic}, enabling full-body-manipulation. We demonstrate how to successfully leverage human demonstrations to produce diverse, physically plausible, and human-like solutions to underspecified tasks involving coupled human-object interactions.}
    \item \tracker: A physics-based motion tracker that accurately infers actions and reconstructs dexterous human full-body-manipulation sequences from reference motion data.
    \item \generator: A unified generative full-body-manipulation policy that enables diverse spatio-temporal goal-conditioning for both humanoid body parts and manipulated objects.
\end{enumerate}

\section{Related Work}

Our work builds upon advances in demonstration-driven control, object manipulation, and versatile goal-conditioned policies for physics-based characters.

\paragraph{Demonstration-driven control.} Traditional reinforcement learning (RL) for character control often requires meticulous reward engineering. Imitation learning (IL) offers a powerful alternative by leveraging expert demonstrations. DeepMimic \cite{peng2018deepmimic} showed RL can robustly replicate reference motions, a paradigm extended to object manipulation by works like \citep{yu2025skillmimic} for dexterous sequences and InterMimic \cite{xu2025intermimic} for retargeting interactions across morphologies. Our \tracker\ stage is inspired by these approaches, using RL to train a single policy to precisely track the entire diverse human full-body-manipulation demonstrations from GRAB \cite{taheri2020grab}, thereby recovering actions and ensuring physical plausibility.

\paragraph{Object manipulation and grasping} Integrating object manipulation into humanoid control substantially increases task complexity. Early approaches achieved progress without reference data by relying on hand-crafted rewards and large-scale reinforcement learning \cite{akkaya2019solving, lin2024twisting}. More recent work has explored demonstration-driven methods: for example, \edited{OmniGrasp \cite{luo2024omnigrasp} synthesizes full-body motions conditioned on dense object trajectories and trained via dense rewards. It achieves this by leveraging a scene-agnostic generative prior that is steered through hierarchical reinforcement learning. Compared to our approach, this design presents two key limitations: (1) the reliance on dense reward signals restricts OmniGrasp to short-horizon training objectives, making it unsuitable for more general goal-conditioned tasks such as bringing an object to a target position \cite{pignatelli2023survey}; (2) the scene-agnostic prior hinders dexterous control. As a result, OmniGrasp struggles reproducing dexterous behaviors such as grasping a teapot by its handle or performing hand-to-hand transfers. In contrast, our method first learns to reproduce dexterous human-object demonstrations from dense goals and then distills this knowledge into a flexible, long-horizon, goal-conditioned controller. The resulting controller, \generator, supports a range of object and/or humanoid body-part goals, enabling the synthesis of diverse full-body-manipulation behaviors from sparse constraints and supporting more adaptable, long temporal sequence control.}

\paragraph{Versatile and goal-conditioned control} A trend in robotics and animation involves hierarchical control, where high-level systems specify abstract goals (e.g., walk targets \cite{peng2022ase}, grasp poses \cite{li2024virt}, 2D paths \cite{li2025hamster}) for low-level execution, akin to System 1/System 2 reasoning \cite{kahneman2011thinking}. MaskedMimic \cite{tessler2024maskedmimic} significantly advanced System1 versatility by framing character control as motion inpainting, allowing a unified model to synthesize diverse motions from sparse future joint targets. Our work extends this by enabling explicit spatio-temporal goal-conditioning on both humanoid body parts and the manipulated object within \generator. This provides direct, versatile control over the entire coupled human-object system, crucial for sophisticated full-body-manipulation tasks.

\section{Preliminaries}

We formulate our problem within the framework of goal-conditioned Markov Decision Process (MDP) \cite{puterman2014markov}, defined by a tuple $M = (S, G, A, P, R, \gamma)$, consisting of states $S$, goals $G$, actions $A$, transition dynamics $P$, a reward function $R$, and a discount factor $\gamma$. The objective is to learn a policy $\pi(a_t | s_t, g_t)$ that selects an action $a_t$ based on the current state $s_t$ and a task-specific goal $g_t$.

The state of the humanoid's $J$ links at time $t$ is described by their individual 3D positions $p_{j,t} \in \mathbb{R}^3$ orientations $\theta_{j,t}$, linear $v_{j,t}$ and angular $\omega_{j,t}$ velocities. Each orientation $\theta_{j,t}$ is represented using a continuous 6D rotation vector \cite{zhou2019continuity}. All link positions and orientations constitute the full body kinematic state, denoted $q_t = (\{p_{j,t}\}_{j=1}^J, \{\theta_{j,t}\}_{j=1}^J)$. The corresponding linear and angular velocities collectively form $\dot{q}_t$. When interacting with the scene, a body part may be subjected to contact forces, which we denote by $c_{j,t}$. Object states $q^{\text{obj}}_t$ are similarly defined by 3D position $p^{\text{obj}}_t$, 6D orientation $\theta^{\text{obj}}_t$, velocities $\dot{q}^{\text{obj}}_t$, and contact forces $c^{\text{obj}}_t$.

Reference kinematic and contact quantities, sourced from Motion Capture (MoCap) data or a target trajectory generator, are denoted with a hat accent (e.g., $\hat{q}_t, \hat{\dot{q}}_t, \hat{q}^{\text{obj}}_t, \hat{c}_t, \hat{c}^{\text{obj}}_t$). In contrast to the simulated information, the reference $\hat{c}_t$ are contact indicators. Quantities without this accent refer to states within our physics simulation.

\section{Method}

Our approach for generating versatile humanoid full-body-manipulation capabilities is structured around a two-stage learning process. Human motion capture data provides rich kinematic descriptions of complex interactions but typically lacks the underlying low-level motor commands (actions) required to drive a physics-based character. Our first stage, \tracker, addresses these challenges by training a high-fidelity motion tracker within a physics-based simulation. Operating in a full-information setting with respect to the reference kinematics, \tracker\ learns not only to recover the necessary actions but also to physically reconstruct the target full-body-manipulation sequences, thereby ensuring dynamic feasibility and emphasizing the nuances of object handling. Subsequently, the robust interaction skills learned by \tracker\ is distilled into a versatile control policy, \generator. This second stage yields a policy that can generate novel, physically plausible behaviors by dynamically conditioning on spatio-temporal goals specified for various parts of the humanoid body or the manipulated object itself. The following sections detail the objectives, architecture, and training procedures for each of the two stages.

\subsection{\tracker}

The first stage focuses on reproducing the intricate behaviors required for dexterous full-body-manipulation. We train a single policy $\pi_{\text{track}}$, a motion-tracker denoted by \tracker, to accurately reproduce the complex full-body-manipulation sequences from the processed GRAB dataset.
The reference kinematic data provides a rich description of the desired high-level behavior but does not contain the low-level motor commands (actions $a_t$) necessary to execute these motions in a physically simulated environment. Therefore, we formulate this task as a reinforcement learning problem where $\pi_{\text{track}}$ learns to infer these actions by imitating the reference motion. 

\subsubsection{Task design}

\begin{figure}[t]
     \centering
     \begin{subfigure}[b]{0.33\linewidth}
        \centering
        \includegraphics[trim={6cm 4cm 8cm 4cm},clip,width=\textwidth]{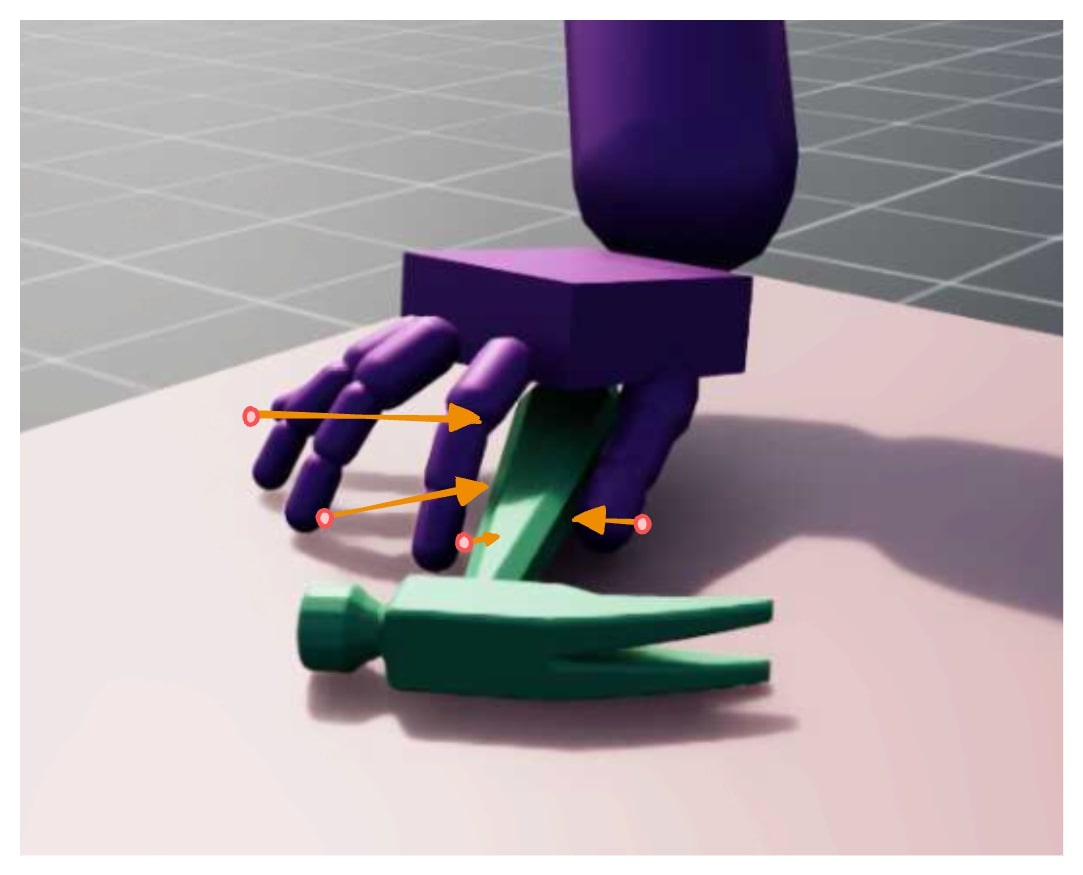}
        \caption{{Pre-contact}}
         \label{fig: rew pre contact}
    \end{subfigure}\hfill
     \begin{subfigure}[b]{0.33\linewidth}
        \centering
        \includegraphics[trim={6cm 4cm 8cm 4cm},clip,width=\textwidth]{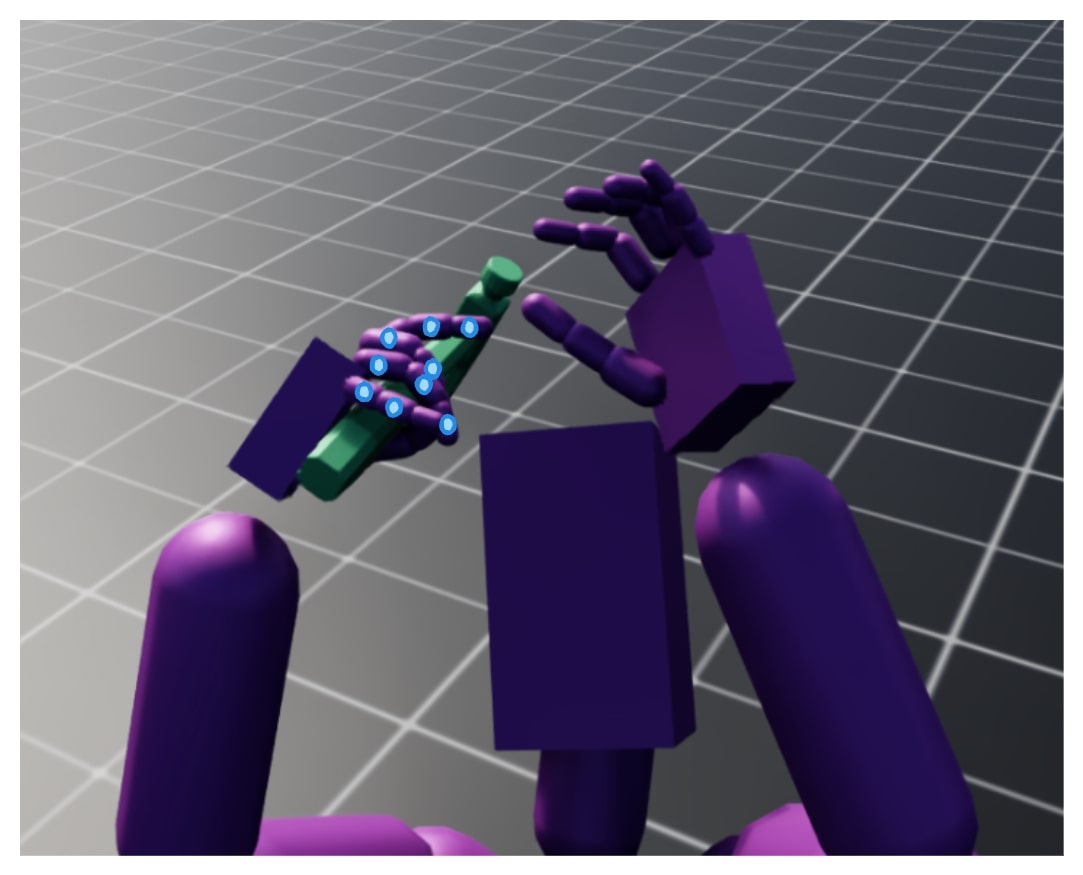}
        \caption{{During contact}}
         \label{fig: rew during contact}
    \end{subfigure}\hfill
    \begin{subfigure}[b]{0.33\linewidth}
        \centering
        \includegraphics[trim={6cm 4cm 8cm 4cm},clip,width=\textwidth]{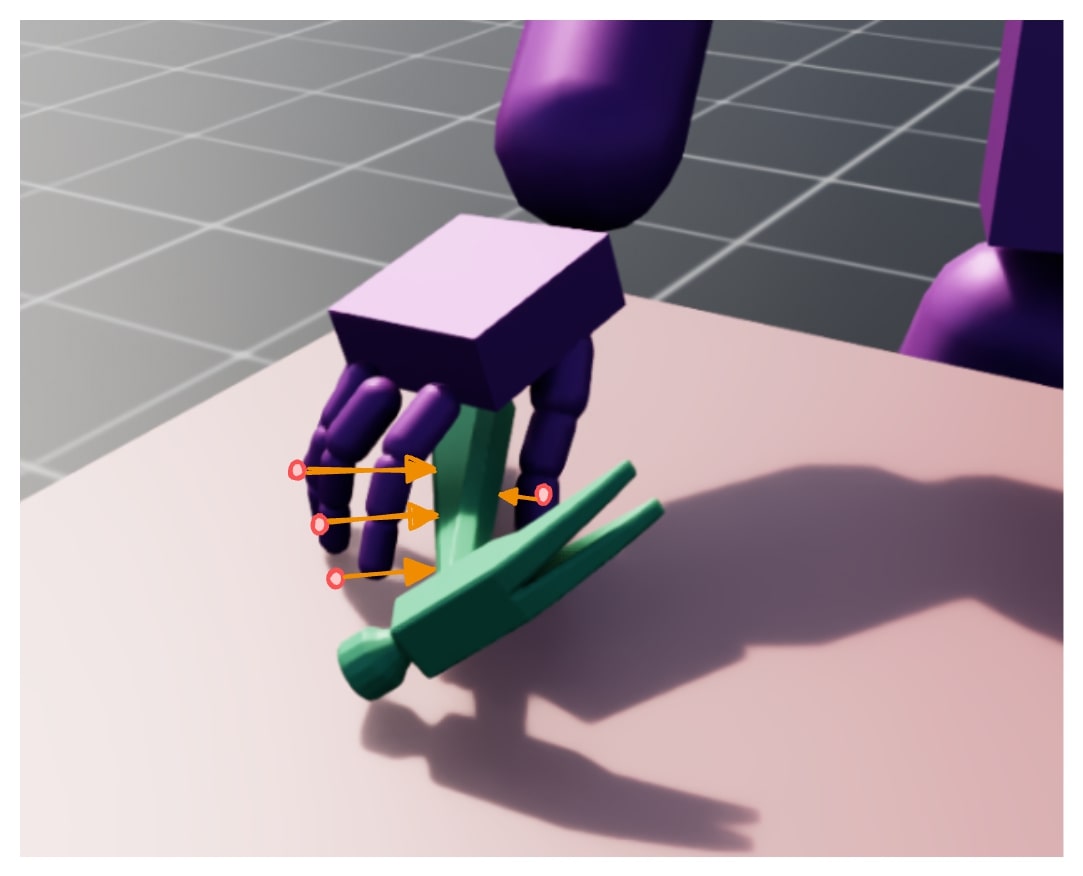}
        \caption{{Post-contact}}
         \label{fig: rew post contact}
    \end{subfigure}
\caption{\textbf{Phased contact reward for precise manipulation.} We design a three-stage contact reward: (a) Approach: Tracks the reference motion while aligning the hand's path relative to the object surface. (b) Engagement: Ensures critical contacts are maintained according to the reference. (c) Release: Promotes a smooth and timely object disengagement mirroring the demonstration. The spheres illustrate the reference joints and the arrows illustrate vectors to the closest point on the object's surface.}
\label{fig: contact reward} \vspace{-0.2cm}
\end{figure}

\paragraph{Observation.} To enable object interactions, the observation $s_t$ comprises of the character's proprioception $q_t$, object information $q^\text{obj}_t$ and character-object relational features. All components are expressed in a character's local coordinate frame and aligned with it's current root heading (yaw) for policy invariance. The character-object relational features capture the spatial relationship between the humanoid and the object, by providing the vector (direction and distance) from each body part to the object surface. \textbf{Object structure:} Precise interaction requires awareness of the object geometry. To accomodate this, we also provide a Basis-Point-Set \cite[BPS]{prokudin2019efficient} representation, which represents all objects by the surface distances to a pre-sampled set of points. Compared to point-cloud representations, the pre-sampled and fixed BPS allow processing using simpler architectures (e.g., MLP), allowing faster training and inference, while still being expressive. \textbf{Future reference trajectory (goal)}: To reproduce the target motions, \tracker\ observes the the goal future poses in the next frame $K=1$ and $K=30$ frames (1 second), denoted by $g_t^{\text{track}}$. Each target future pose observation $\{ \hat{q}_{t+k} \Theta q_t \} = \{ \hat{p}_{t+k} - p_t , \hat{q}_{t+k} \Theta q_t \}$ is a future pose $\hat{q}_{t+k}$ normalized with respect to the current simulated pose $q_{t}$, where $\Theta$ represents the inverse quaternion. In addition, we provide $\hat{c}_{t+k}$ to indicate which body parts should have contact with the object.

\paragraph{Reward and termination.} Successfully reconstructing long and intricate human-object interaction sequences requires high precision. Minor inaccuracies during critical interaction phases (e.g., grasping, placing) can compound, leading to failures much later in the motion. This delayed consequence creates a challenging credit assignment problem for reinforcement learning.

Reference demonstrations implicitly encode successful, precise interaction strategies. For example, where to grab the object in order to naturally reach a target orientation, or to successfully perform a hand-to-hand transfer. We leverage this data and design reward $R_{\text{track}}$ and termination conditions to synergistically guide the \tracker\ agent towards accurate and physically plausible reconstructions. 

The reward $R_{\text{track}}$ is a product of several components. \edited{We found that the multiplicative design helps learning precise, full-body-manipulation over long temporal sequences by strictly requiring competence in all elements of the complex motion} \cite{park2019learning,won2020scalable,xu2025intermimic}.
\begin{equation}\label{eqn: reward}
    R_{\text{track}} = r^{\text{pose}} \cdot r^{\text{contact}} \cdot r^{\text{energy}} \cdot r^{\text{interaction}} \,.
\end{equation}
Each component $r^{(\cdot)}$ is an exponential of a negatively weighted error or cost (e.g., $r = \exp(-w \cdot \text{cost})$), valued between (0, 1].
\edited{
\begin{equation}
    r^{\text{humanoid}} = r^{\text{ht}} \cdot r^{\text{hr}} \,\, \text{and} \,\, r^{\text{obj}} = r^{\text{ot}} \cdot r^{\text{or}}   
\end{equation}
}
components consider the translation and rotation errors for the human and object respectively. The energy component attempts to mitigate artifacts such as jitters. The interaction component aims to prevent human-object penetrations by encouraging a gentle approach and interaction.

\edited{The contact component, illustrated in \cref{fig: contact reward}, is formulated to encourage the policy to reproduce the contact patterns and timing observed in the reference motion. The reward is structured into three distinct phases, each corresponding to a specific stage of the interaction, thereby guiding the agent to accurately replicate the demonstrated behavior. To prevent abrupt behaviors, the approach and engagement phase starts 1 second before contact and ends 0.2 seconds after the reference motion indicates contact has started. Similarly, the disengagement phase beins 1 second before the reference motion indicates that contact has ended. We provide additional details in the supplementary material.}

\edited{In the \textbf{approach and engagement} phase, we identify future contact bodies from the reference trajectory. For each future contact body, the closest point on the object mesh is determined in both the simulation and the reference motion. The reward then penalizes discrepancies in distances and surface normals between each simulated and reference contact body, thereby promoting similarity in the approach trajectory and orientation. This mechanism ensures that end-effectors (e.g., hands) are guided smoothly toward the target contact regions in a manner consistent with the demonstration.}

\edited{During the \textbf{contact maintenance} phase, we identify which reference body parts $\hat{c}_{j,t}$ are marked as should-be-in-contact. For each such body part, a reward is computed based on its proximity to the object mesh. This formulation encourages stable contact, discourages premature detachment, and ensures alignment of the contact configuration with that of the reference motion.}

\edited{Finally, in the \textbf{disengagement phase}, previously contacting body parts are incentivized to move away from the object in a controlled and timely manner. Rewards are again based on distance and surface normal alignment, encouraging a smooth release that mirrors the reference disengagement dynamics.}

Complementing the reward design, we leverage early termination to define a feasibility envelope for the training phase \cite{peng2018deepmimic,luo2023perpetual,tessler2024maskedmimic}. An episode is terminated if any body part deviates by >25cm from reference, or the object deviates >10cm from reference, or when there's a prolonged unintended loss of contact for >10 consecutive frames during a contact phase, or if the character maintains contact >0.4sec after the reference has released.

\paragraph{Prioritized training for complex interactions.} We train a single policy to reconstruct the entire dataset. The reference sequences span a wide range of difficulty. This variation arises from both the object's characteristics (e.g., grasping a mug by its handle is more demanding than grasping a banana) and the intricacy of the interaction itself (e.g., bimanual operations or tasks requiring precise tool-like dexterity). During training we periodically evaluate the agent's performance and over-sample sequences in which it has failed \cite{luo2024omnigrasp,tessler2024maskedmimic,luo2023perpetual}.

\begin{figure}[t]
    \centering
\includegraphics[trim={0cm 4cm 3cm 4cm},clip,width=1\columnwidth]{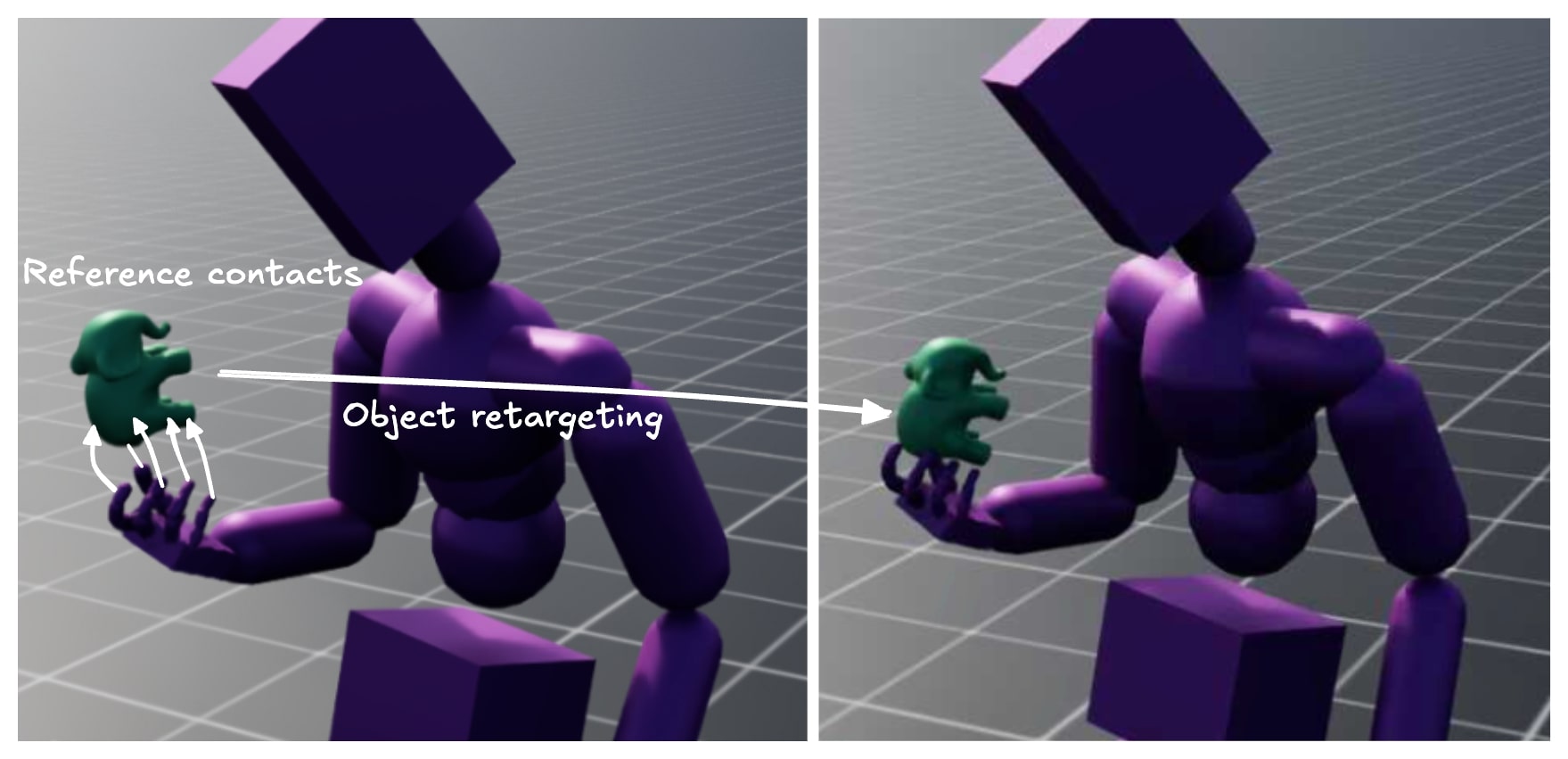}
\caption{\textbf{Object Retargeting for Morphological Differences.} Transferring motion between characters of varying shapes can misalign human-object interactions (left). Our method leverages original contact data to retarget the object's trajectory, preserving interaction consistency (right).}
\label{fig: retargeting} \vspace{-0.2cm}
\end{figure}

\subsubsection{Data processing}

\begin{figure*}
    \centering
    \begin{subfigure}[b]{0.27\textwidth}
        \centering
        \includegraphics[width=\linewidth]{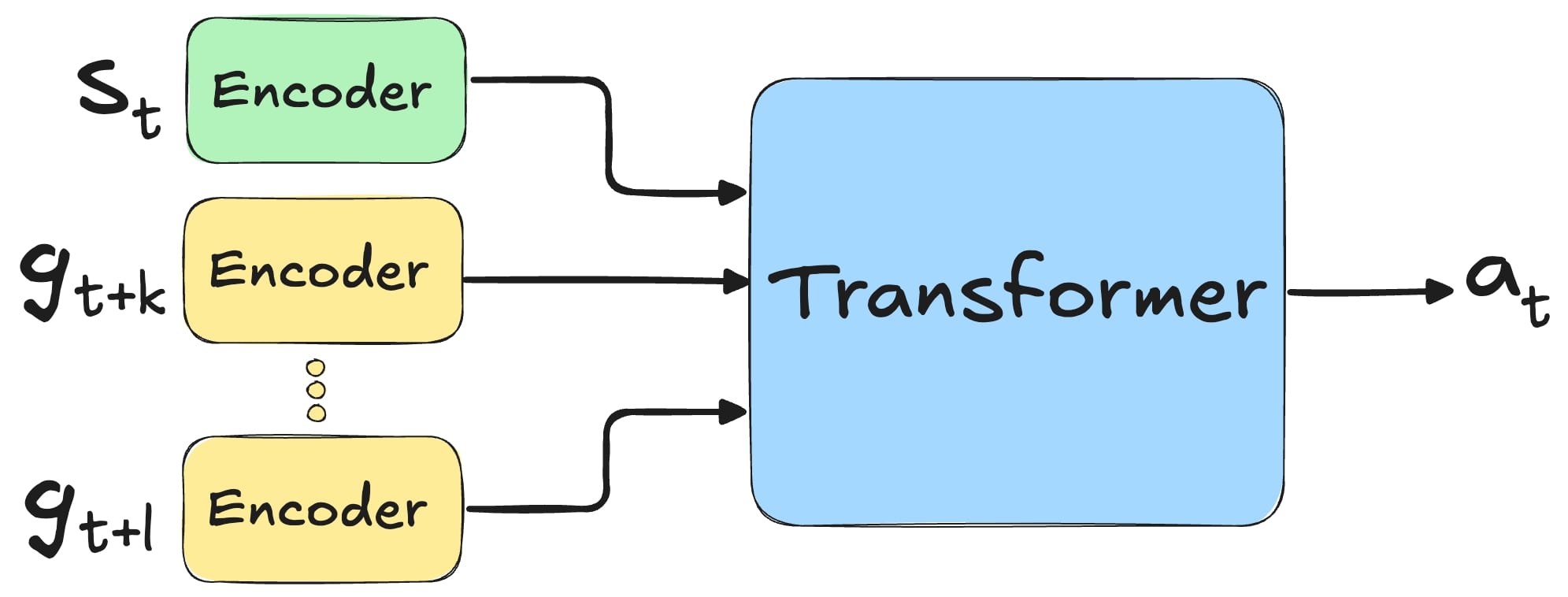}
        \caption{Deterministic}
         \label{fig: deterministic}
    \end{subfigure}\hfill
    \begin{subfigure}[b]{0.39\textwidth}
        \centering
        \includegraphics[width=\linewidth]{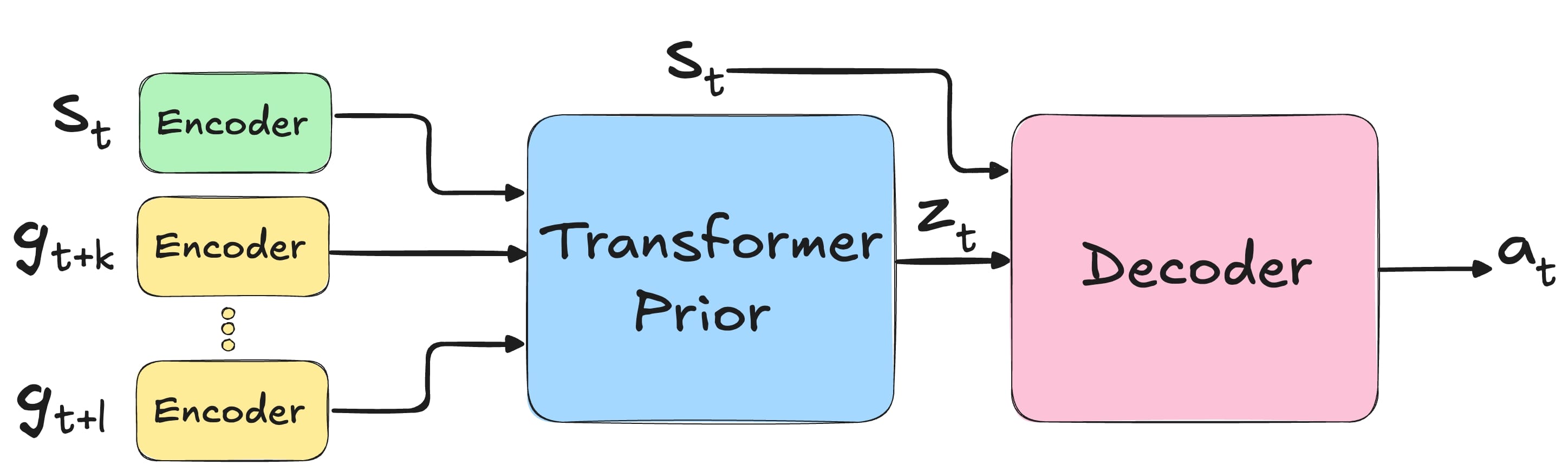}
        \caption{C-VAE: A prior is trained to represent the set of plausible solutions for each state $s_t$ and sparse objectives $g_t^\text{versatile}$.}
         \label{fig: vae}
    \end{subfigure}\hfill 
    \begin{subfigure}[b]{0.30\textwidth}
        \centering
        \includegraphics[width=\linewidth]{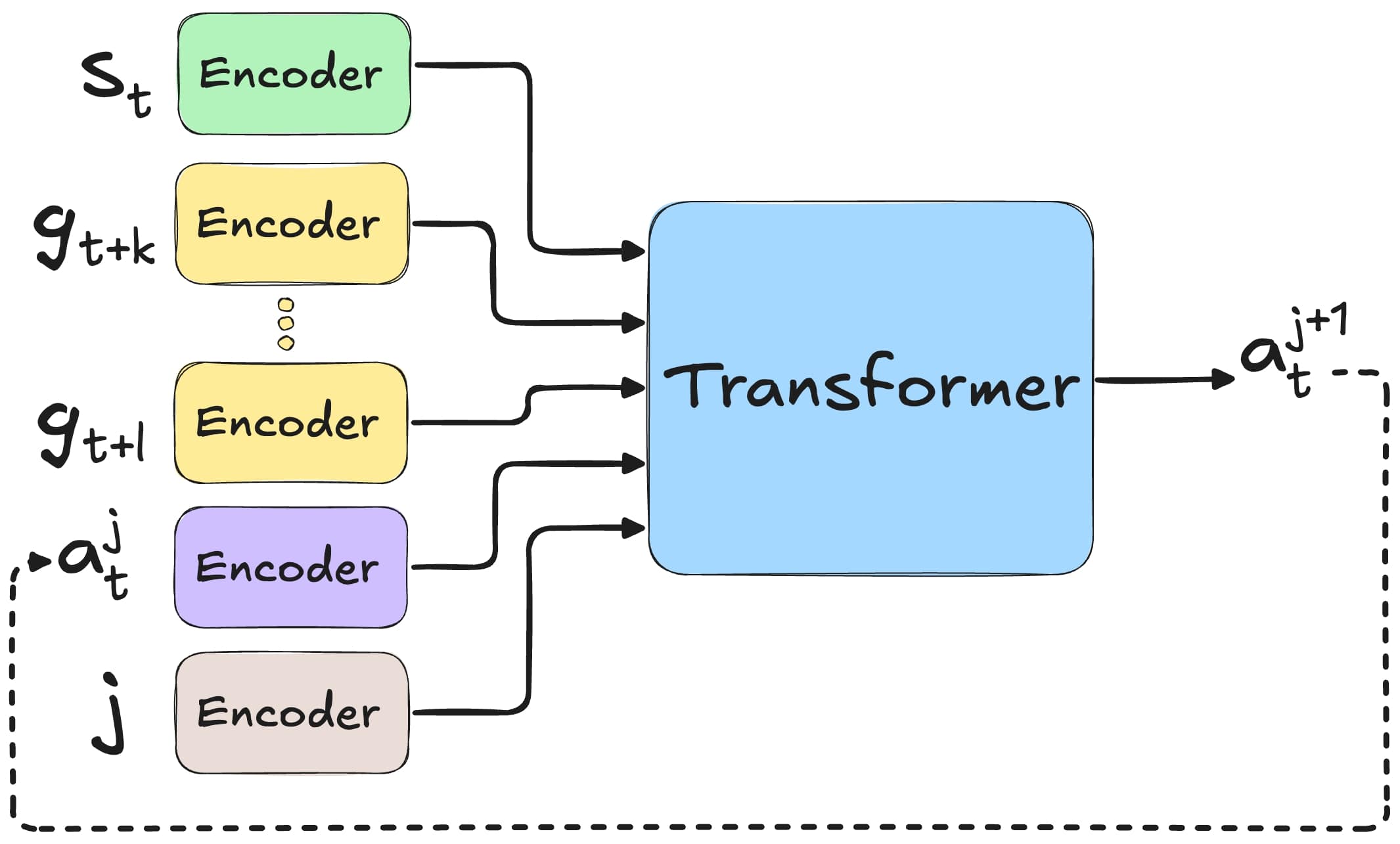}
        \caption{Diffusion: Conditioned on the noisy action $a_t^j$ and the current diffusion timestep $j$.}
         \label{fig: diffusion}
    \end{subfigure}
    \caption{\textbf{\generator\ architectures:} We illustrate the inference procedure of the considered architectures.}
    \label{fig: architectures}
    \vspace{-0.1cm}
\end{figure*}

The GRAB dataset consists of motion sequences collected across 10 different human subjects, differing by gender and size. This poses a challenge for training the controller, as it requires awareness and generalization across different morphologies. To simplfy the control problem, we use a single, canonical humanoid body shape (mean SMPL-X). The core difficulty lies in accurately retargeting the diverse source motions from different subjects to our canonical character while preserving the semantically salient characteristics of the original human-object interaction.

We propose a simple yet effective approach. First, we map the motion from source to target skeletons by transferring the local DoF rotations. \edited{Transferring rotations to a new body size leads to misaligned human-object motion. For example, when rotated downwards, a longer upperarm will result in the hand being beneath the object. Our second step is object retargeting. We extract contacts from the human-object aligned source motion. This tells us which body part is in contact with the object's mesh, and where. We then retarget the object's global position to best preserve these contact relationships with the canonical character's end-effectors (additional details in the supplementary material and illustrated in \cref{fig: retargeting}).}

Despite retargeting, some motions remain unsuitable for physical reconstruction. We filter these, primarily due to: Complex bimanual interactions where our object-centric retargeting may not fully resolve hand-object-hand alignment (\edited{e.g., transferring motion to a shorter character leads to a large gap between hands)}, and interactions reliant on features our simplified humanoid model lacks (e.g., placing sunglasses on a detailed face). This data processing, including retargeting and filtering, yields a training set of 1007 motion sequences and a test set of 141 sequences (GRAB subject 10).

\subsection{\generator}

The second stage of our framework distills the rich, physically-grounded interaction expertise learned by $\pi_\text{track}$ (from \tracker) into a versatile, generative control policy, $\pi_\text{versatile}$ (our \generator). The ultimate goal for \generator\ is to produce diverse, human-like, and physically plausible full-body manipulation behaviors in response to sparse future goals. These goals might specify the desired pose for a hand, the target location for an object, or a full-body stance at a future time, effectively leaving many other aspects of the motion for the policy to ``inpaint''.

Directly training such a versatile policy from scratch is exceptionally challenging. Sparse goals often lead to a misspecified objective with an overwhelmingly large space of potential solutions, posing significant exploration and credit assignment difficulties. However, \tracker\ ($\pi_{\text{track}}$), having learned to reconstruct a wide array of complex human demonstrations, implicitly captures a distribution of high-quality, physically feasible solutions to various interaction sub-tasks. Our approach extends the methodology of MaskedMimic \cite{tessler2024maskedmimic}, which demonstrated that framing control as a ``physics-based inpainting'' problem -- where the policy fills in missing parts of a motion sequence -- can effectively learn versatile behaviors from diverse data.

\paragraph{Goal Specification and Distillation:}
The core idea is to train \generator\ ($\pi_{\text{versatile}}$) to predict the detailed actions that \tracker\ ($\pi_{\text{track}}$) would execute to complete a motion, but conditioned only on the current state $s_t$ and a sparse set of future goals $g_t^{\text{versatile}}$. We employ an online teacher-student distillation process based on DAgger \cite{ross2011reduction}.

At each step, the student policy $\pi_{\text{versatile}}$ receives the current state $s_t$ and a randomly masked version of the future reference trajectory, $g_t^{\text{versatile}}$. This sparse goal $g_t^{\text{versatile}}$ specifies target future states (e.g., pose of a hand, object location, or full body stance) for a subset of entities at one or more future timesteps. All other future state information is masked. The student then predicts an action $a_t^\text{versatile}$ which is used to control the character in the simulation.

Concurrently, the teacher policy $\pi_{\text{track}}$ observes the same current state $s_t$ but with the unmasked, dense future reference trajectory $g_t^{\text{track}}$ (as used during its own training). The distillation objective is to minimize the L2 distance between the actions predicted by the student and the teacher:
\begin{equation}
\mathcal{L}_{\text{distill}} = - \log \pi_{\text{versatile}}\left( a_t^{\text{track}} | s_t, g_t^{\text{versatile}} \right)
\label{eq:distill_loss}
\end{equation}
We adapt the structured masking scheme from MaskedMimic \cite{tessler2024maskedmimic} to define $g_t^{\text{versatile}}$, crucially extending it to include conditioning on the future pose of the manipulated object alongside humanoid body parts.

\paragraph{Policy Architecture and Training:}
To accommodate a variable number of sparsely specified future goal constraints, each defined by a target pose and a future timestep, \generator\ utilizes a transformer architecture \cite{vaswani2017attention}. The input to the transformer is a sequence of tokens. Fixed-information inputs, such as current proprioception (joint states, root state), hand-to-object vectors, object and table pose, and object and table BPS, are jointly mapped to a shared token. Each specified future goal (e.g., target pose of hand $j$ at time $t+k$) is also encoded into its own token. Which each target future pose is represented using a unique token, they are generated using a shared encoder.

Given that a sparse goal $g_t^{\text{versatile}}$ can often be achieved through multiple valid and human-like motion sequences (i.e., the problem is multi-modal), we experiment with three classes of policy architectures for $\pi_{\text{versatile}}$, illustrated in \cref{fig: architectures}:

\textbf{(1) Deterministic:} This standard approach learns to predict a single, mean action based on the teacher's demonstrations. While computationally efficient, this can average over the diverse solutions present in the data, potentially leading to less expressive or less performant behaviors when multiple valid options exist.

\textbf{(2) Conditional Variational Autoencoder (C-VAE):} To explicitly model the multi-modality of solutions, we employ a C-VAE architecture with a learned prior, drawing inspiration from \cite{rempe2021humor,tessler2024maskedmimic}. The prior network takes the sparse goal $g_t^{\text{versatile}}$ and current state $s_t$ to predict a latent distribution $\mathcal{N}(\mu_{prior}, \sigma_{prior})$ representing plausible general solutions. During training, a residual encoder provides the offset in the latent space, predicting the precise solution from the reference data. We extend the architecture in MaskedMimic. There, the encoder observes both the reference motion $g_t^{\text{track}}$ and the sparse goals $g_t^{\text{versatile}}$. As a result, the encoder is required to implicitly predict both the prior distribution and the offset. We simplify this design by providing the encoder with the prior's output $\mathcal{N}(\mu_{prior}, \sigma_{prior})$. We found it reduces parameter count and improves performance. During inference, sampling from the learned prior allows \generator\ to generate diverse actions satisfying the sparse goal.

\begin{figure*}[t]
     \centering
     \begin{subfigure}[b]{0.164\textwidth}
         \centering
         \includegraphics[trim={10cm 4cm 10cm 4cm},clip,width=1\textwidth]{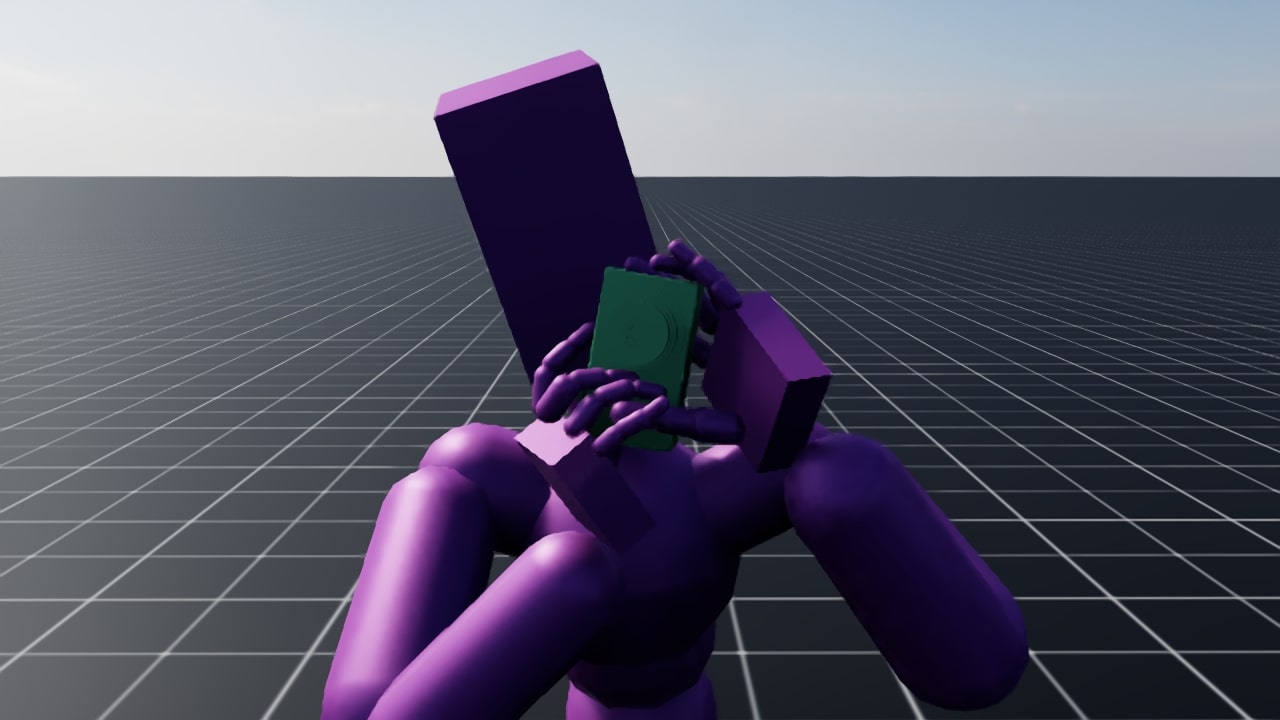}
         \caption{{Taking a picture}}
         \label{fig: camera}
    \end{subfigure}\hfill
    \begin{subfigure}[b]{0.164\textwidth}
         \centering
         \includegraphics[trim={8cm 4cm 12cm 4cm},clip,width=1\textwidth]{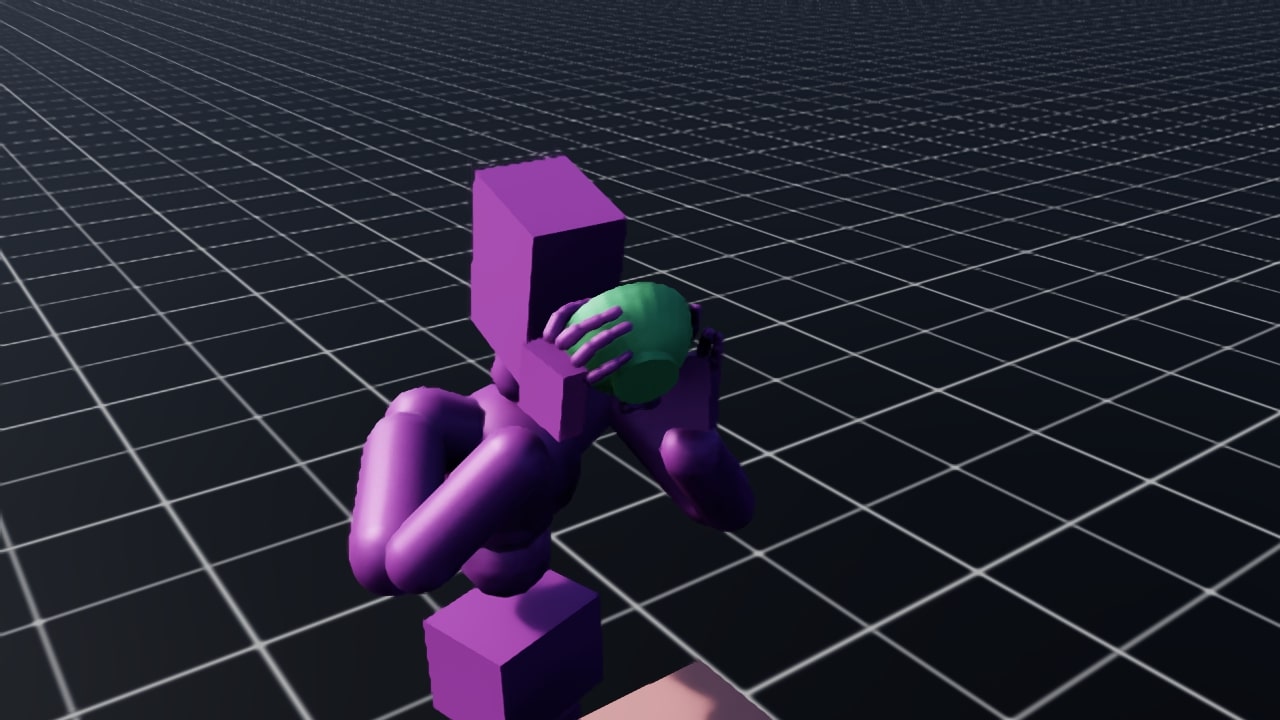}
         \caption{{Drinking soup}}
         \label{fig: bowl}
    \end{subfigure}\hfill
    \begin{subfigure}[b]{0.164\textwidth}
         \centering
         \includegraphics[trim={10cm 6cm 10cm 2cm},clip,width=1\textwidth]{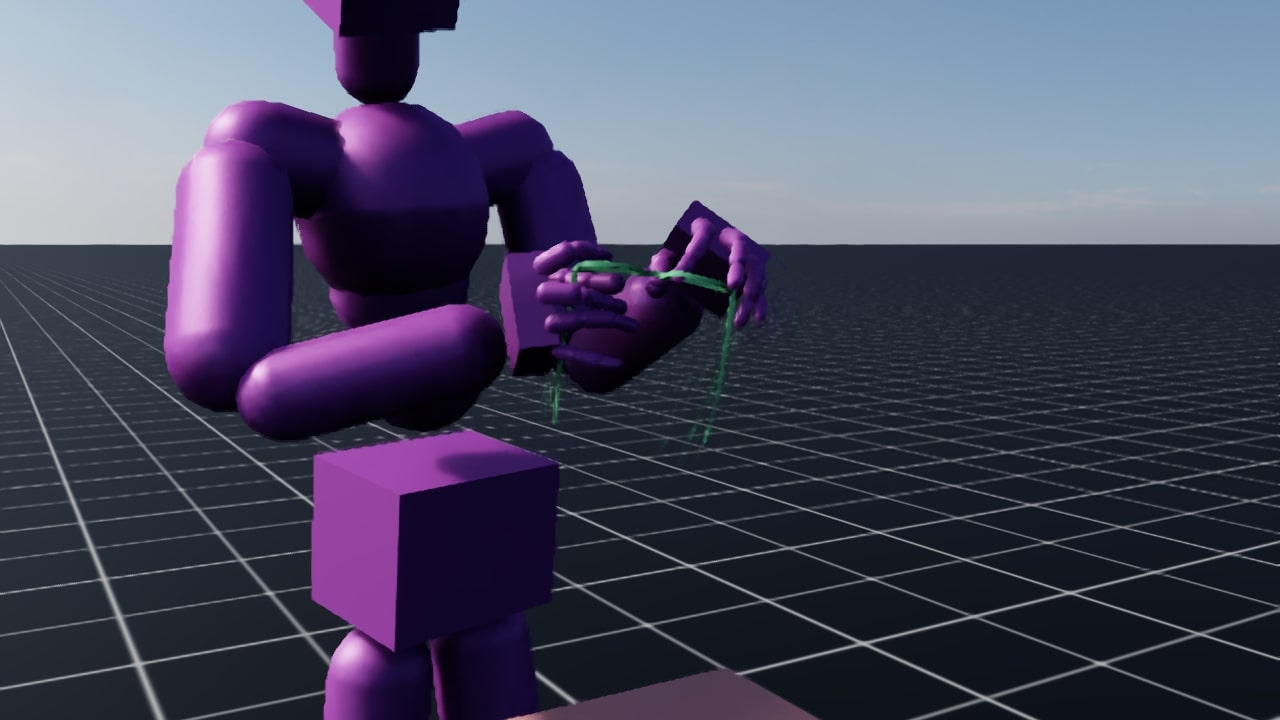}
         \caption{{Cleaning glasses}}
         \label{fig: glasses}
    \end{subfigure}\hfill
    \begin{subfigure}[b]{0.164\textwidth}
         \centering
         \includegraphics[trim={10cm 4cm 10cm 4cm},clip,width=1\textwidth]{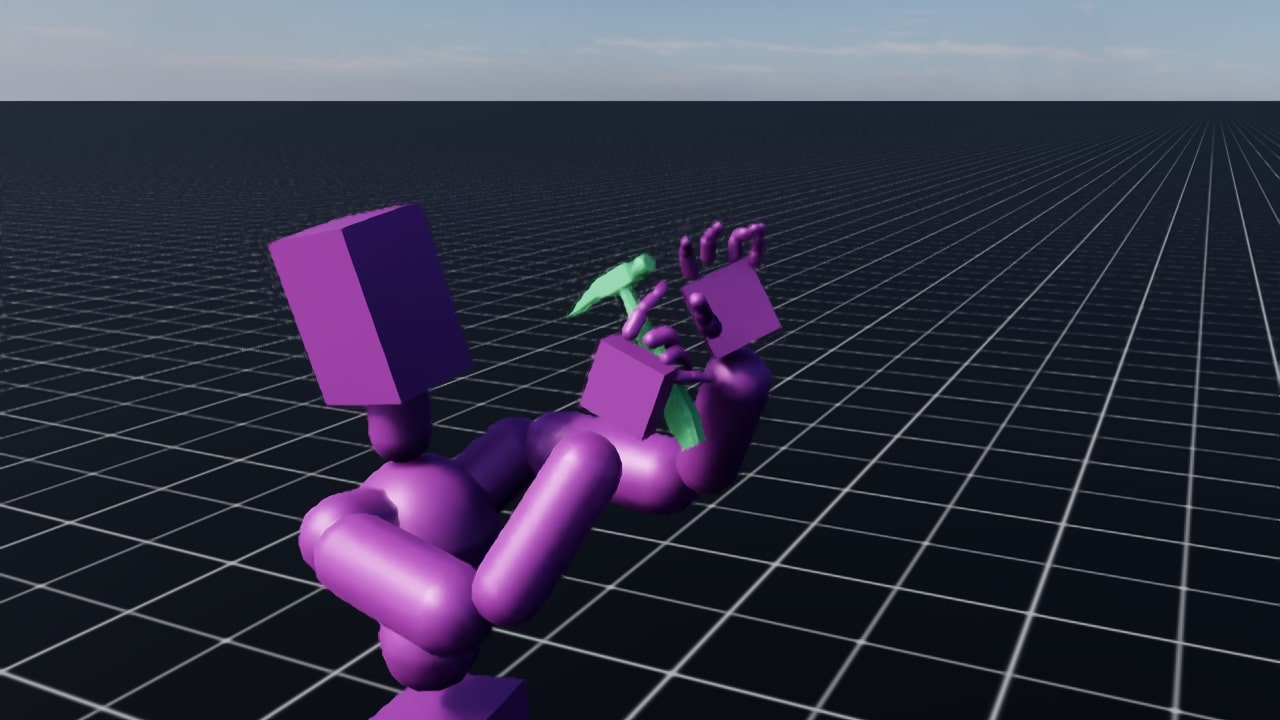}
         \caption{{Using a hammer}}
         \label{fig: hammer}
    \end{subfigure}\hfill
    \begin{subfigure}[b]{0.164\textwidth}
         \centering
         \includegraphics[trim={10cm 8cm 10cm 0cm},clip,width=1\textwidth]{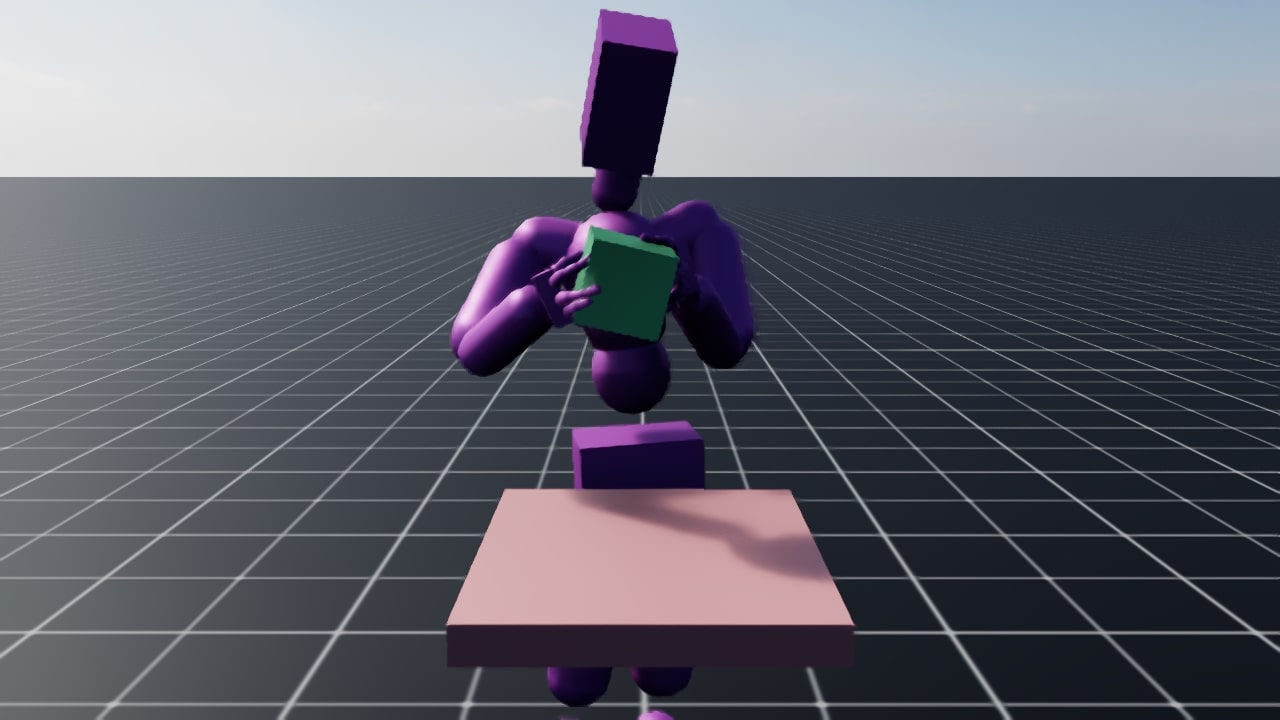}
         \caption{{Inspecting a cube}}
         \label{fig: large cube}
    \end{subfigure}\hfill
    \begin{subfigure}[b]{0.164\textwidth}
         \centering
         \includegraphics[trim={10cm 6cm 10cm 2cm},clip,width=1\textwidth]{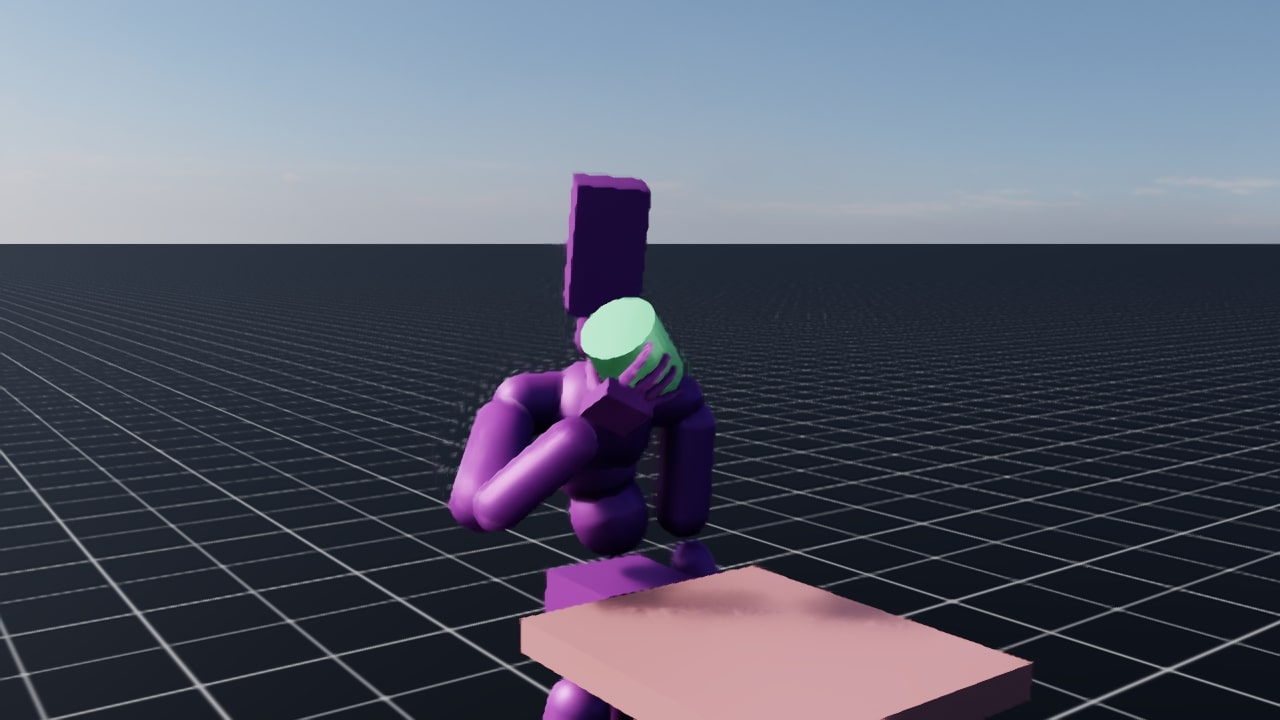}
         \caption{{Inspecting object}}
         \label{fig: large cylinder}
    \end{subfigure}\\
    \begin{subfigure}[b]{0.164\textwidth}
         \centering
         \includegraphics[trim={10cm 0cm 10cm 8cm},clip,width=1\textwidth]{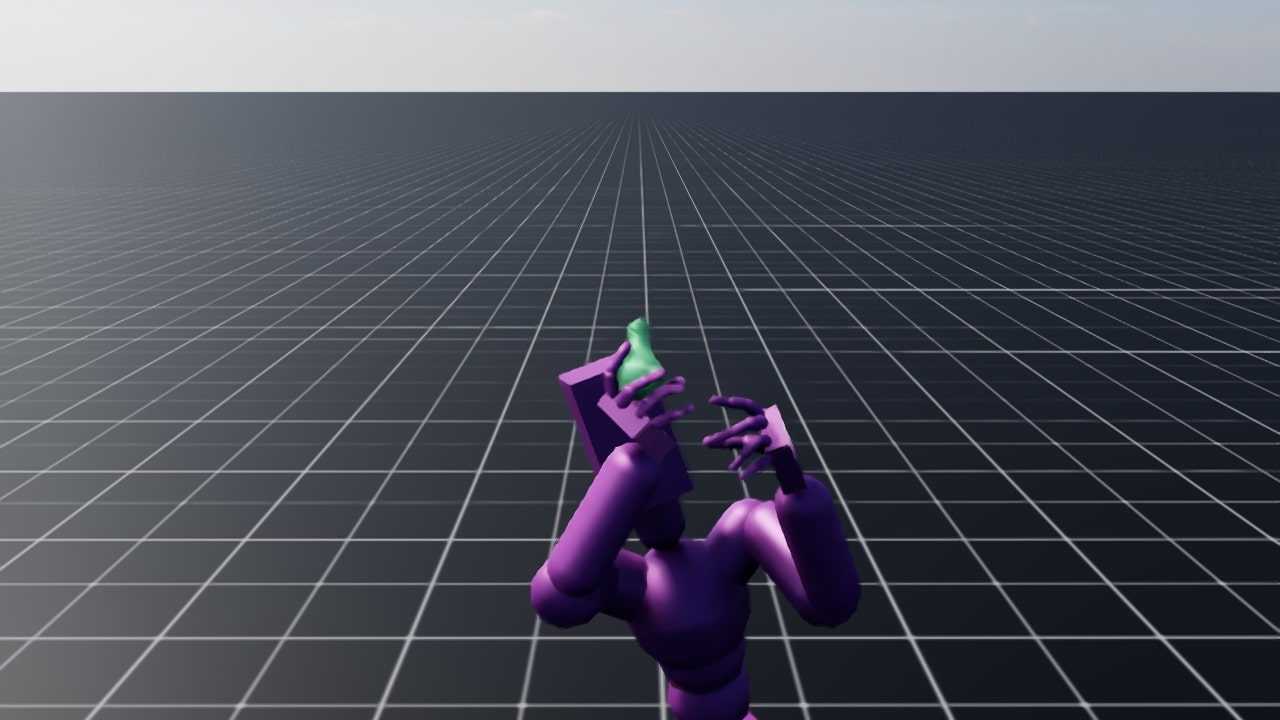}
         \caption{{Screwing lightbulb}}
         \label{fig: lightbulb}
    \end{subfigure}\hfill
    \begin{subfigure}[b]{0.164\textwidth}
         \centering
         \includegraphics[trim={10cm 10.1cm 13cm 0cm},clip,width=1\textwidth]{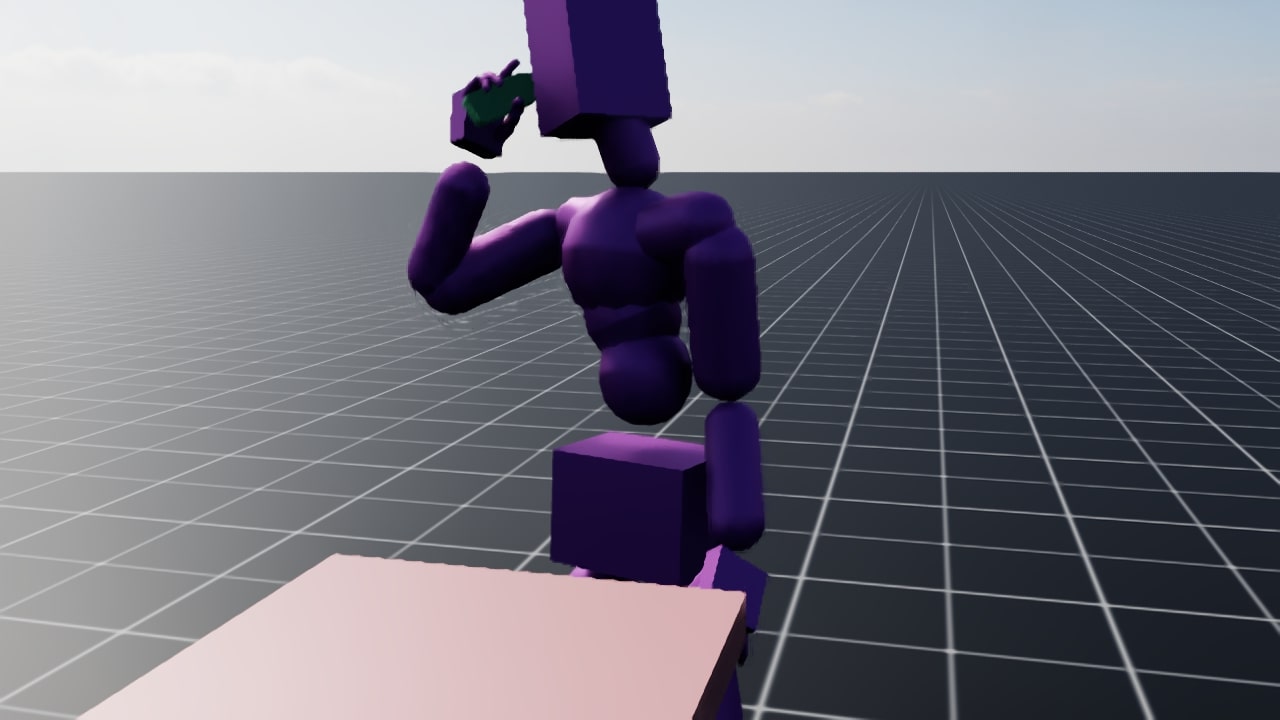}
         \caption{{Using a phone}}
         \label{fig: phone}
    \end{subfigure}\hfill
    \begin{subfigure}[b]{0.164\textwidth}
         \centering
         \includegraphics[trim={10cm 6cm 10cm 2cm},clip,width=1\textwidth]{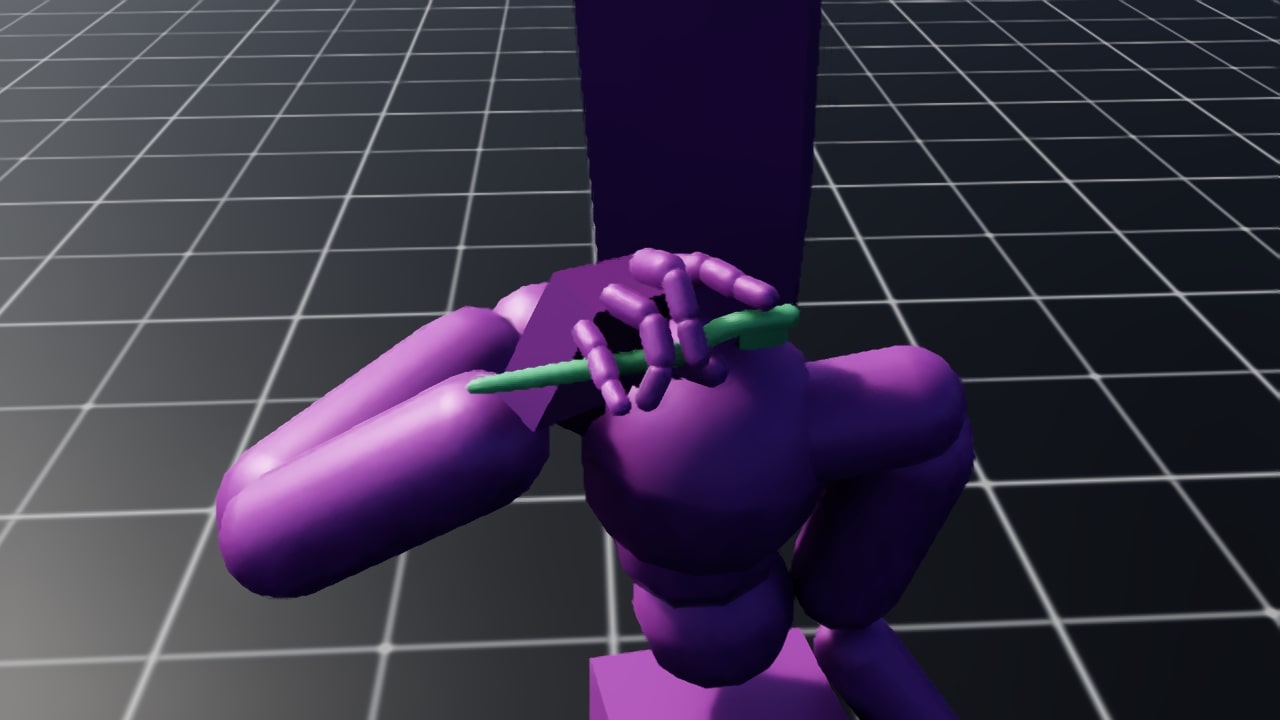}
         \caption{{Brushing teeth}}
         \label{fig: toothbrush}
    \end{subfigure}\hfill
    \begin{subfigure}[b]{0.164\textwidth}
         \centering
         \includegraphics[trim={10cm 4cm 10cm 4cm},clip,width=1\textwidth]{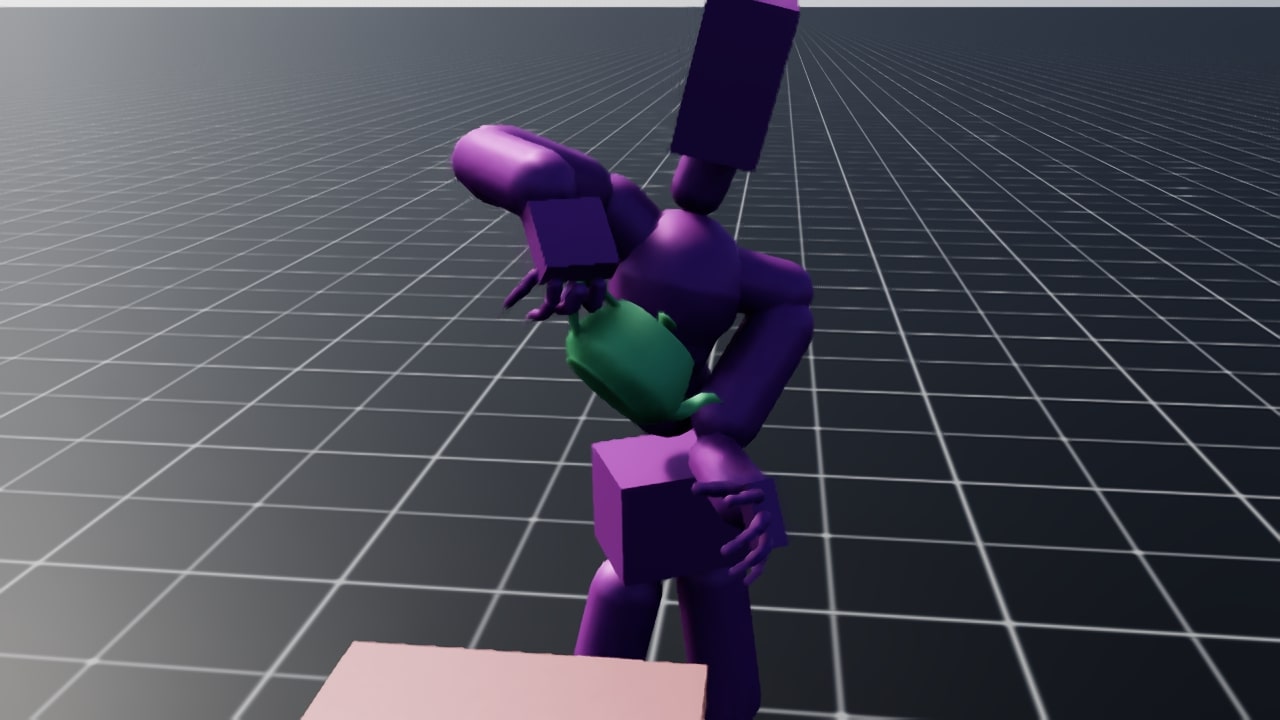}
         \caption{{Pouring tea}}
         \label{fig: teapot}
    \end{subfigure}\hfill
    \begin{subfigure}[b]{0.164\textwidth}
         \centering
         \includegraphics[trim={8cm 4cm 12cm 4cm},clip,width=1\textwidth]{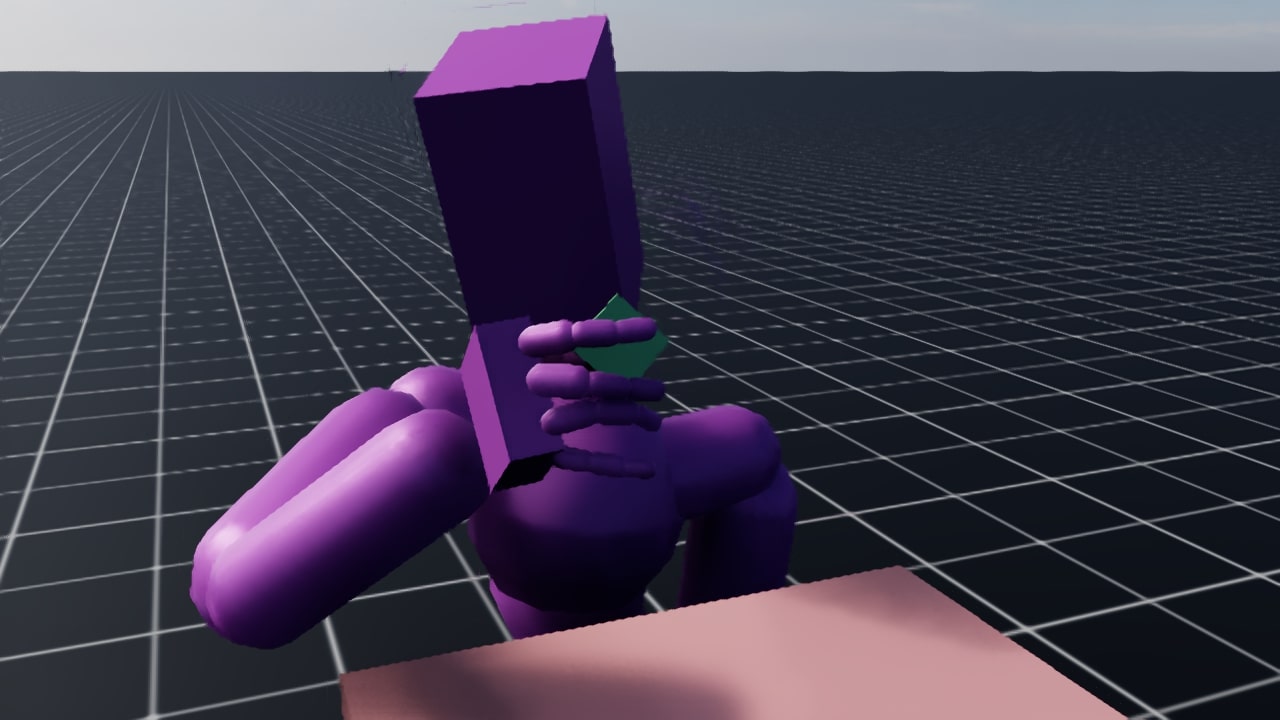}
         \caption{{Inspecting a cube}}
         \label{fig: small cube}
    \end{subfigure}\hfill
    \begin{subfigure}[b]{0.164\textwidth}
         \centering
         \includegraphics[trim={16cm 14cm 17cm 3cm},clip,width=1\textwidth]{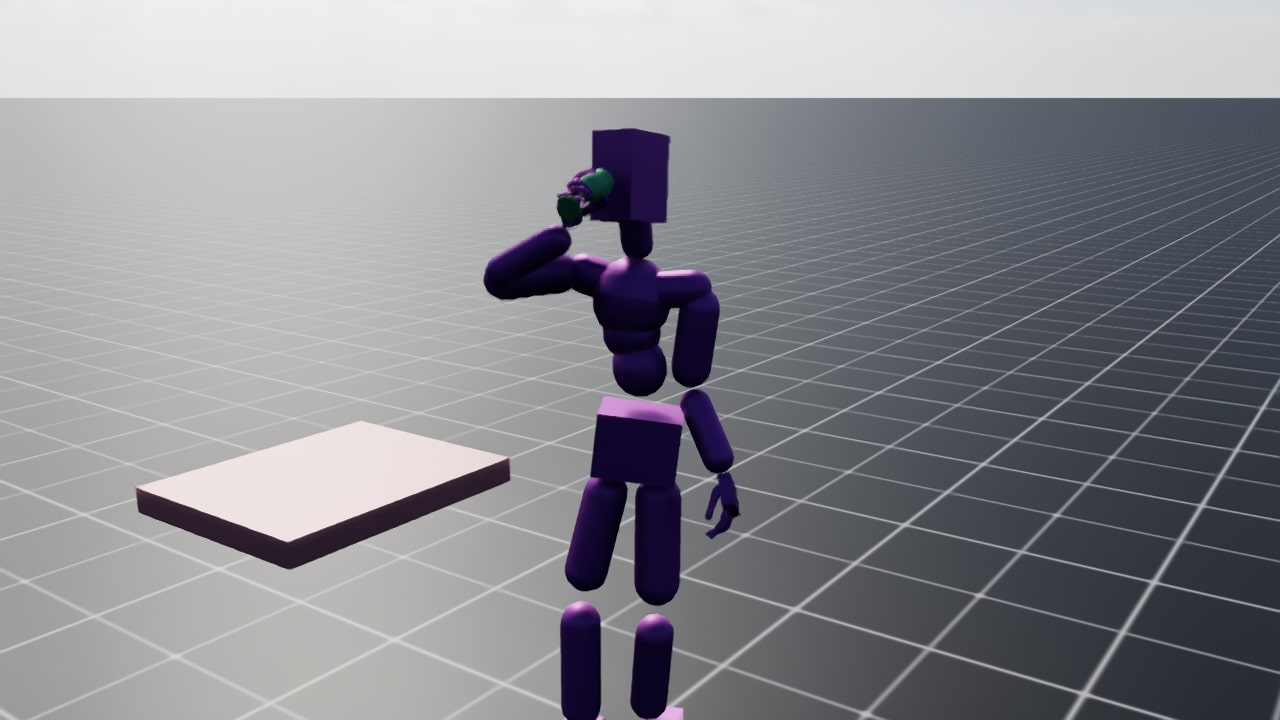}
         \caption{{Drinking wine}}
         \label{fig: wineglass}
    \end{subfigure}\\
     \begin{subfigure}[b]{\textwidth}
         \centering
         \includegraphics[trim={14cm 8cm 12cm 4cm},clip,width=0.164\textwidth]{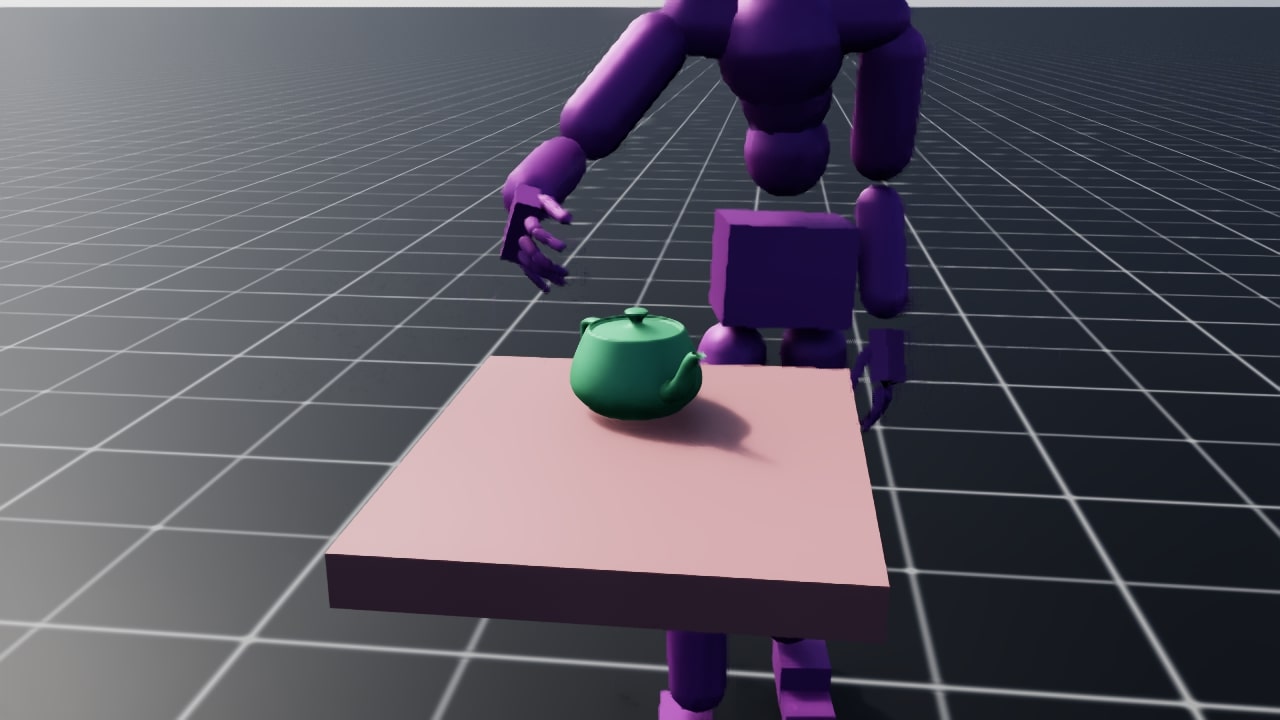}\hfill
         \includegraphics[trim={14cm 8cm 12cm 4cm},clip,width=0.164\textwidth]{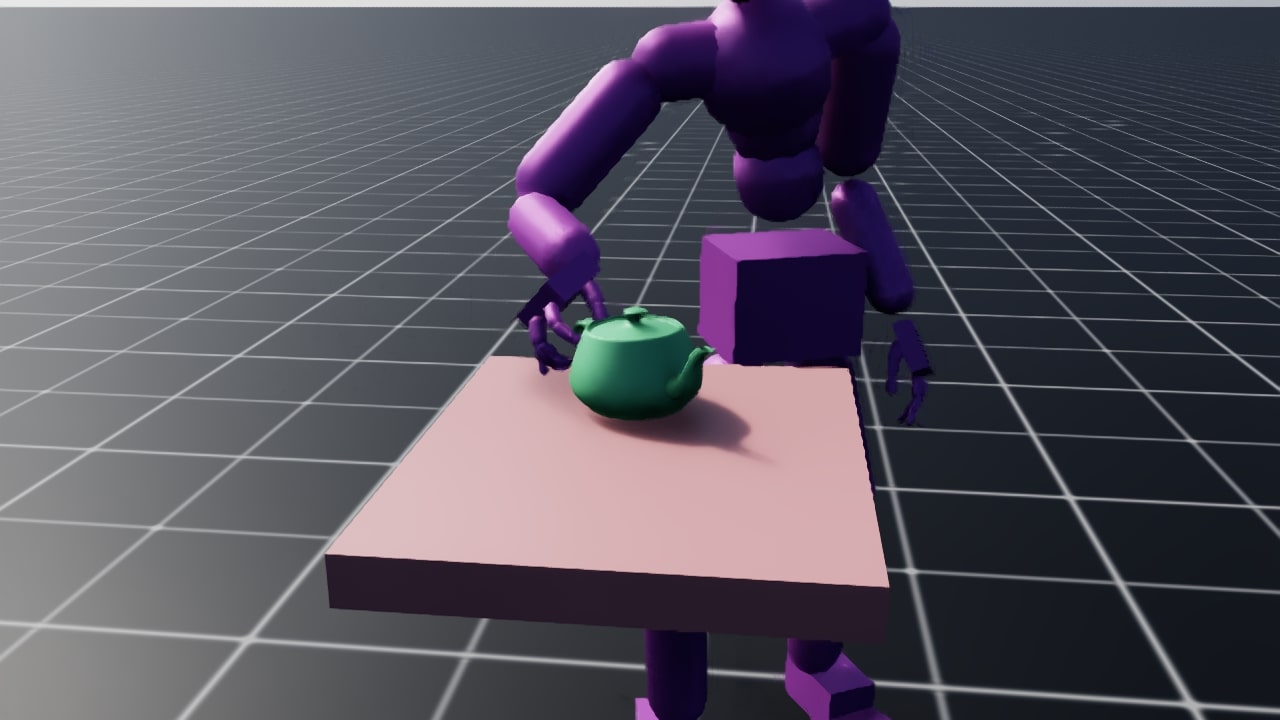}\hfill
         \includegraphics[trim={16cm 7cm 10cm 5cm},clip,width=0.164\textwidth]{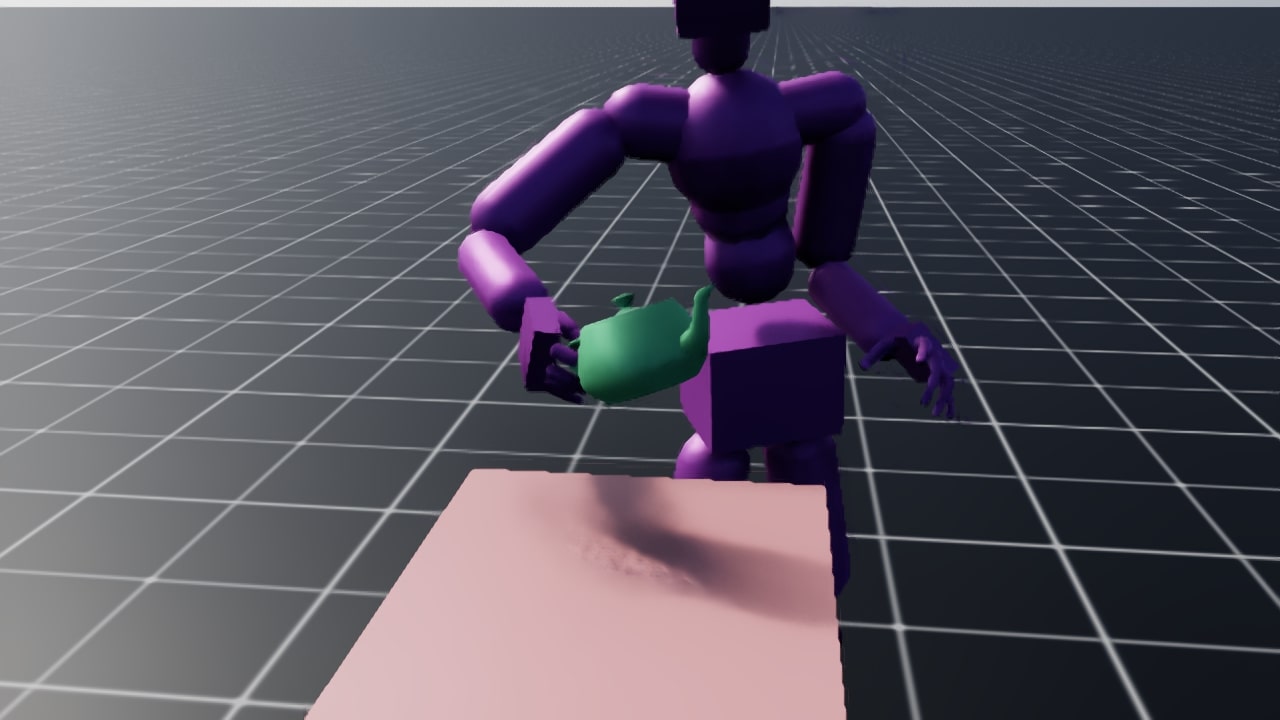}\hfill
         \includegraphics[trim={14cm 4cm 12cm 8cm},clip,width=0.164\textwidth]{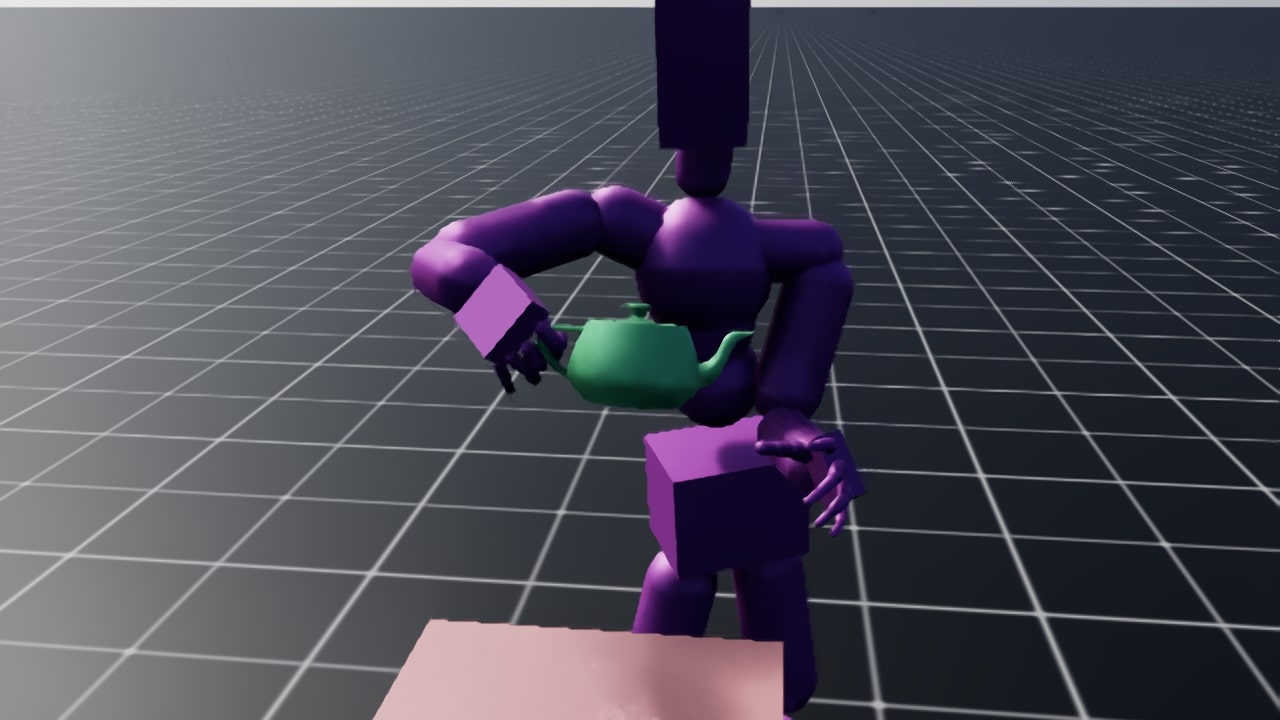}\hfill
         \includegraphics[trim={14cm 4cm 12cm 8cm},clip,width=0.164\textwidth]{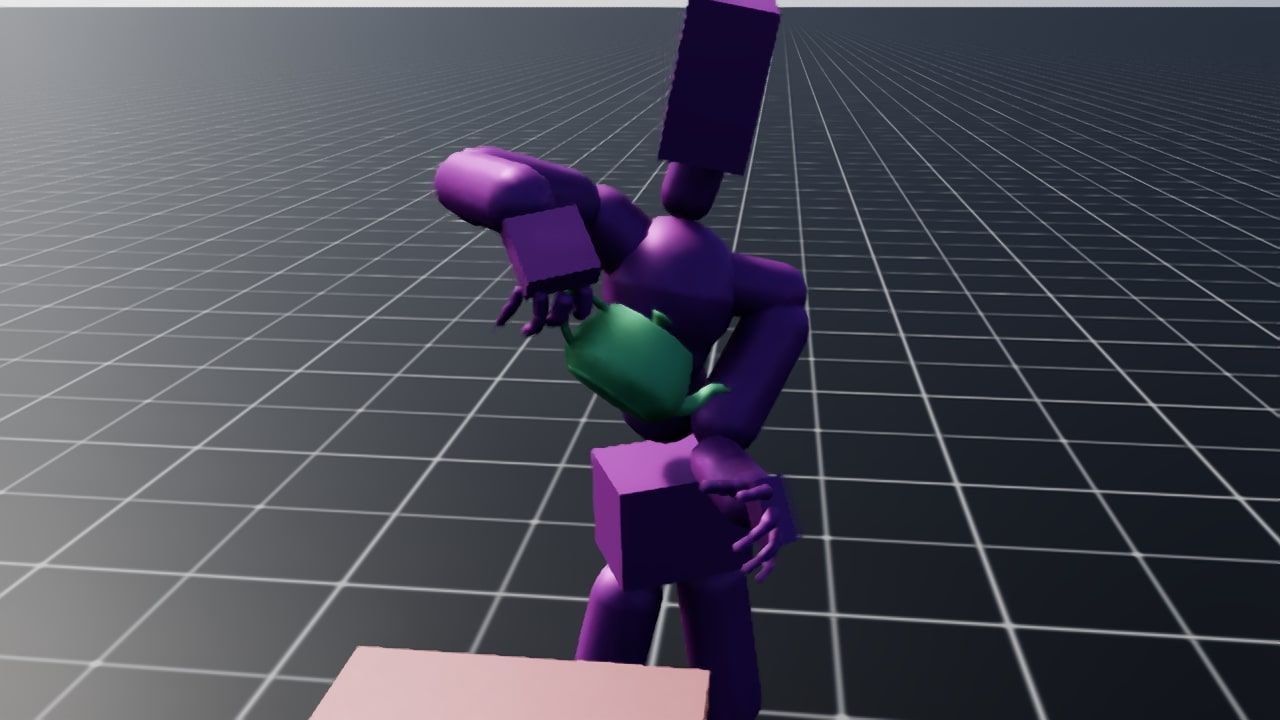}\hfill
         \includegraphics[trim={14cm 7cm 12cm 5cm},clip,width=0.164\textwidth]{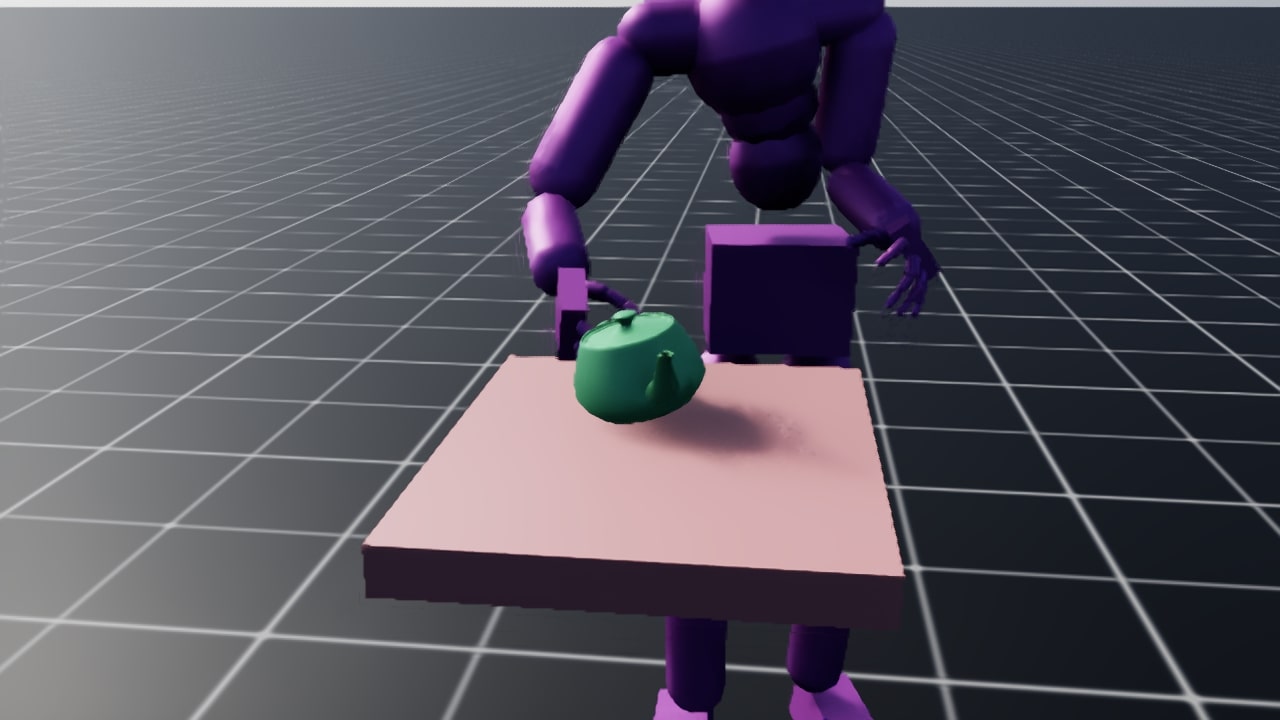}
         \caption{\textbf{Pouring tea:} \tracker\ picks up the teapot by the handle to imitate a pouring motion into an imaginary cup.}
         \label{fig: full body pouring tea}
    \end{subfigure}\\
    \begin{subfigure}[b]{\textwidth}
         \centering
         \includegraphics[trim={14cm 7cm 12cm 5cm},clip,width=0.164\textwidth]{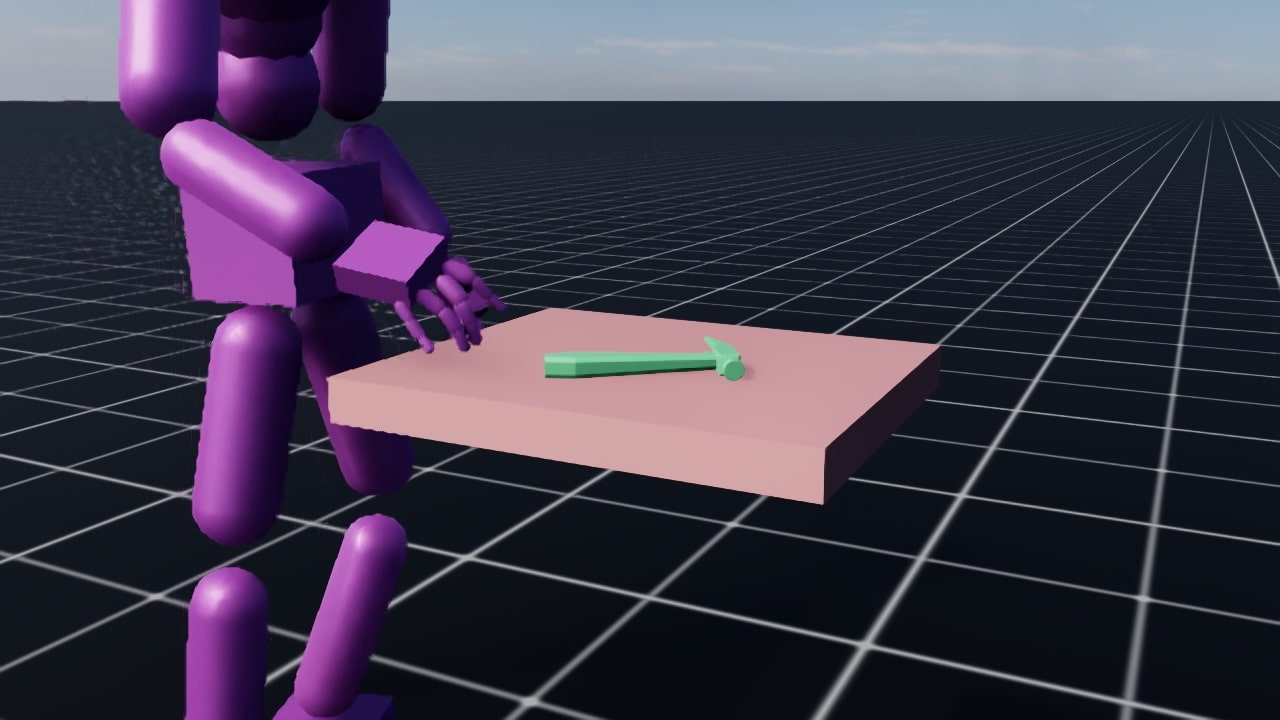}\hfill
         \includegraphics[trim={14cm 7cm 12cm 5cm},clip,width=0.164\textwidth]{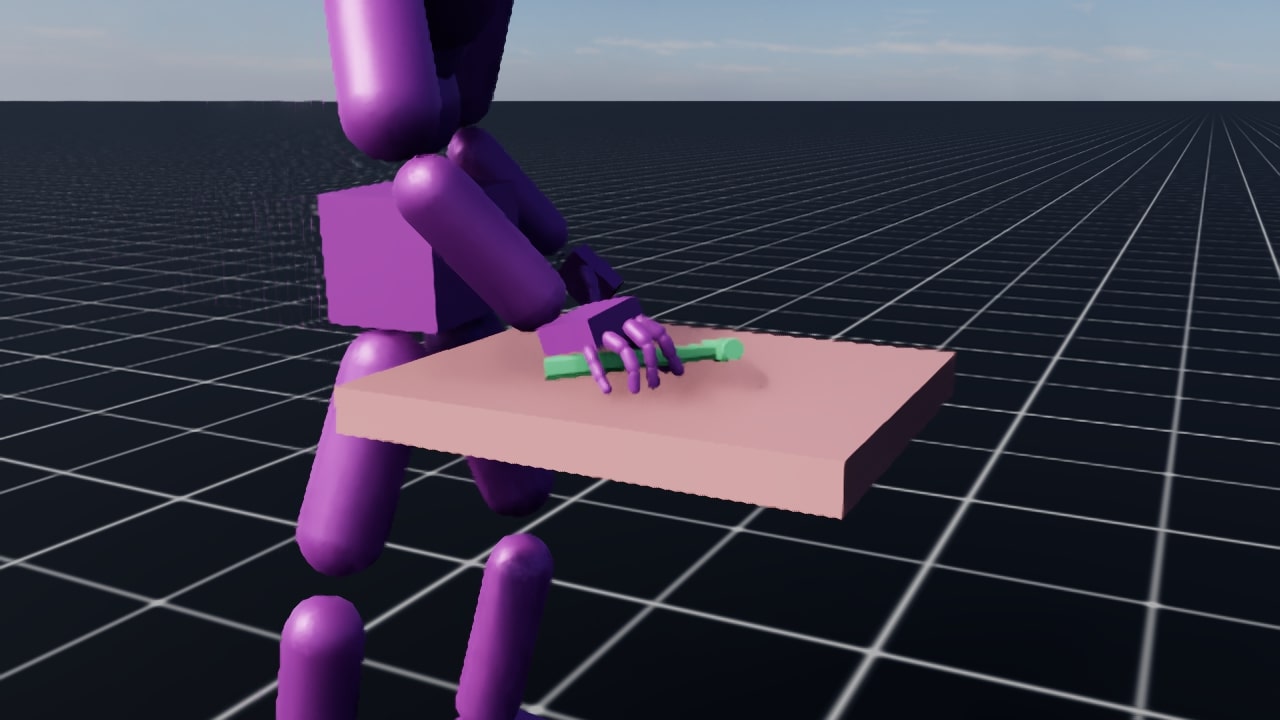}\hfill
         \includegraphics[trim={14cm 7cm 12cm 5cm},clip,width=0.164\textwidth]{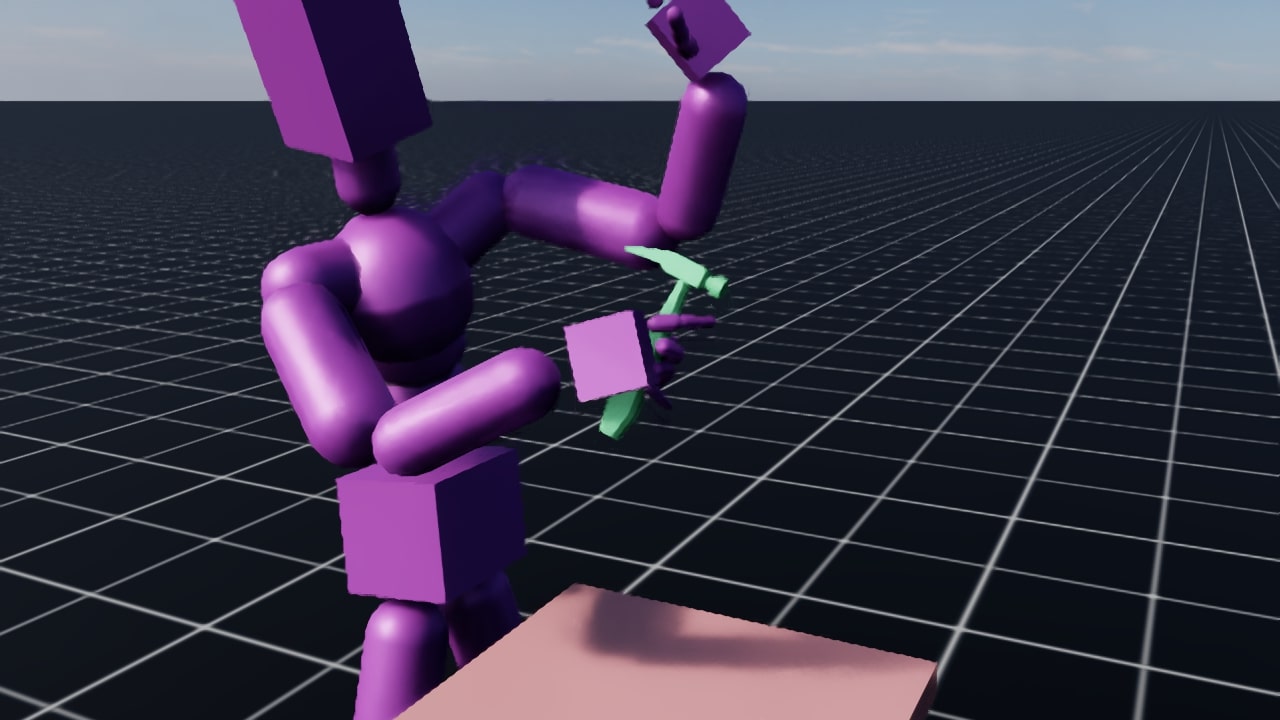}\hfill
         \includegraphics[trim={14cm 7cm 12cm 5cm},clip,width=0.164\textwidth]{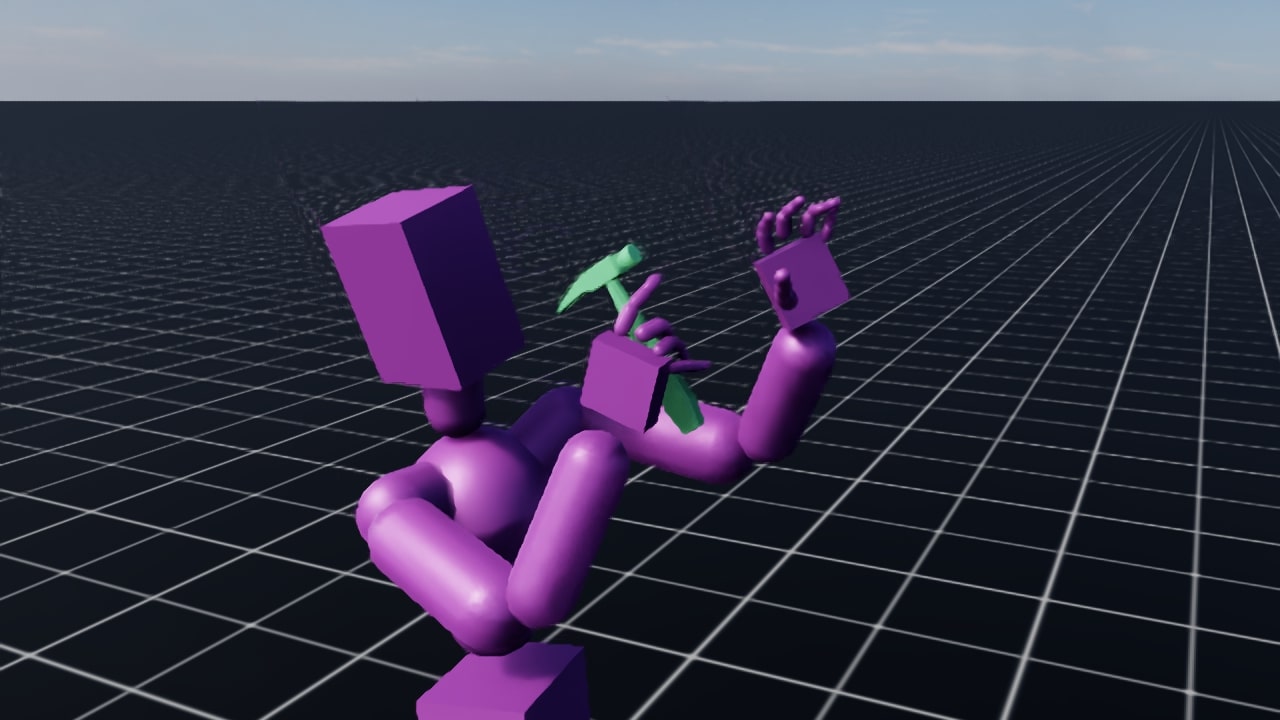}\hfill
         \includegraphics[trim={14cm 7cm 12cm 5cm},clip,width=0.164\textwidth]{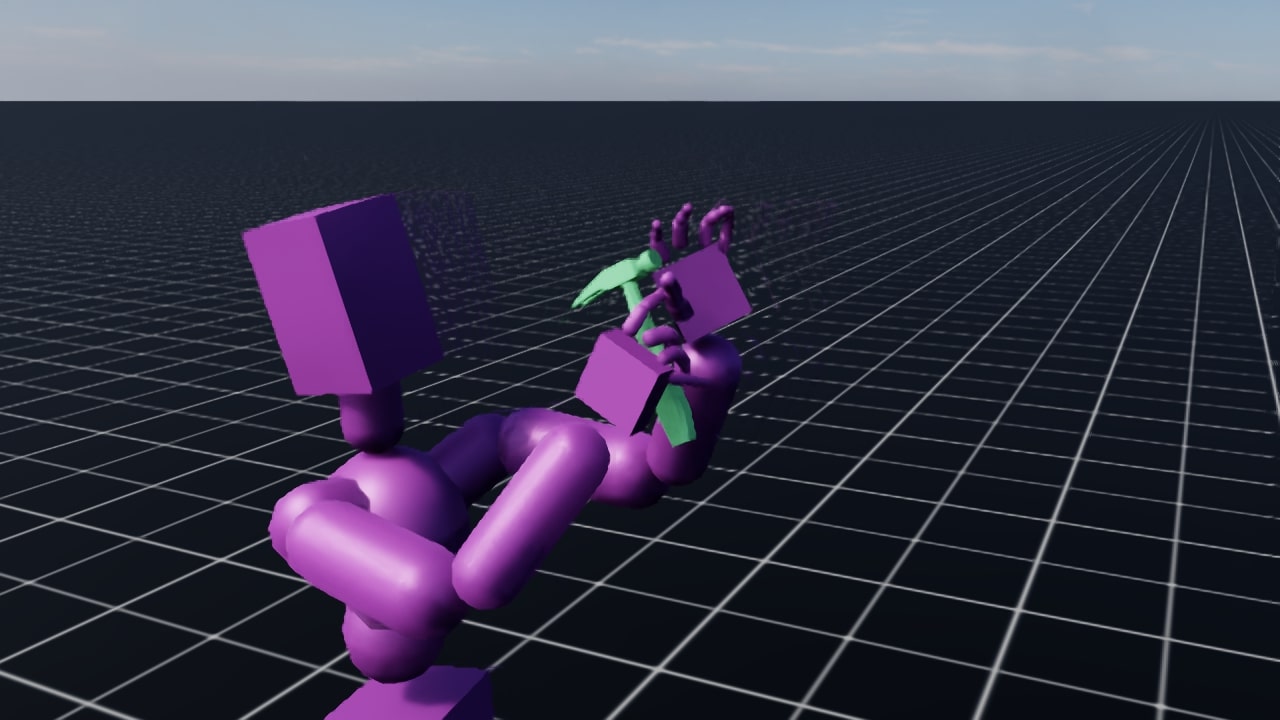}\hfill
         \includegraphics[trim={14cm 7cm 12cm 5cm},clip,width=0.164\textwidth]{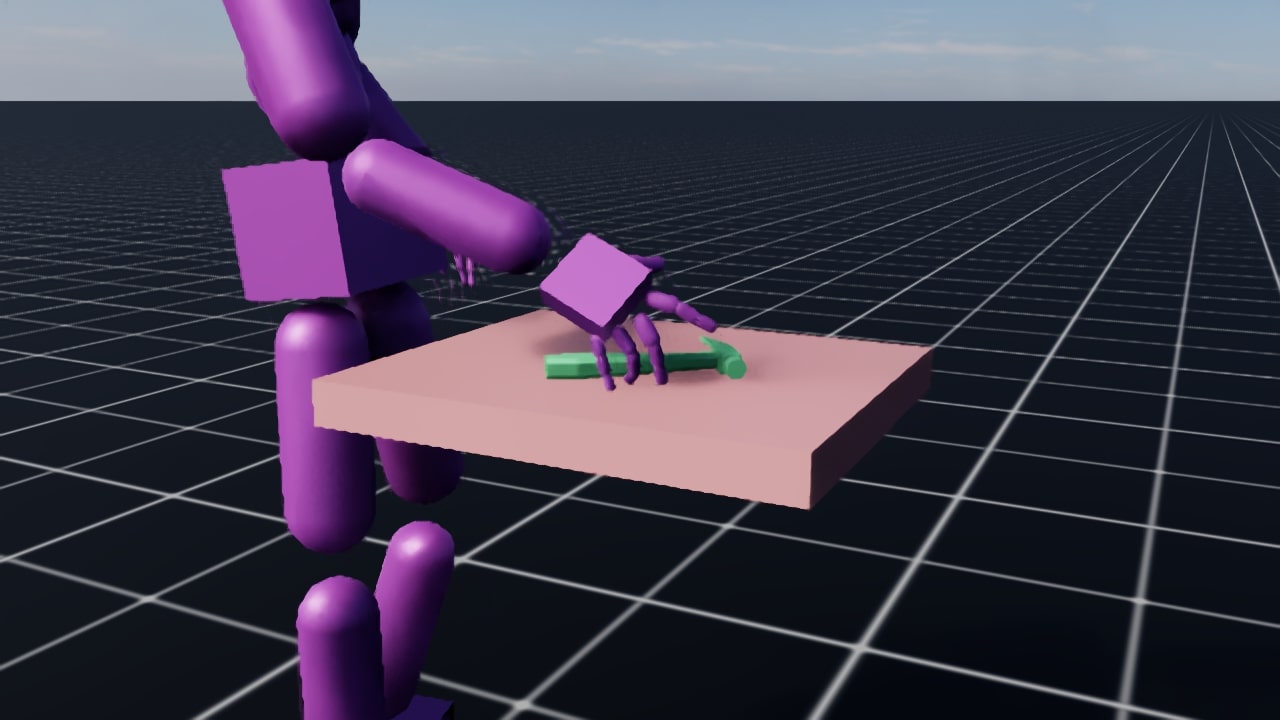}
         \caption{\textbf{Using a hammer:} \tracker\ precisely handles a hammer, maintaining proper orientation while demonstrating how to hit a nail.}
         \label{fig: full body banging a nail}
    \end{subfigure}
    \caption{\textbf{\tracker\ -- full-body tracking:} \tracker\ successfully interacts with a wide range of objects, reconstructing various daily human behaviors. \edited{\textit{Textual descriptions are provided for the reader's convenience. The model itself is not conditioned on textual labels.}}}
    \label{fig: full body tracker}
\end{figure*}

\textbf{(3) Diffusion Policy:} As an additional generative approach, we investigate a diffusion-based policy \cite{chi2023diffusion,ho2020denoising,lu2025what}. Unlike C-VAEs, diffusion models do not require an explicit prior network or separate encoder for the posterior during training for their generative process. The policy is trained to denoise an action that is initially pure Gaussian noise, iteratively refining it over $N$ steps to produce a clean action $a_t^N$. This iterative refinement process is conditioned on the current state $s_t$ and sparse goal $g_t^{\text{versatile}}$. To provide the necessary conditioning for each denoising step $j$, we augment our transformer architecture with two additional input heads: one tokenizing the noisy action $a_t^j$ from the current step, and another representing the current denoising timestep $j$. At each step $j$, the policy predicts the less noisy action $a_t^{j-1}$ (or equivalently, the noise to remove). The final denoised action $a_t^0$ (conventionally, or $a_t^N$ if $N$ is the start of denoising from pure noise) is used to control the character. In addition to Diffusion using DAgger, we also experiment with a fully-offline version, where \generator\ is trained using supervised learning on trajectories collected by \tracker.

Both the C-VAE and Diffusion approaches are explicitly chosen for their capacity to model the inherent diversity and multi-modality in the solution space when presented with sparse goals. This empowers \generator\ not merely to fulfill specified constraints, but to do so with a rich variety of human-like behaviors, reflecting the breadth of solutions learned by \tracker.

\section{Experiments}\label{sec: experiments}

\paragraph{Experimental setup:} All experiments are conducted using the ProtoMotions framework \cite{ProtoMotions} with IsaacLab \cite{mittal2023orbit} as the underlying physics simulator. The simulation runs at 120 FPS, and with 4 physics decimation steps, our control policies operate at an effective rate of 30 FPS. To manage interpenetrations robustly, we set the simulation's depenetration velocity to 100. To better approximate the contact dynamics of human soft tissue interacting with objects using rigid body simulation, we use a global friction coefficient of 1.5.

\paragraph{Training details:} All \tracker\ tracking policies ($\pi_{\text{track}}$) are trained using Proximal Policy Optimization (PPO) \cite{schulman2017proximal}. The \generator\ variants ($\pi_{\text{versatile}}$) are trained using DAgger \cite{ross2011reduction} with $\pi_{\text{track}}$ as the expert teacher. Each policy variant is trained for approximately 40,000 episodes, distributed across 8 NVIDIA A100 GPUs. Specific hyperparameters for PPO and the distillation process are detailed in the Appendix.

\paragraph{Evaluation metrics:} The full-body-manipulation sequences from GRAB are often long and complex, involving multiple pick-and-place iterations, and hand-to-hand transfers. To comprehensively evaluate performance, we employ several metrics:

\textit{Full-Sequence Success Rate:} The percentage of test sequences successfully completed from start to finish.
    
\textit{First-Interaction Success Rate:} To isolate early interaction performance from later complexities, this measures success from the motion's start until the second required contact with the object.

\textit{Mean Per Joint Position Error (MPJPE):} The average Euclidean distance between the simulated joint positions and the reference motion's joint positions, providing a measure of tracking fidelity.

\textit{Average Sequence Length:} The average episode length until termination is encountered (or episode ends).

We define failure (leading to episode termination for success rate calculation) if any humanoid joint deviates by more than 50cm from its reference position, or if the manipulated object deviates by more than 25cm from its reference position \cite{luo2024omnigrasp}.

\section{Results}

\begin{figure*}[t]
     \centering
     \begin{subfigure}[b]{0.495\textwidth}
         \centering
         \includegraphics[trim={14cm 10cm 40cm 3.9cm},clip,width=0.33\textwidth]{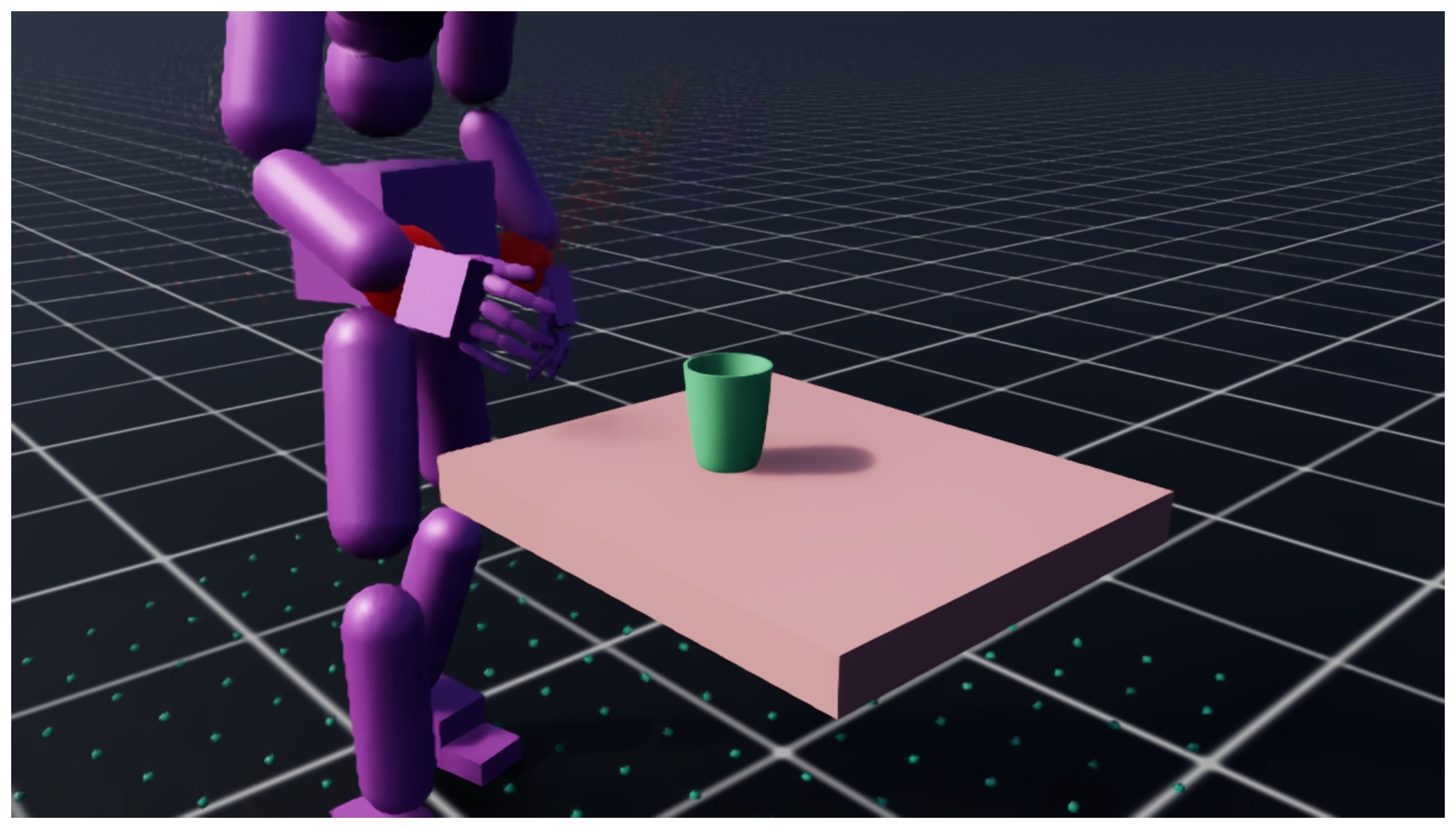}\hfill
         \includegraphics[trim={18cm 10cm 36cm 3.9cm},clip,width=0.33\textwidth]{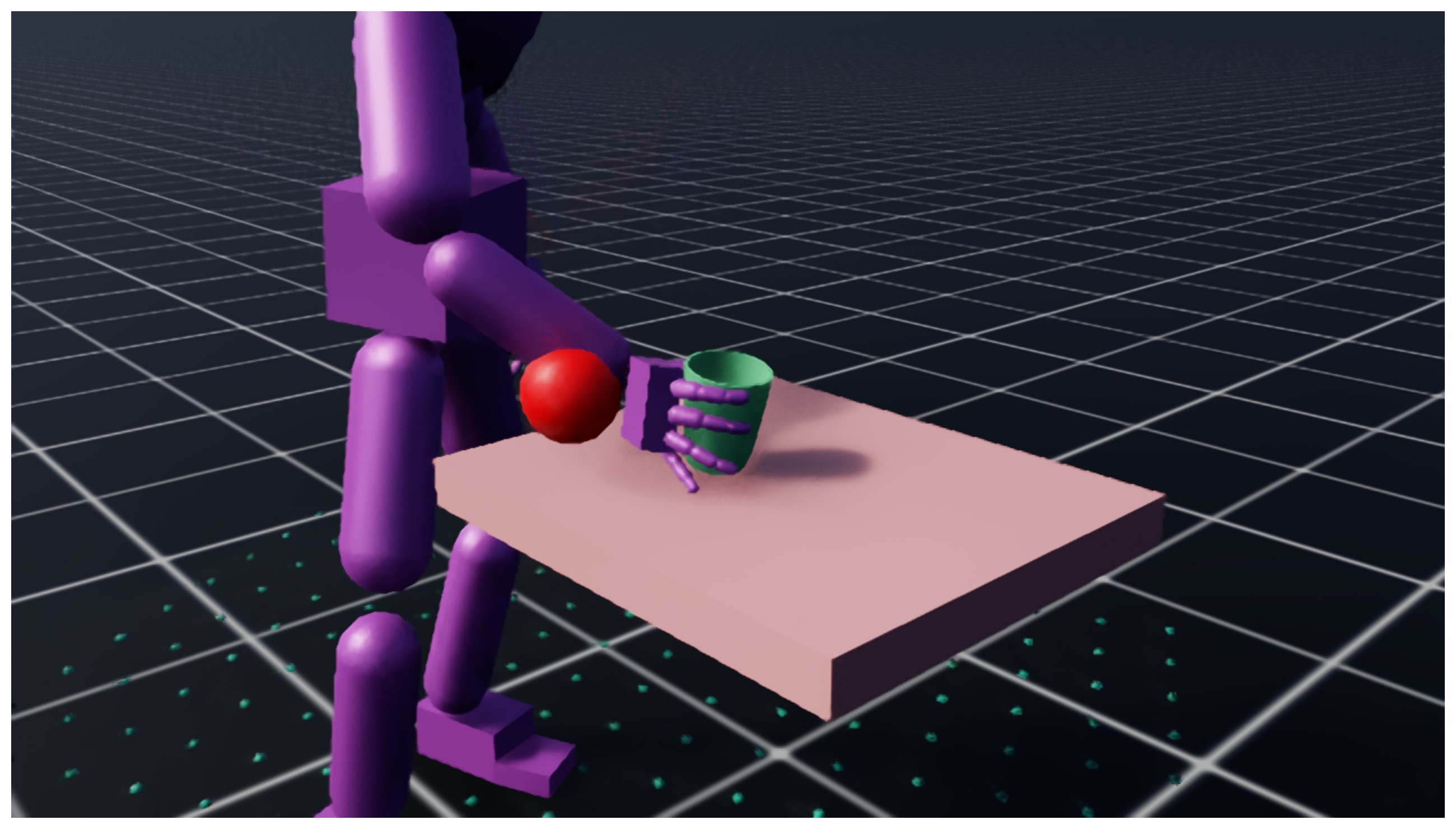}\hfill
         \includegraphics[trim={24cm 10cm 30cm 3.9cm},clip,width=0.33\textwidth]{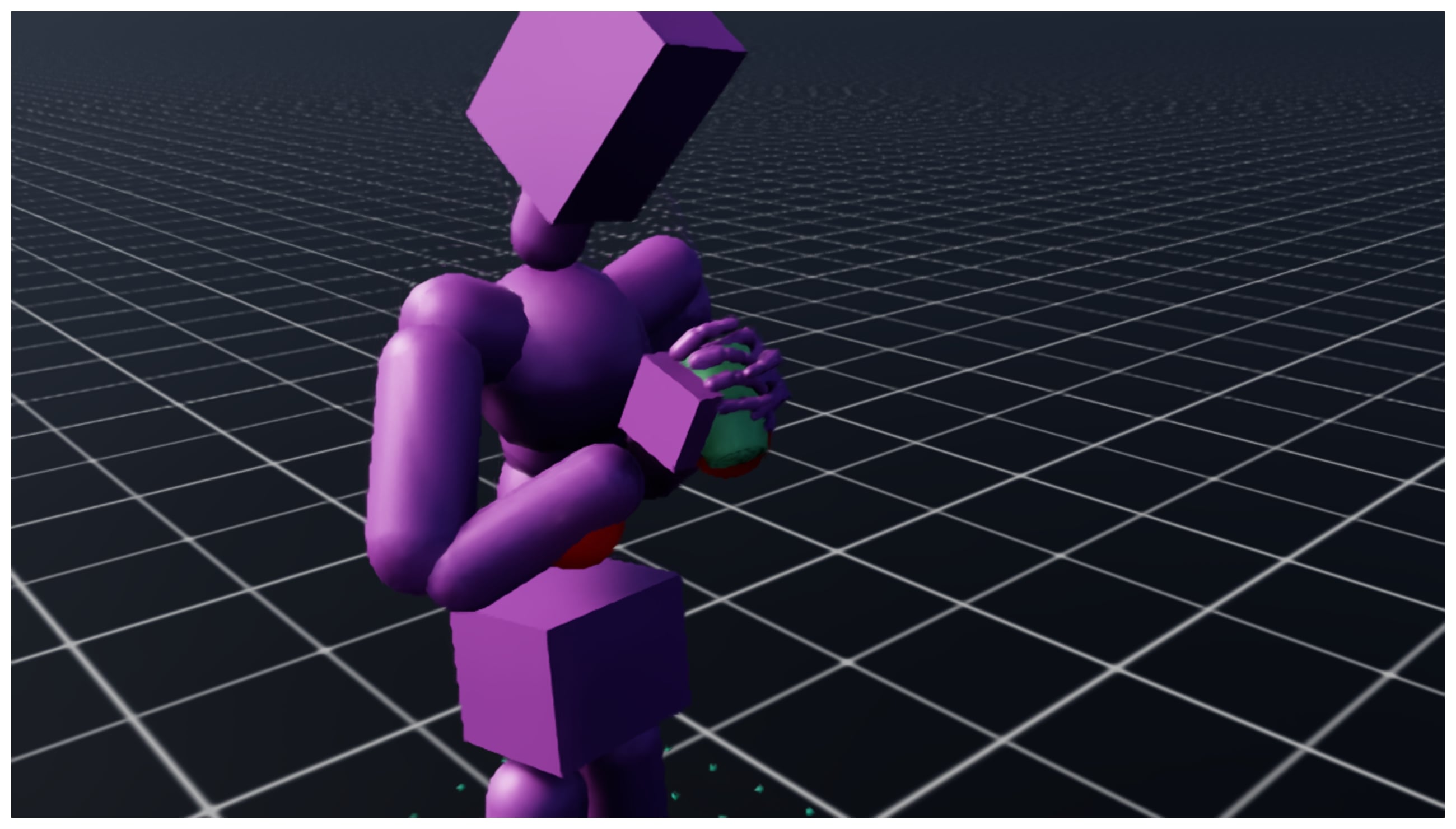}
         \caption{\textbf{Wrist control:} Conditioned only on the wrist position and orientation, the agent predicts when the motion is intended to interact with an object.}
         \label{fig: wrists cup}
    \end{subfigure}\hfill
    \begin{subfigure}[b]{0.495\textwidth}
         \centering
         \includegraphics[trim={24cm 8cm 30cm 5.9cm},clip,width=0.33\textwidth]{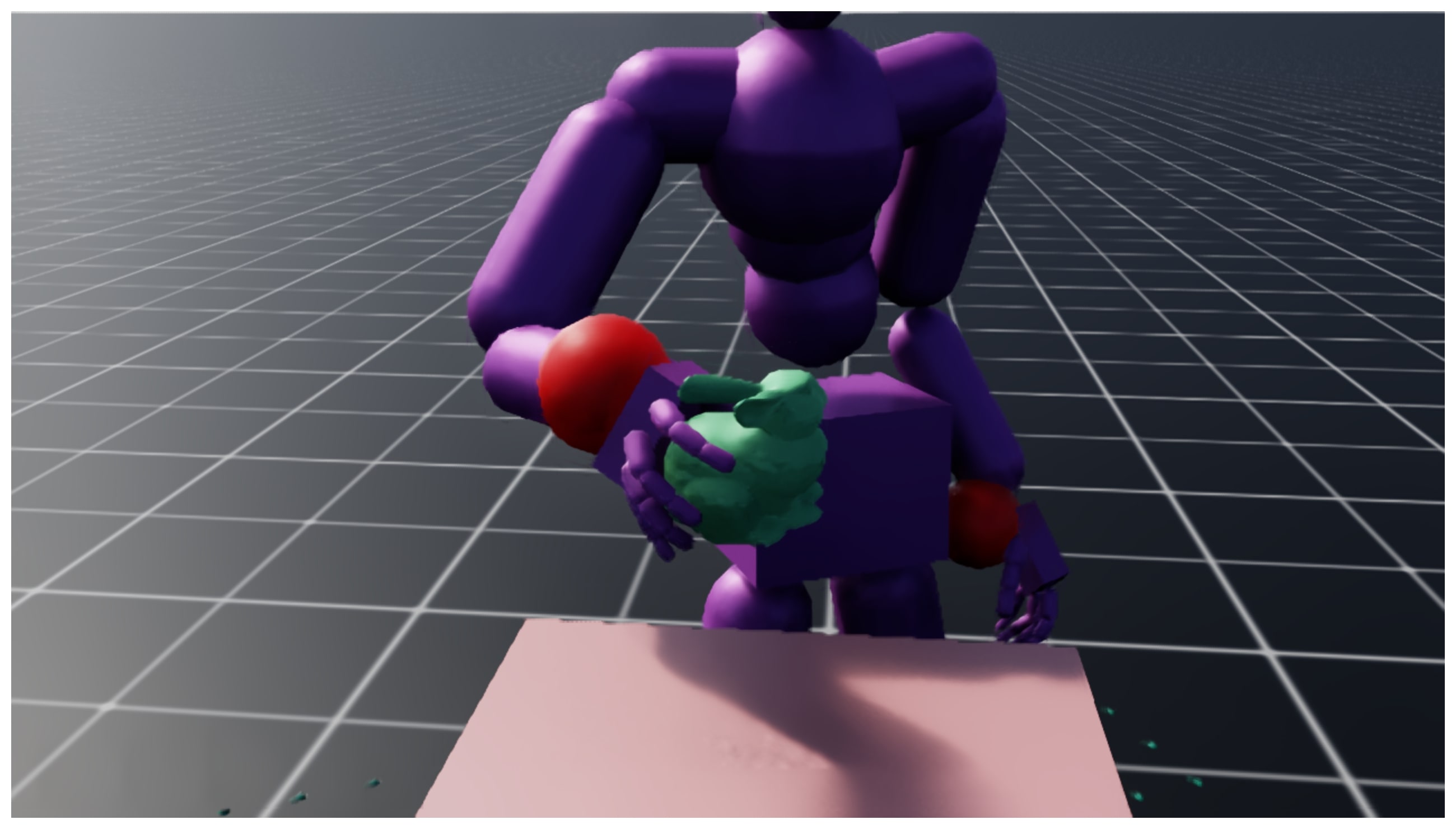}\hfill
         \includegraphics[trim={22cm 10cm 32cm 3.9cm},clip,width=0.33\textwidth]{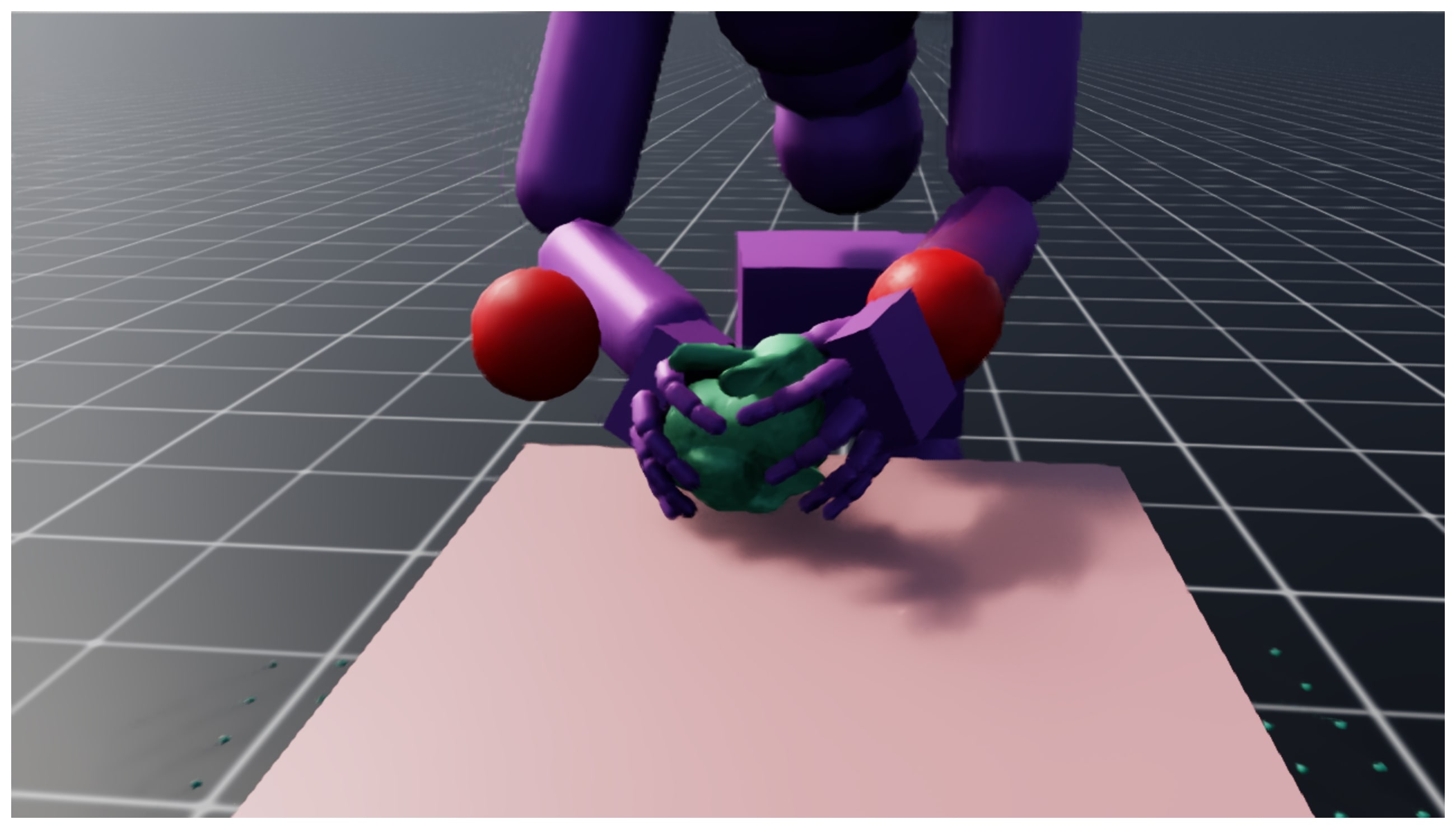}\hfill
         \includegraphics[trim={26cm 10cm 28cm 3.9cm},clip,width=0.33\textwidth]{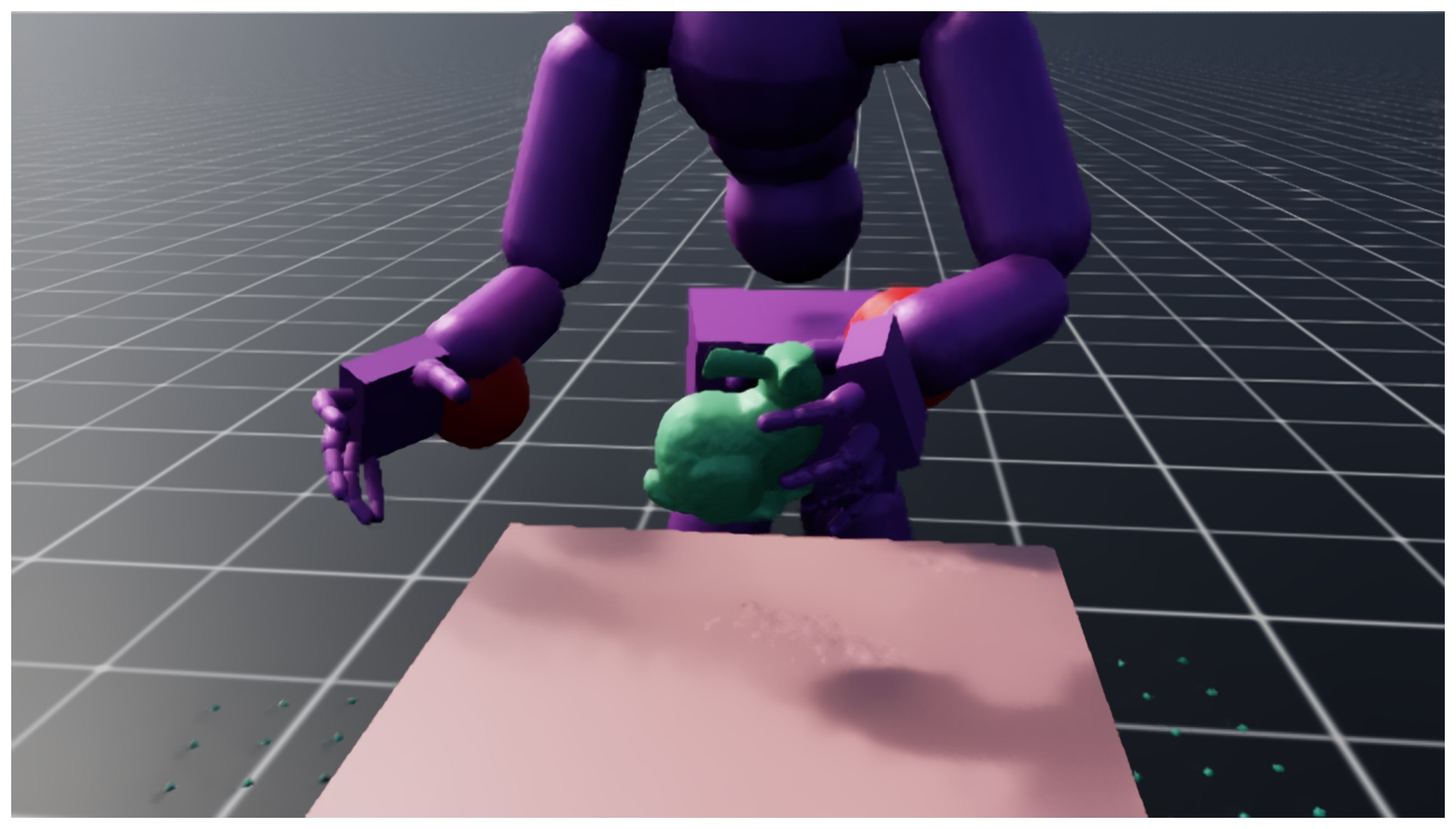}
         \caption{\textbf{Wrist control, hand-off:} When only conditioned on the wrists, the agent is required to `guess' which hand is more likely to maintain contact.}
         \label{fig: wrists bunny}
    \end{subfigure}\\
    \begin{subfigure}[b]{\textwidth}
         \centering
         \includegraphics[trim={24cm 13.9cm 30cm 0cm},clip,width=0.164\textwidth]{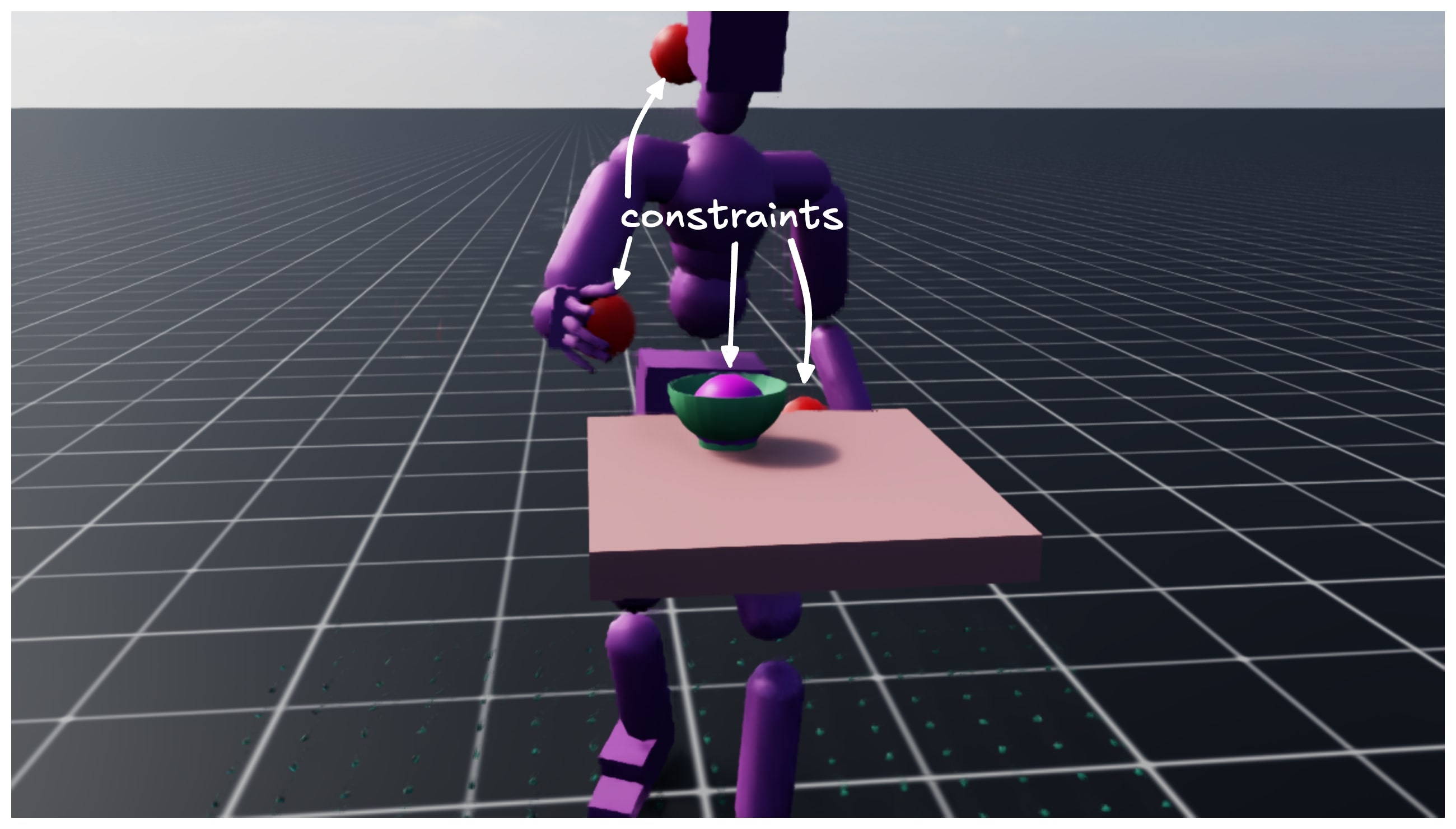}\hfill
         \includegraphics[trim={24cm 10cm 30cm 3.9cm},clip,width=0.164\textwidth]{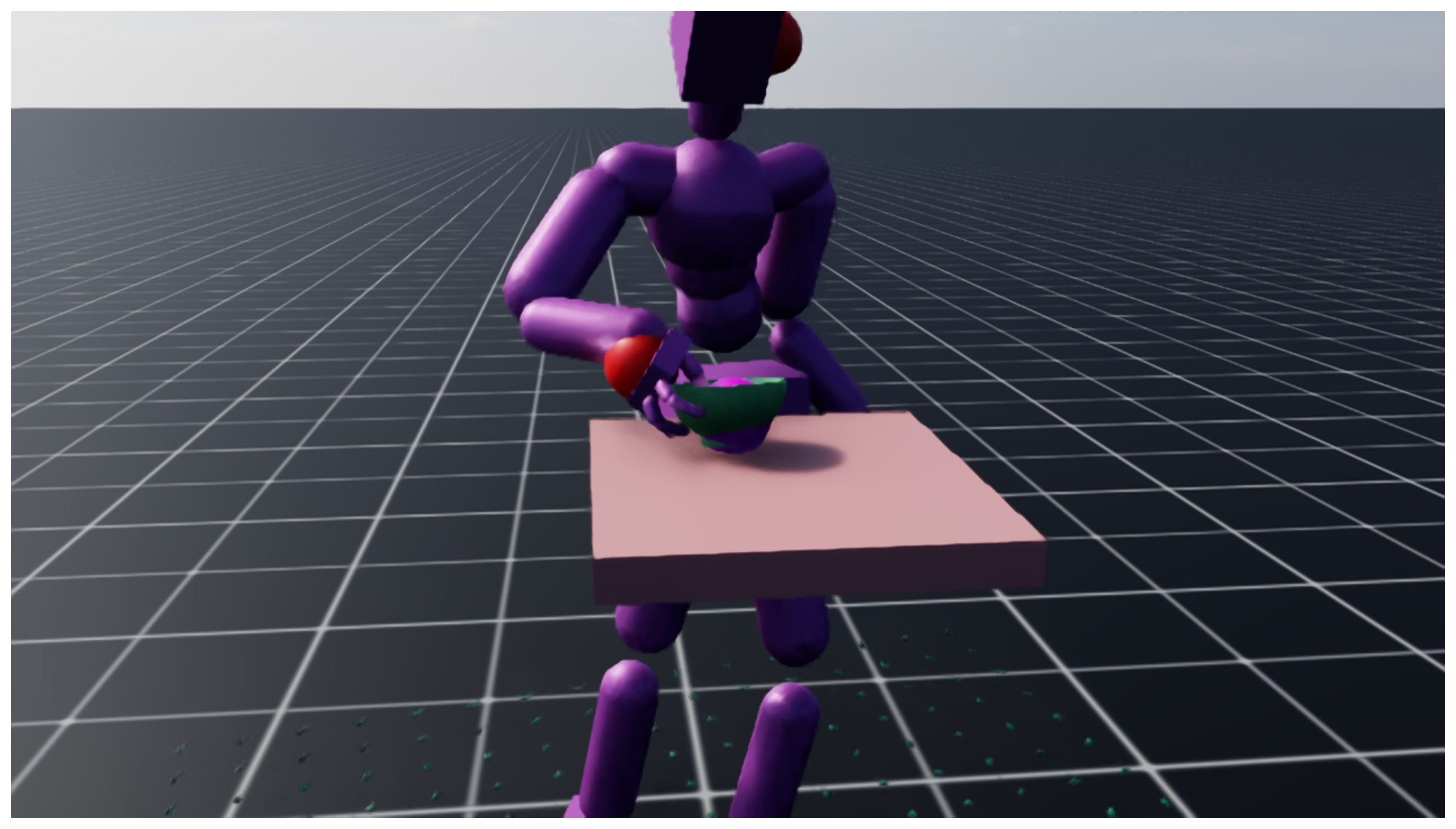}\hfill
         \includegraphics[trim={24cm 5cm 30cm 8.9cm},clip,width=0.164\textwidth]{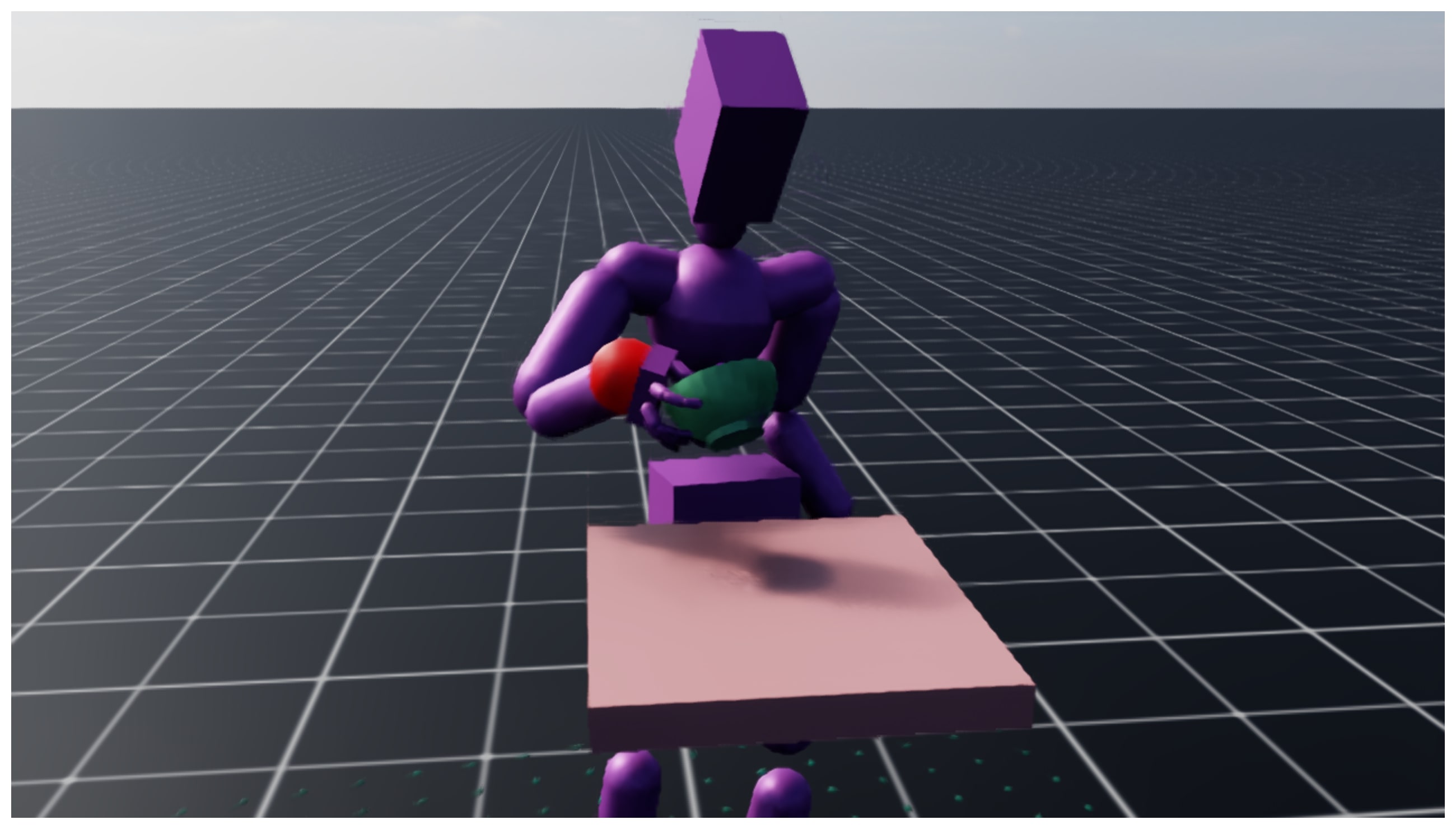}\hfill
         \includegraphics[trim={26cm 10cm 28cm 3.9cm},clip,width=0.164\textwidth]{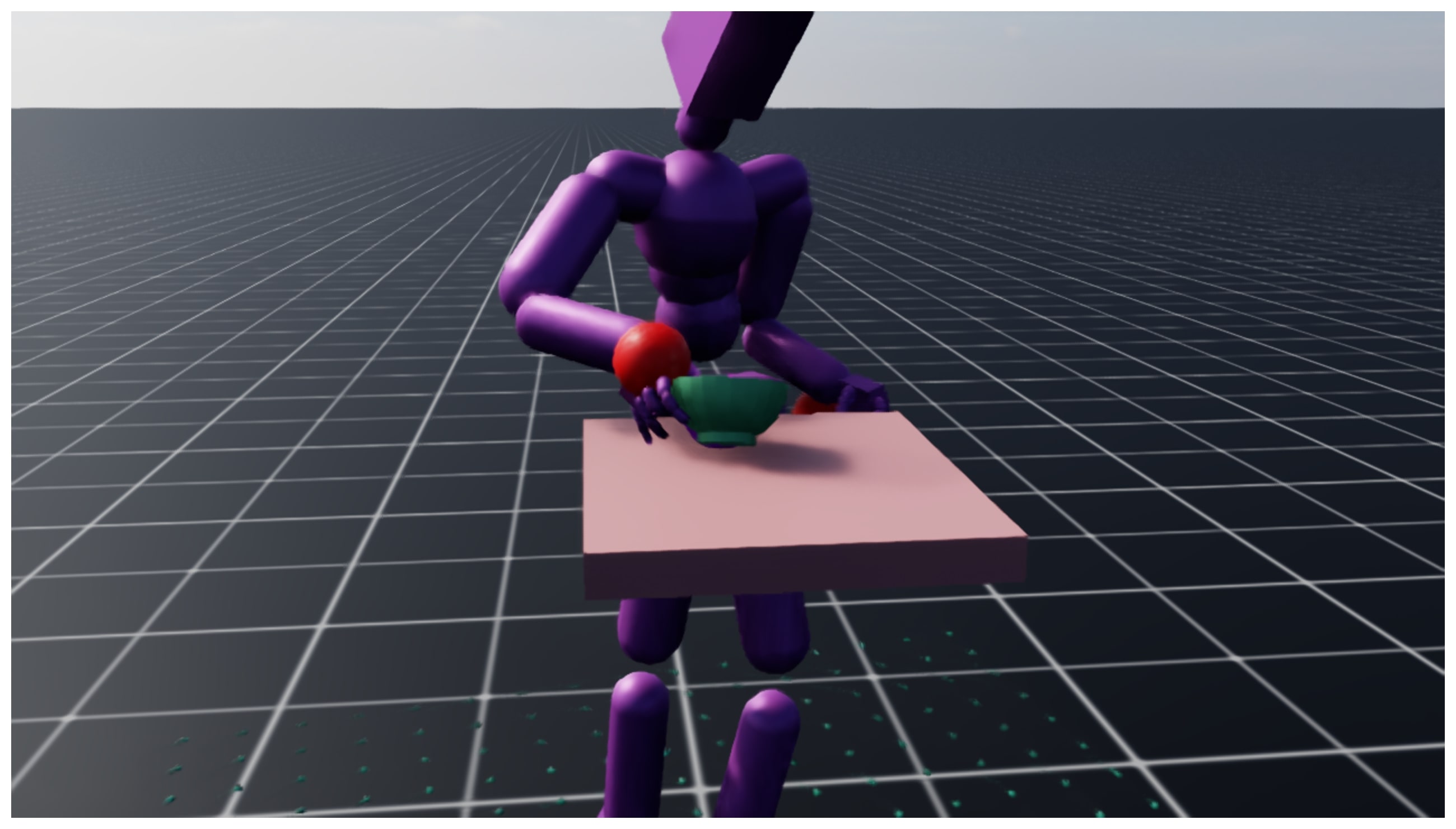}\hfill
         \includegraphics[trim={24cm 10cm 30cm 3.9cm},clip,width=0.164\textwidth]{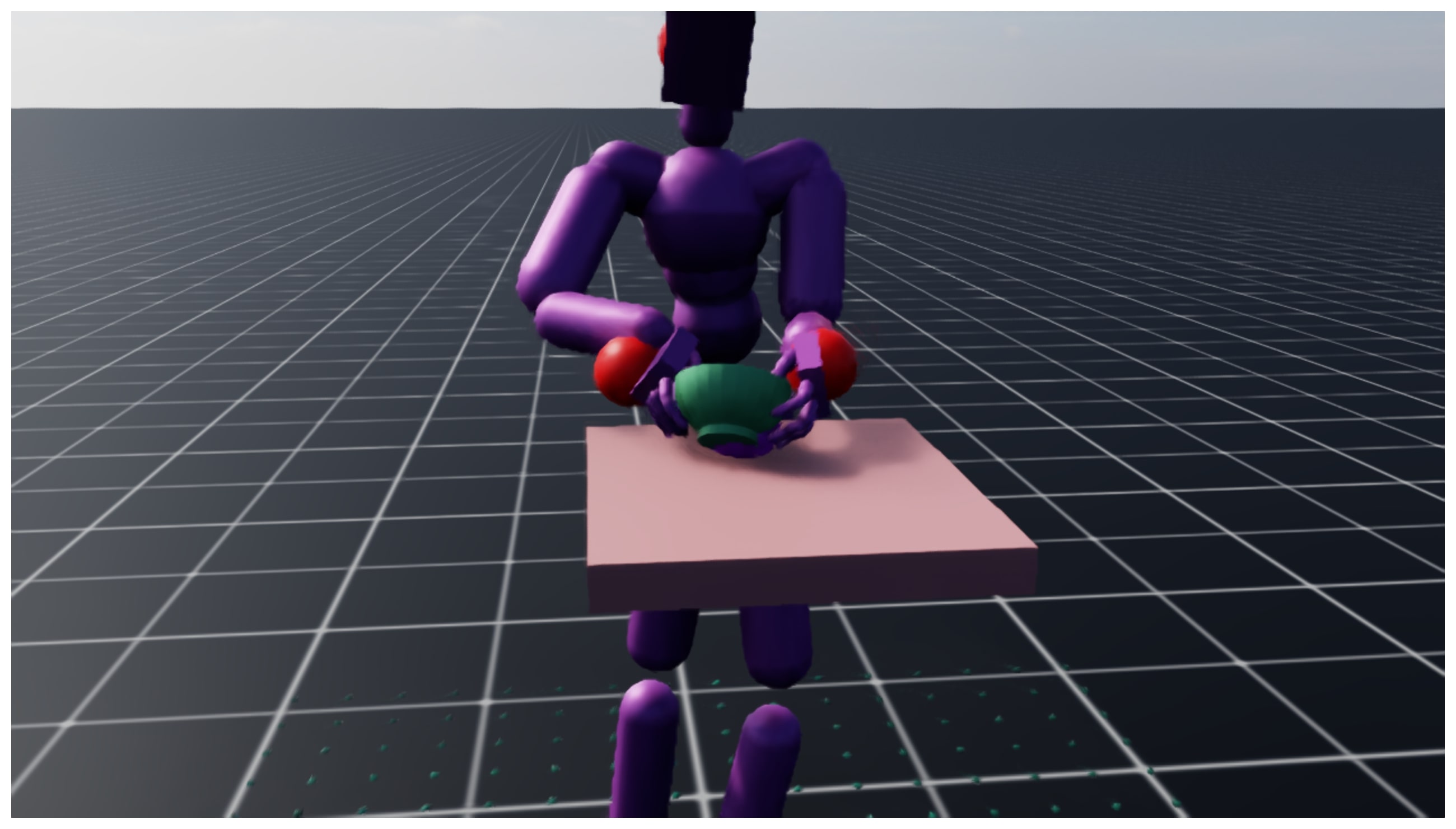}\hfill
         \includegraphics[trim={22cm 10cm 32cm 3.9cm},clip,width=0.164\textwidth]{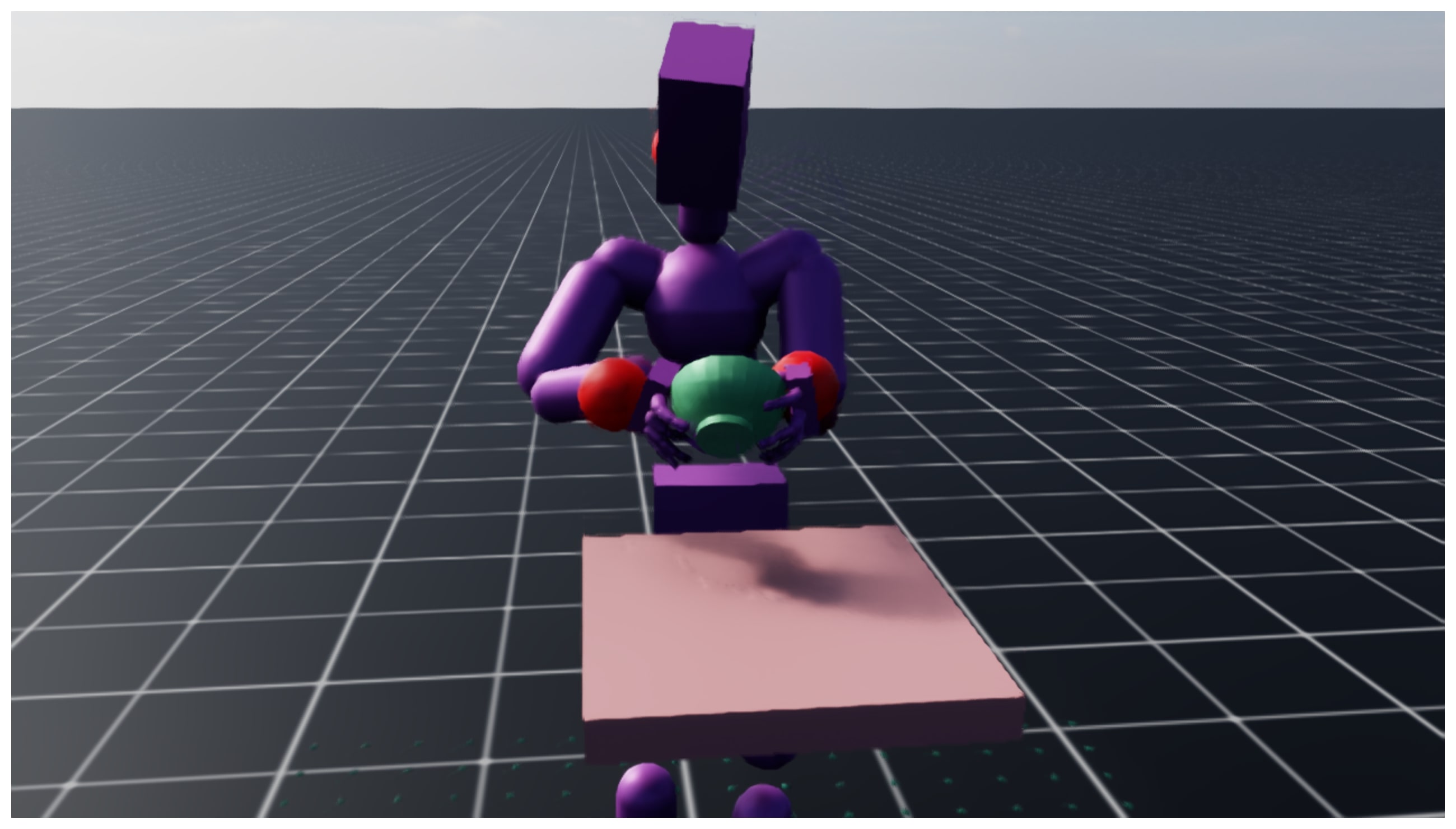}
         \caption{\textbf{Simulating teleoperation with object control:} Conditioning on the head position provides additional information on the body position, while conditioning on the object pose helps the policy infer how the hands should interact with the object.}
         \label{fig: wrists head object}
    \end{subfigure}
    \caption{\textbf{\generator\ -- wrists, head, and/or object conditioning:} \generator\ is conditioned on when and where to transport the object. The red spheres indicate target joint positions, whereas the purple where indicates the target object position.}
    \label{fig: teleop}
\end{figure*}

\begin{table*}[]
    \centering
    \caption{\textbf{\tracker:} We ablate various design decisions, stripping another component and measuring the cumulative importance. E.g., the ``Contact guidance'' is without ``Tight termination'' and ``Prioritized scenes''. We also compare with our implementation of InterMimic \cite{xu2025intermimic} (see \cref{subsec: tracker}).}
    \small
    \tabcolsep=0.1cm
    \begin{tabular}{l|cccc|cccc}
                            & \multicolumn{4}{c|}{\textbf{Train}} & \multicolumn{4}{c}{\textbf{Test}} \\ %\hline
                            & \thead{Sequence\\Success} & \thead{First Sequence\\Success} & MPJPE [mm] & \thead{Tracking\\Length} [s] & \thead{Sequence\\Success} & \thead{First Sequence\\Success} & MPJPE [mm] & \thead{Tracking\\Length} [s] \\ \hline\hline

        \textbf{\tracker} (ours)       & \cellcolor{green!25} 80.7\% & \cellcolor{green!25} 93.5\% & \cellcolor{green!25} 9.8 & \cellcolor{green!25} 9.6 & \cellcolor{green!25} 60.2\% & \cellcolor{green!25} 83.7\% & \cellcolor{green!25} 13.2 & \cellcolor{green!25} 7.3 \\ \hline

        (-) Tight termination      & 74.6\% & 91.2\% & 15.3 & 9.3 & 51.8\% & 80.1\% & 22.8 & 7 \\ \hline

        \quad(-) Prioritized scenes      & 71.2\% & 89.3\% & 15.5 & 9.2 & 56\% & 73.8\% & 18.4 & 7 \\ \hline

        \quad\quad(-) Contact guidance      & 69.4\% & 86.6\% & 16.4 & 9.2 & 47.5\% & 78\% & 25.5 & 6.5 \\ \hline

        InterMimic* \cite{xu2025intermimic}      & 11\% & 31.1\% & 42.2 & 6.2 & 8.5\% & 29.8\% & 50.5 & 4.9 
    \end{tabular}
    \label{tab: grab tracker}\vspace{-0.2cm}
\end{table*}

\subsection{\tracker}\label{subsec: tracker}

We evaluate ManipulationMimic's performance through ablation studies and quantitative comparisons, and provide qualitative examples of its capabilities.

\paragraph{Ablation study and quantitative analysis} \cref{tab: grab tracker} presents \tracker's performance on the GRAB dataset and the impact of key design choices. We ablated: (1) tight early termination criteria, (2) prioritized training on harder motions, and (3) the phased contact guidance reward. The results demonstrate that each component positively contributes to the overall tracking success, with the strict early termination yielding the most significant improvement by maintaining a tight feasibility envelope.

We also compare \tracker\ with our implementation of InterMimic \cite{xu2025intermimic}. For a fair comparison, we trained a single InterMimic model on the entire GRAB training set, similar to \tracker, omitting InterMimic's subject-specific policy distillation stages. In addition, we did not include physical state initialization as it introduced instabilities in dexterous manipulation tasks. \tracker\ outperforms this adapted InterMimic baseline. We attribute InterMimic's lower performance in this setting to its potentially looser tracking requirements and a lack of explicit object structure awareness (e.g., a BPS representation for contact guidance) which is crucial for precise interactions with diverse objects in GRAB.

\paragraph{Qualitative analysis} As illustrated in \cref{fig: full body tracker} and the supplementary video, \tracker\ successfully reconstructs a diverse range of complex, full-body interactions while preserving natural, human-like motion qualities. For instance, the policy can accurately simulate grasping a teapot by its handle to pour liquid, or picking up a hammer and striking a nail. These examples highlight its ability to learn nuanced, tool-specific behaviors. While \tracker\ reliably reconstructs the majority of motions, occasional uncanny behaviors can be observed, particularly during the precise moments of contact initiation or release. These may stem from the discrete nature of the termination conditions related to contact, which can sometimes force abrupt transitions rather than perfectly smooth engagement or disengagement.

\subsection{\generator}

We quantitatively evaluate \generator\ ($\pi_{\text{versatile}}$), assessing its versatility, generalization, and long-horizon reasoning.

\paragraph{Quantitative analysis.}

\textbf{Teleoperation-Style Pose Matching:} \cref{tab: teleop} compares the proposed \generator\ architectures (Deterministic, C-VAE, Diffusion) on a teleoperation-like task. Here, the policy is conditioned on near-future goals for head and wrist poses, plus the object's pose, aiming to accurately achieve these combined configurations. While training performance was similar for the stochastic C-VAE and Diffusion models, the Diffusion policy generalized better, successfully completing $3.6\%$ more unseen test sequences than the other two. This highlights the advantage of its capacity to model complex, multi-modal action distributions for robust generalization. In addition, we observe a dramatic performance drop when training on offline data, highlighting the importance of self-play for learning to recover from mistakes \cite{ross2011reduction}.

\textbf{Long-Horizon Sparse Object Goal Chaining:} We further test the policies on a challenging long-horizon task (\cref{tab: object goals}) where \generator\ receives a sparse object pose goal 2 seconds into the future. Upon reaching the target time, a new 2-second future object goal is set, repeating until the reference motion ends. This evaluates the policy's ability to reason about distant object states while producing appropriate full-body motion and stable grasps for diverse objects. Performance is measured by the success rate in achieving the sequence of object goals. Consistent with the teleoperation task and prior work \cite{tessler2024maskedmimic}, modeling solution diversity proved crucial. While the C-VAE performed better on the training set (indicating some overfitting), the Diffusion model again exhibited superior generalization, successfully solving $9.3\%$ more unseen test sequences than the C-VAE.

\begin{table}[]
    \centering
    \caption{\textbf{`Teleoperation':} We compare the various architectures on a teleoperation task. Here, \generator\ is conditioned on the positions and rotations of the head, wrists, and object.}
    \small
    \tabcolsep=0.1cm
    \begin{tabular}{l|cc|cc}
                            & \multicolumn{2}{c|}{\textbf{Train}} & \multicolumn{2}{c}{\textbf{Test}} \\ %\hline
                            & Success & MPJPE [mm] & Success & MPJPE [mm] \\ \hline\hline

        Deterministic       & 73.8\% & 16.4 & 54.6\% & 24 \\ \hline
        C-VAE       & 78.5\% & 12.7 & 54.6\% & 24.4 \\ \hline
        Offline Diffusion       & 50\% & 27.8 & 25.5\% & 38.5 \\ \hline
        Diffusion       & \cellcolor{green!25} 78.6\% & \cellcolor{green!25} 12.1 & \cellcolor{green!25} 58.2\% & \cellcolor{green!25} 19.7
    \end{tabular}
    \label{tab: teleop}\vspace{-0.2cm}
\end{table}

\begin{table}[]
    \centering
    \caption{\textbf{Object goals:} The agent is provided a sparse objective indicating where (and when) it should transport the object to.}
    \small
    \tabcolsep=0.1cm
    \begin{tabular}{l|cc|cc}
                            & \multicolumn{2}{c|}{\textbf{Train}} & \multicolumn{2}{c}{\textbf{Test}} \\ %\hline
                            & Success & \thead{Tracking\\Length} [s] & Success & \thead{Tracking\\Length} [s] \\ \hline\hline

        Deterministic       & 54.9\% & 7.8 & 46.1\% & 6.2 \\ \hline
        C-VAE       & 64.7\% & 8.3 & 50.3\% & 6.6 \\ \hline
        Offline Diffusion       & 30.9\% & 6.8 & 32.6\% & 5.8 \\ \hline
        Diffusion       & \cellcolor{green!25} 57.3\% & \cellcolor{green!25} 7.9 & \cellcolor{green!25} 59.6\% & \cellcolor{green!25} 6.8
    \end{tabular}
    \label{tab: object goals}\vspace{-0.2cm}
\end{table}

\begin{figure*}[t]
     \centering
     \begin{subfigure}[b]{\textwidth}
         \centering
         \includegraphics[trim={14cm 10cm 40cm 3.9cm},clip,width=0.164\textwidth]{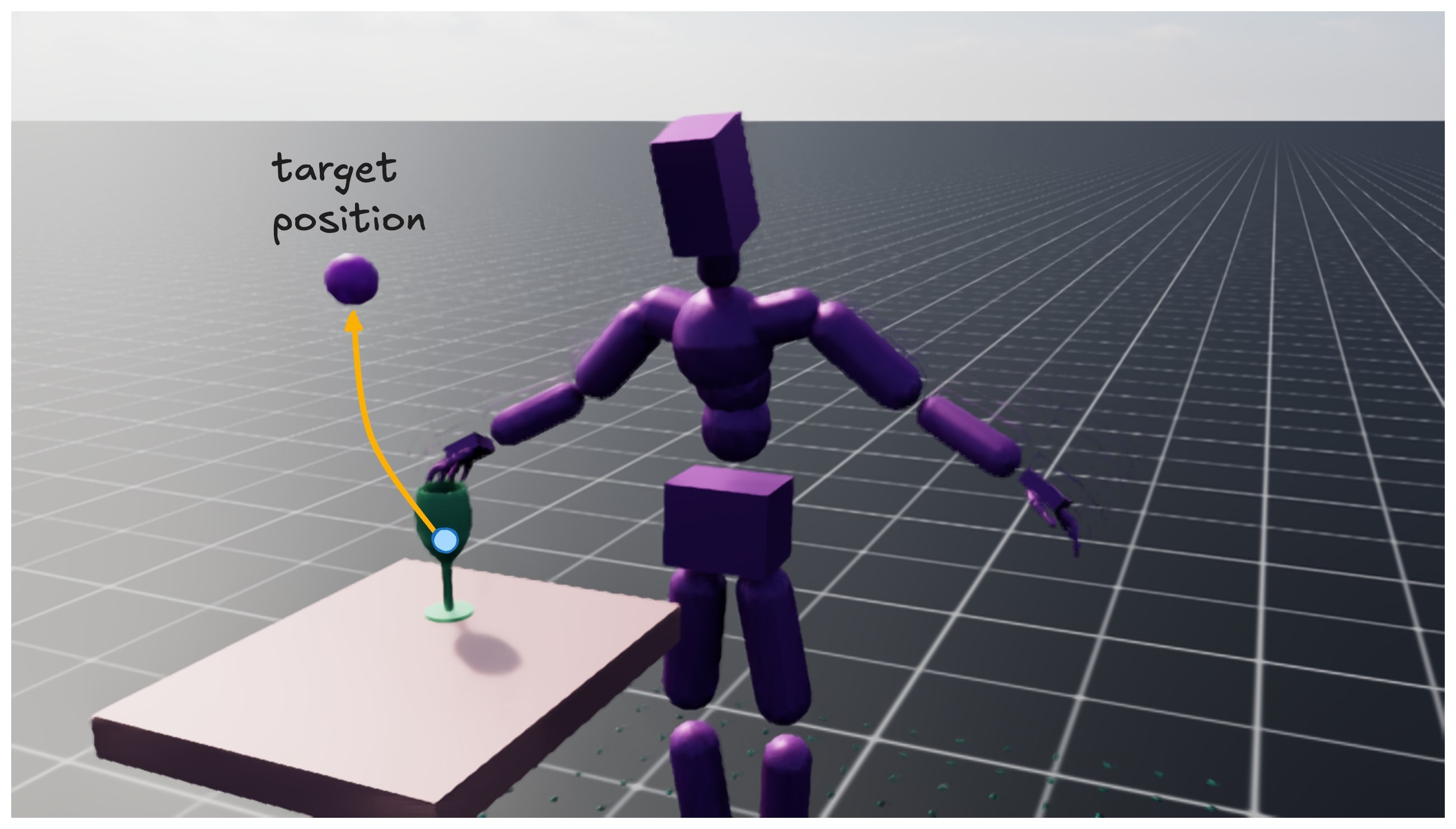}\hfill
         \includegraphics[trim={18cm 10cm 36cm 3.9cm},clip,width=0.164\textwidth]{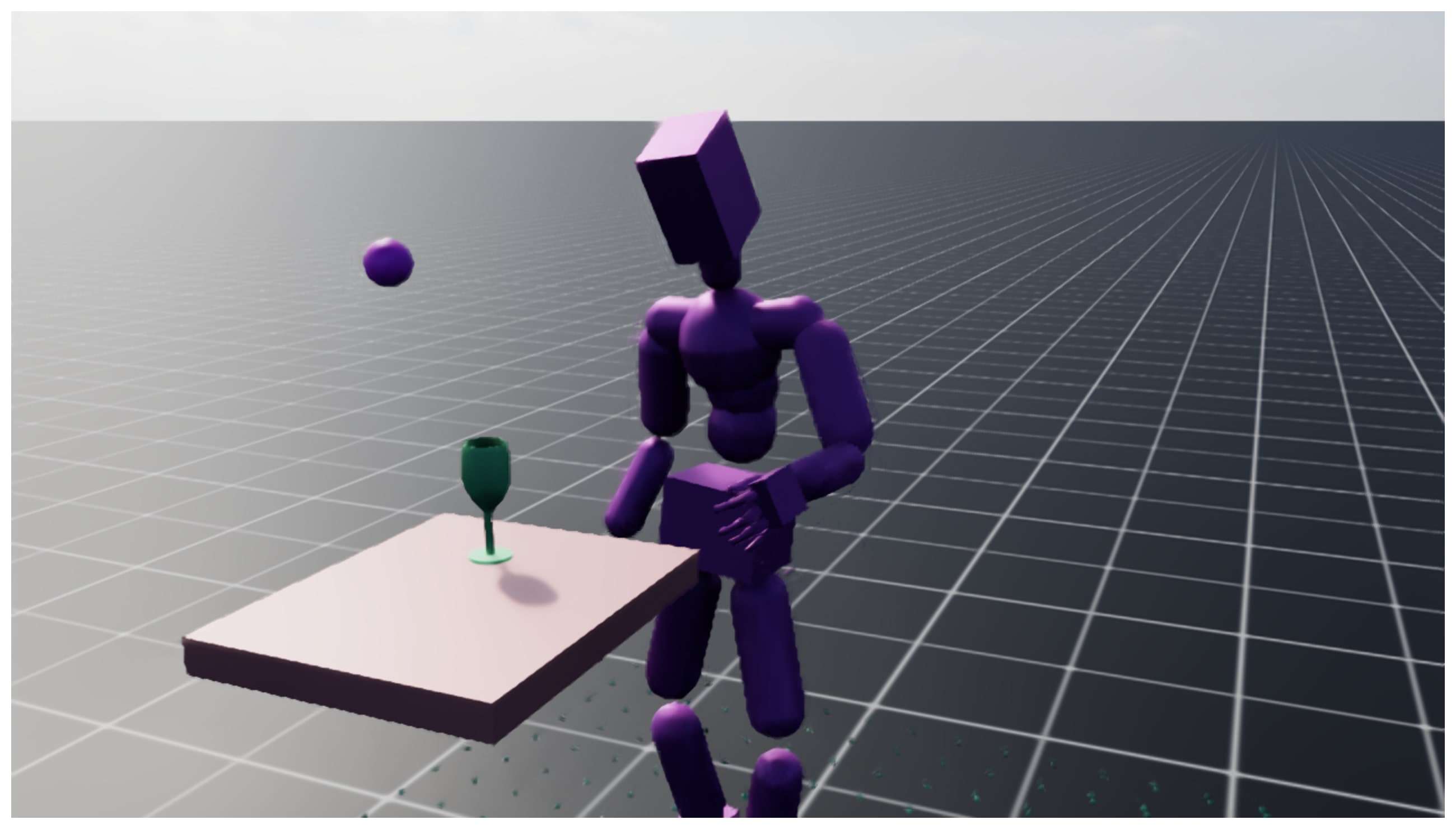}\hfill
         \includegraphics[trim={24cm 10cm 30cm 3.9cm},clip,width=0.164\textwidth]{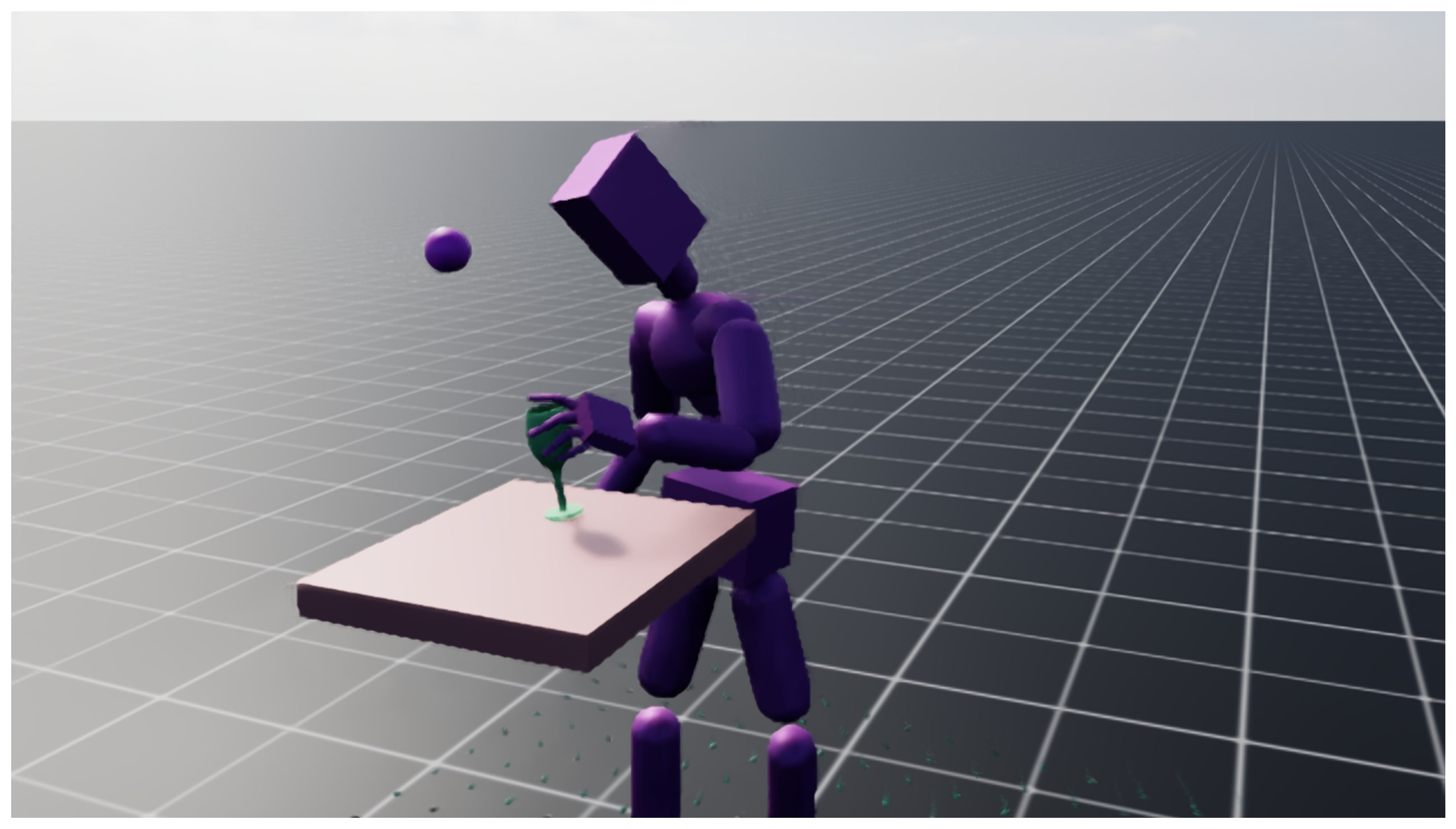}\hfill
         \includegraphics[trim={24cm 10cm 30cm 3.9cm},clip,width=0.164\textwidth]{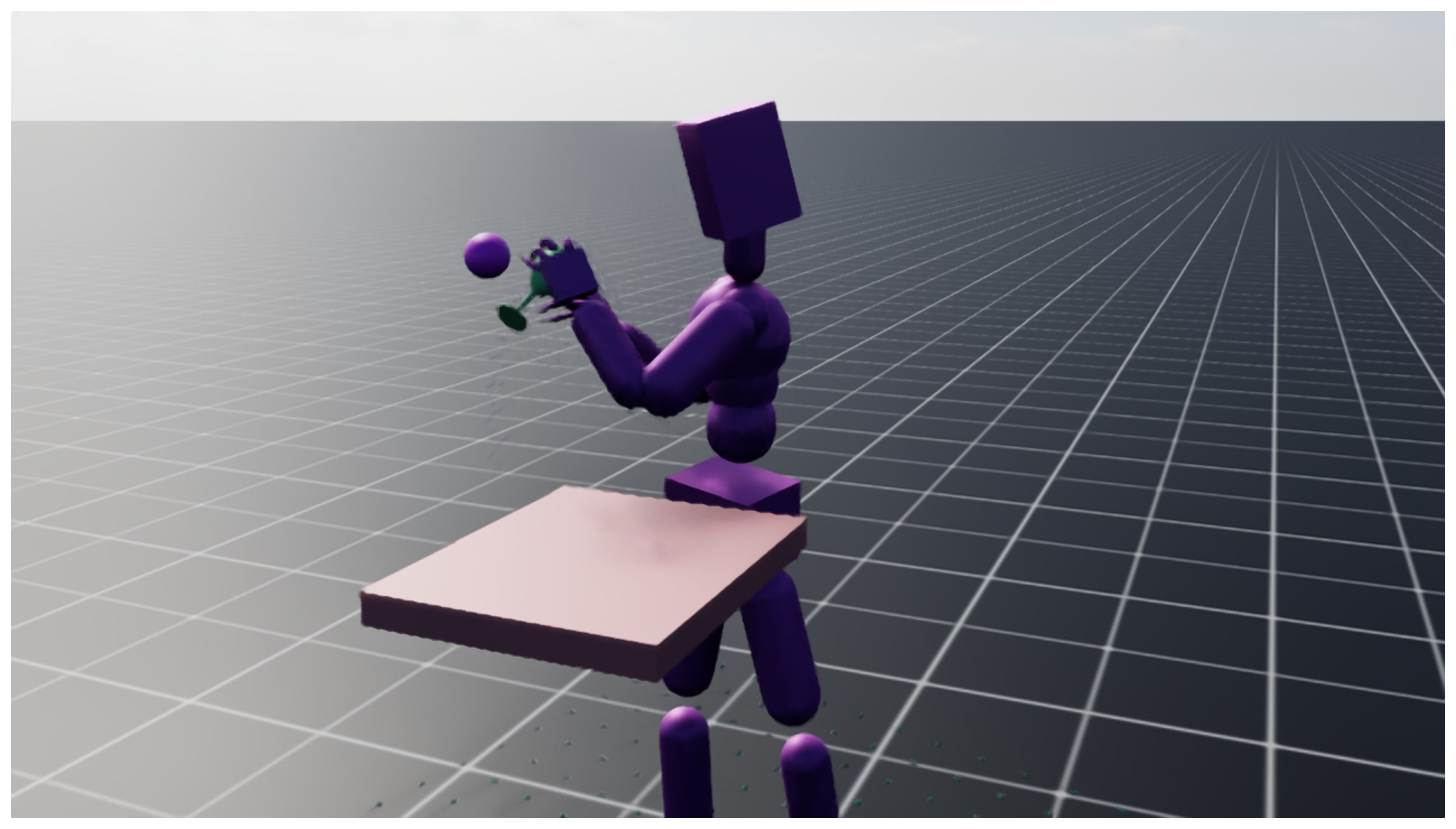}\hfill
         \includegraphics[trim={24cm 10cm 30cm 3.9cm},clip,width=0.164\textwidth]{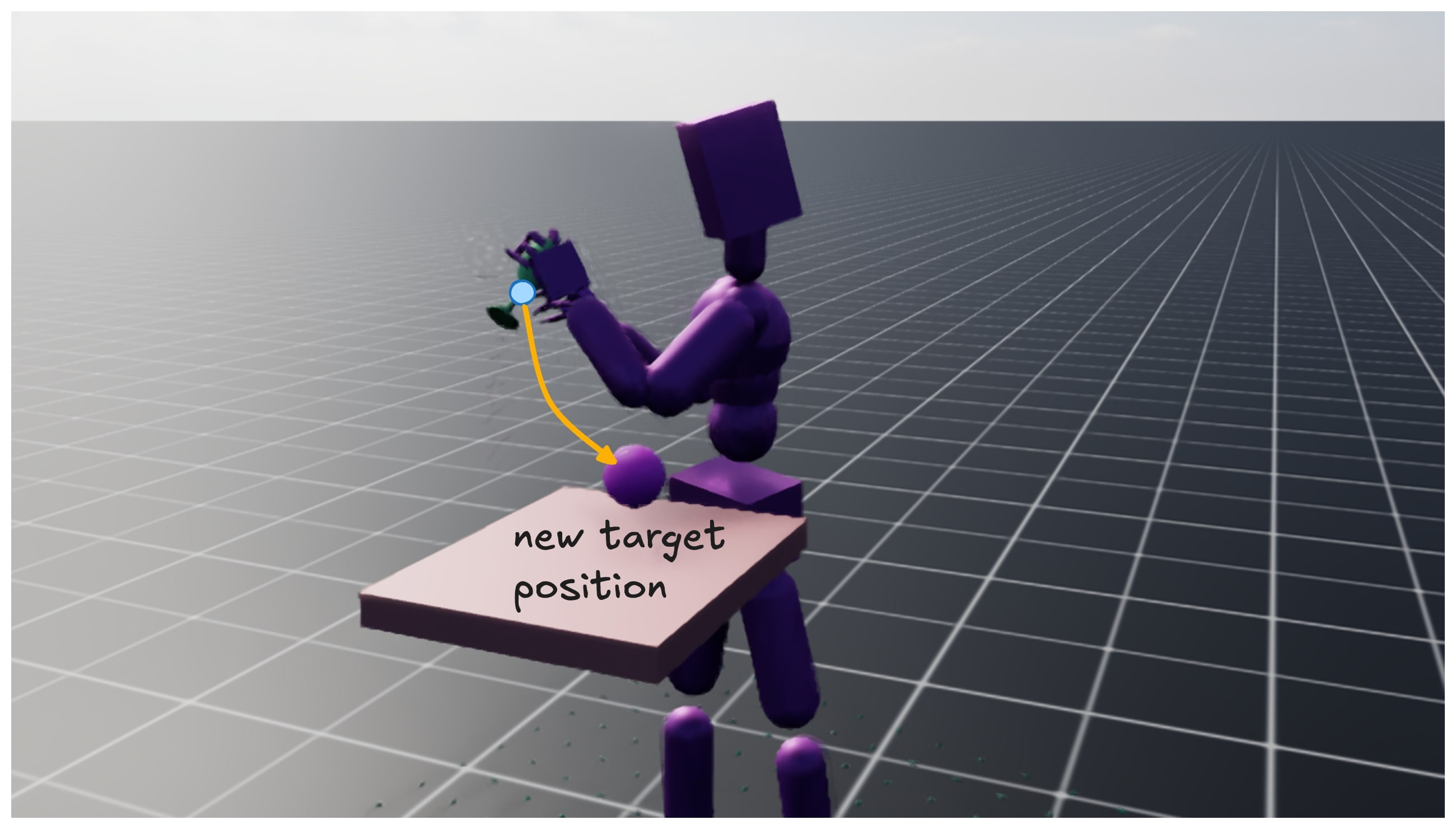}\hfill
         \includegraphics[trim={24cm 10cm 30cm 3.9cm},clip,width=0.164\textwidth]{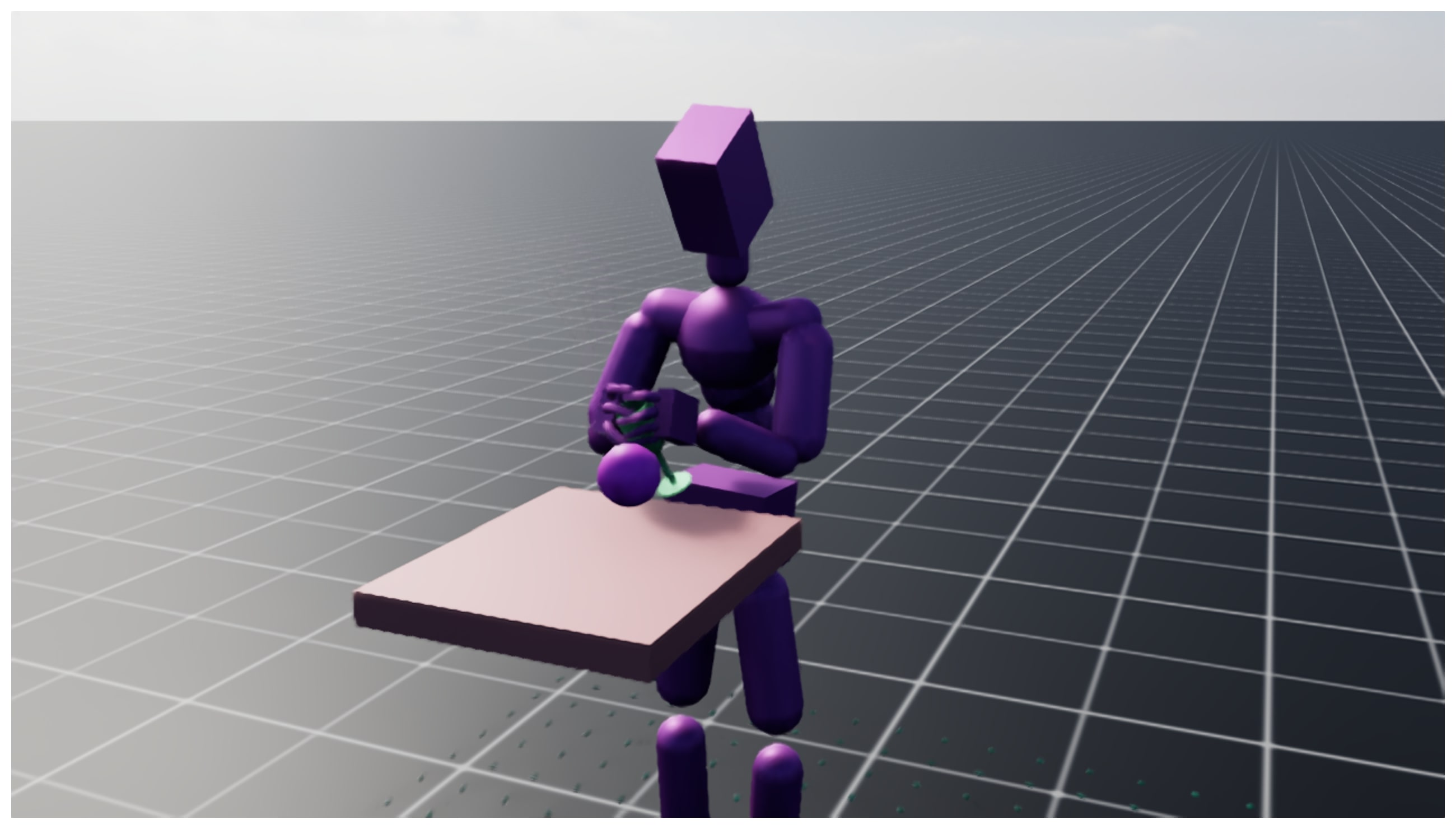}
         \caption{\textbf{Sequential goals:} The agent picks up and transports the object to the target position. It reacts to changes in the objective.}
         \label{fig: wine glass change goal}
    \end{subfigure}\\
    \begin{subfigure}[b]{\textwidth}
         \centering
         \includegraphics[trim={10cm 7cm 44cm 6.9cm},clip,width=0.164\textwidth]{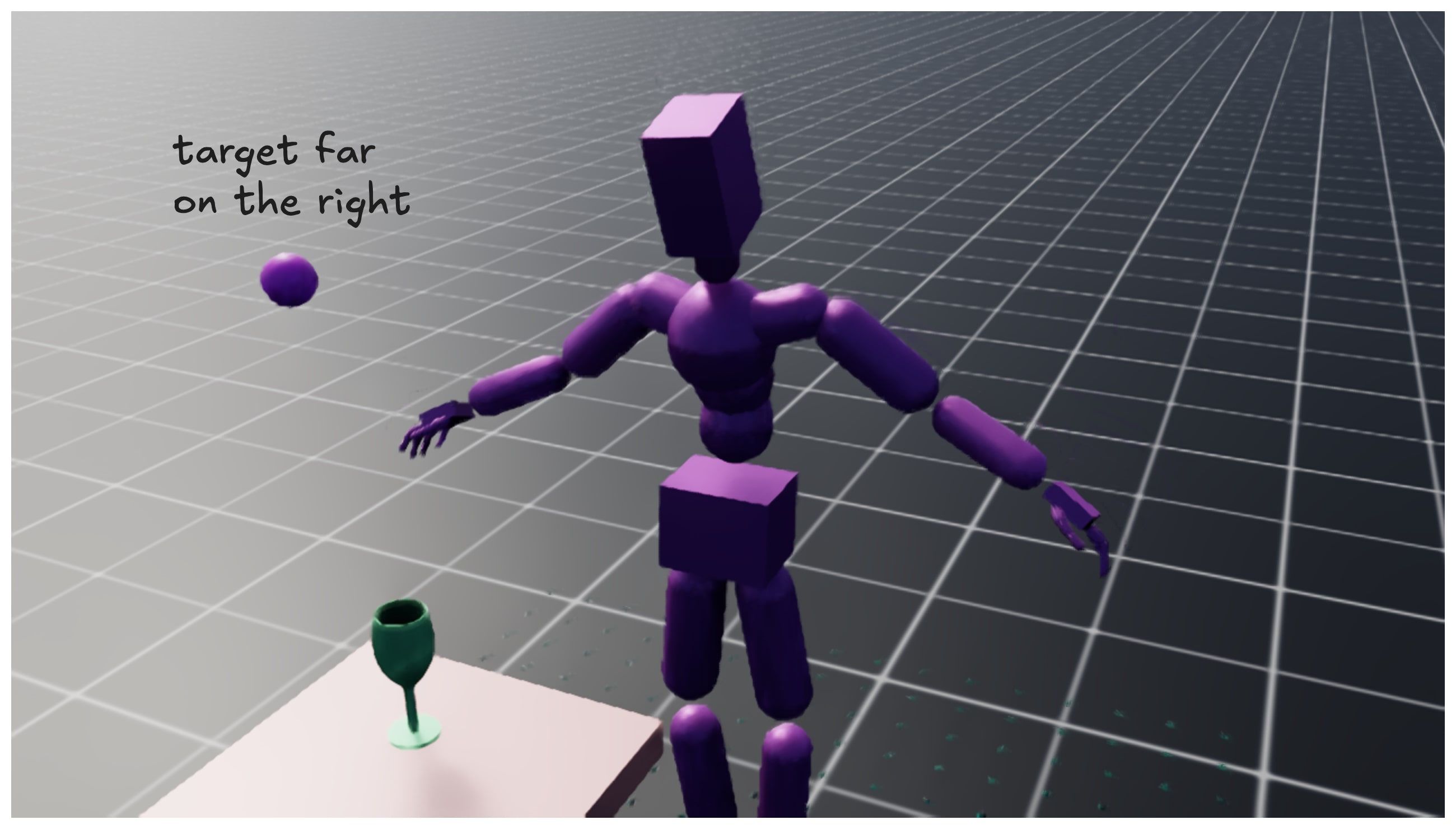}\hfill
         \includegraphics[trim={18cm 10cm 36cm 3.9cm},clip,width=0.164\textwidth]{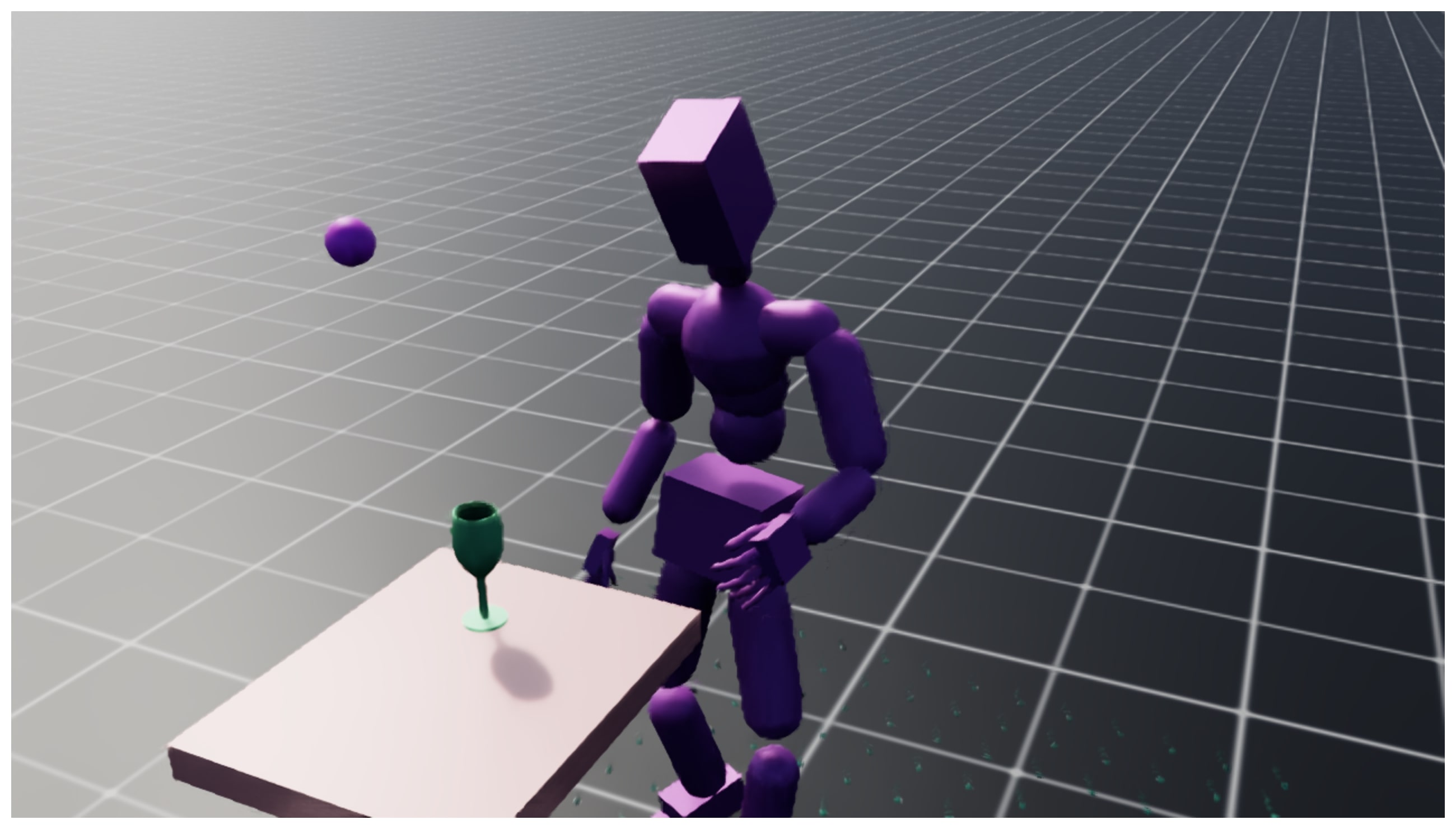}\hfill
         \includegraphics[trim={22cm 5cm 32cm 8.9cm},clip,width=0.164\textwidth]{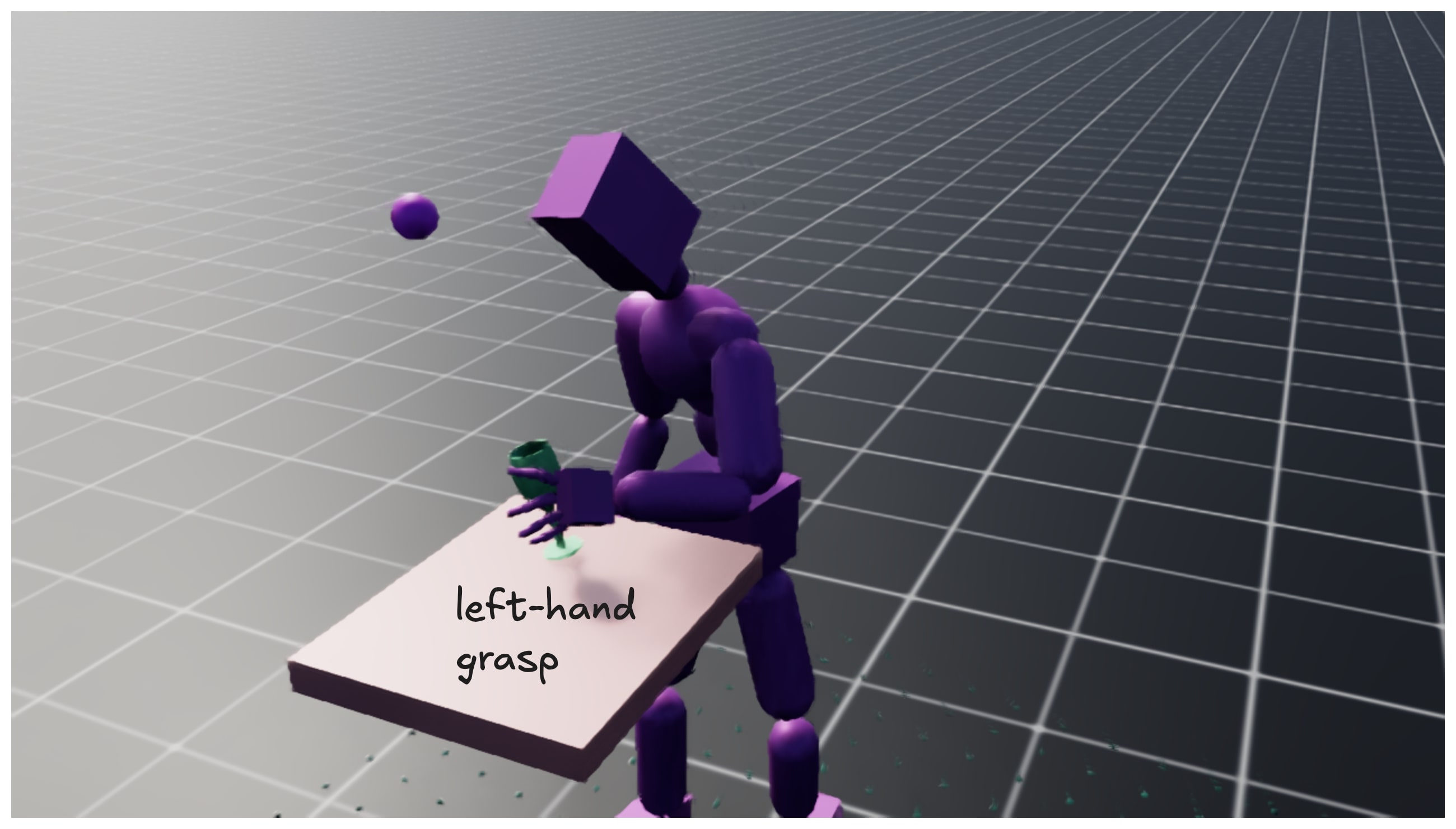}\hfill
         \includegraphics[trim={20cm 10cm 34cm 3.9cm},clip,width=0.164\textwidth]{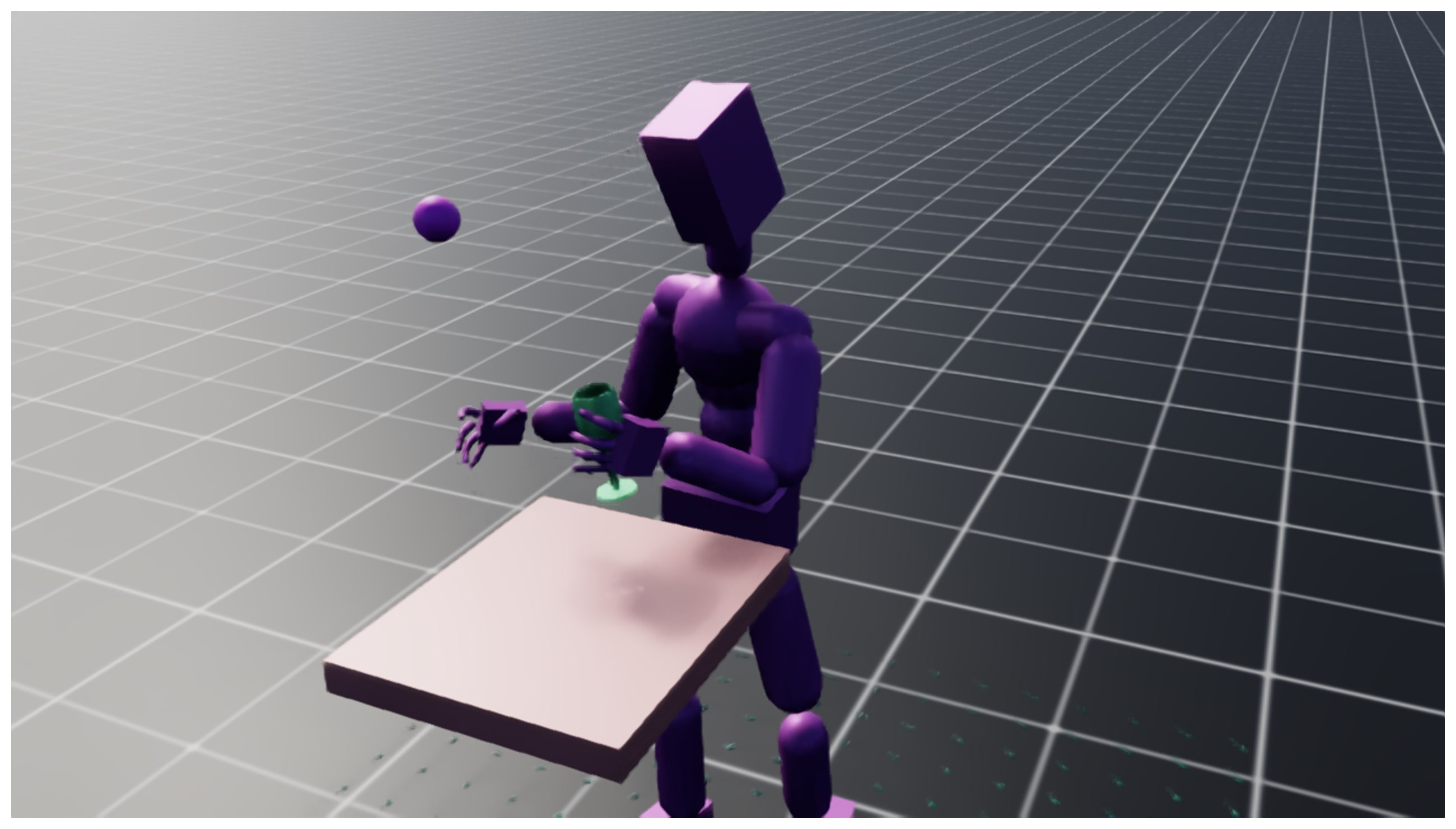}\hfill
         \includegraphics[trim={20cm 10cm 34cm 3.9cm},clip,width=0.164\textwidth]{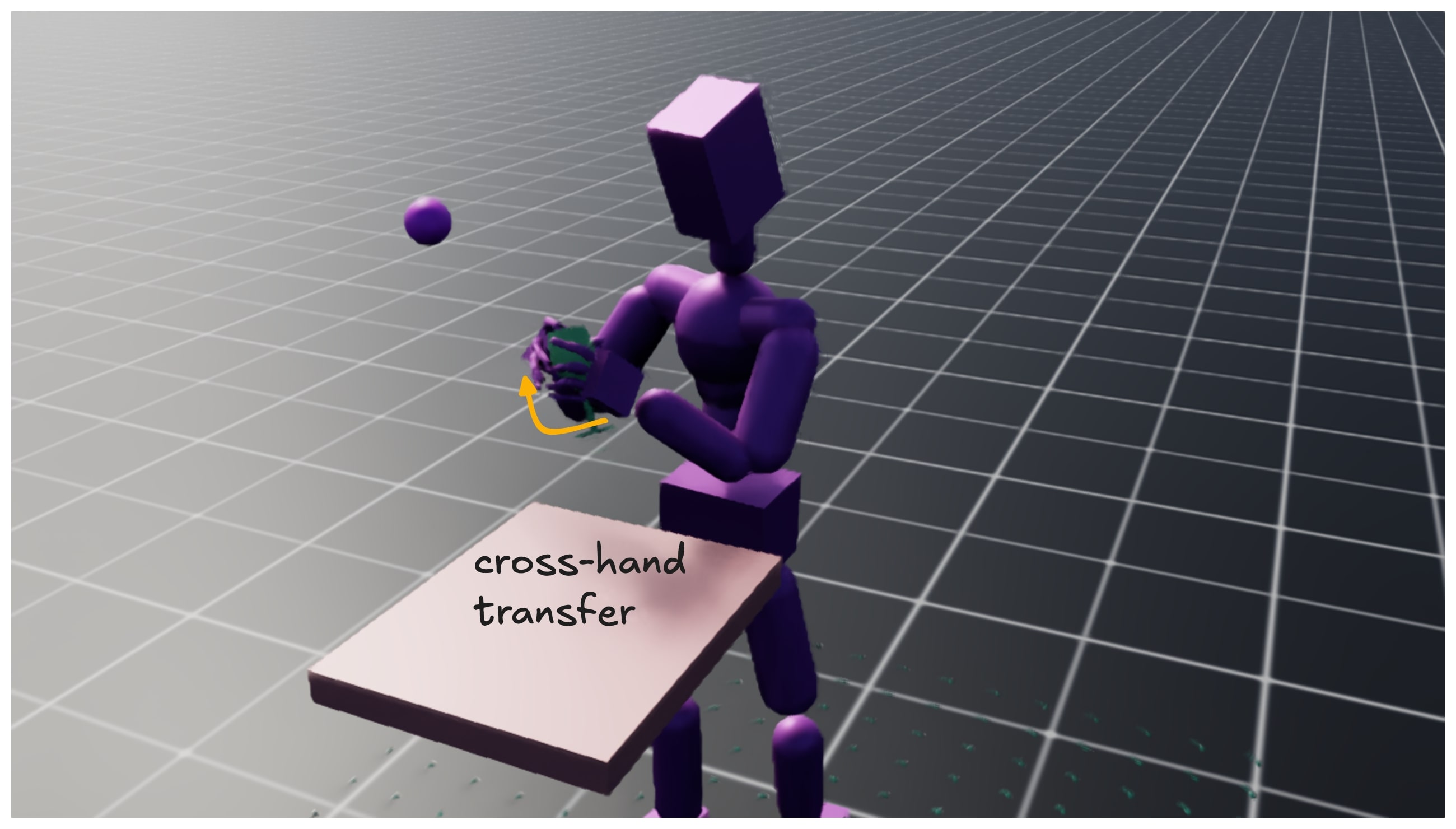}\hfill
         \includegraphics[trim={18cm 10cm 36cm 3.9cm},clip,width=0.164\textwidth]{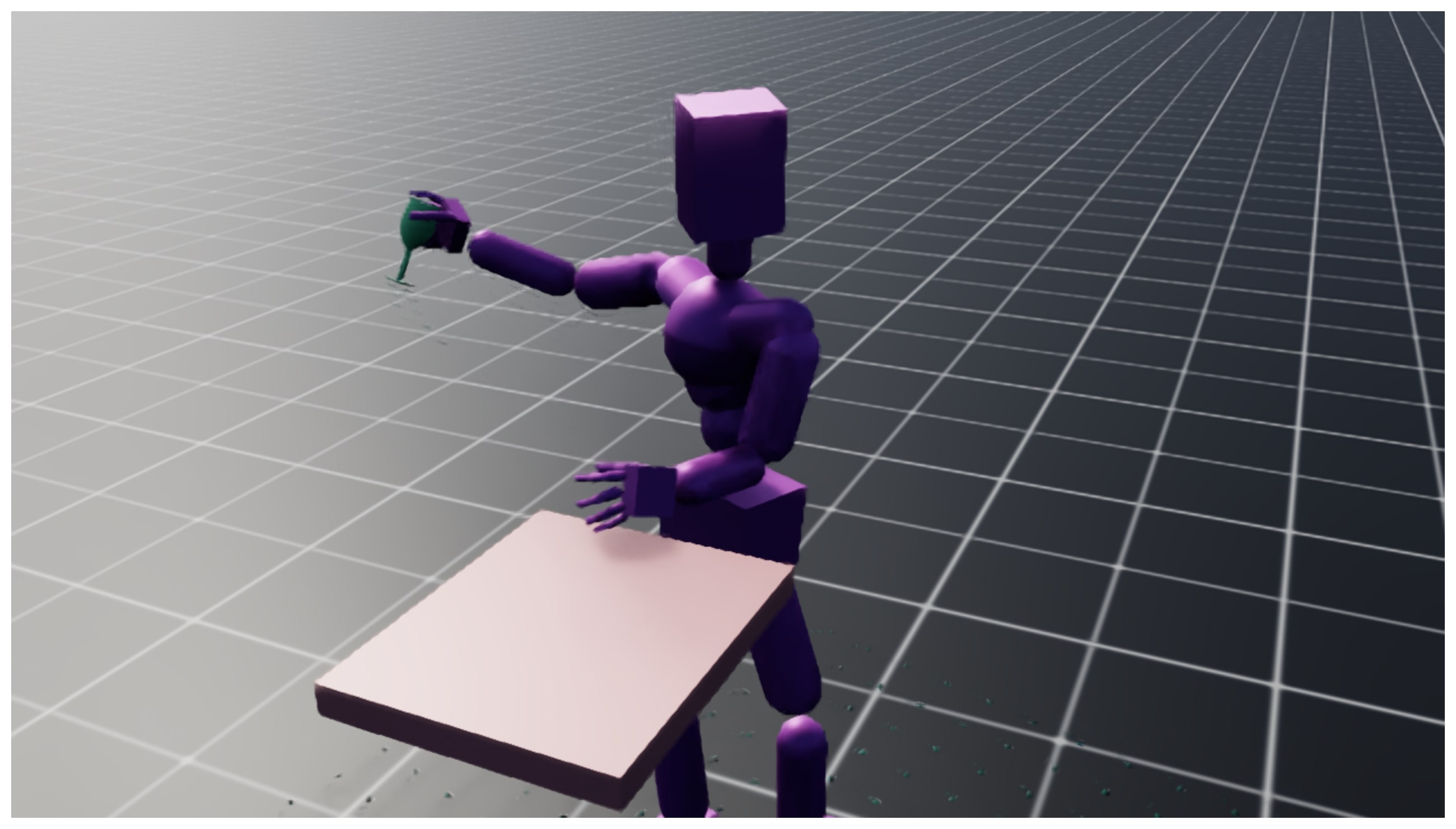}
         \caption{\textbf{Solution adaptation:} The agent picks up the object with its left hand. It then transfers the object to the right hand for a more natural reaching pose.}
         \label{fig: wine glass bi manual}
    \end{subfigure}
    \begin{subfigure}[b]{\textwidth}
         \centering
         \includegraphics[trim={4cm 0cm 20cm 6.2cm},clip,width=0.164\textwidth]{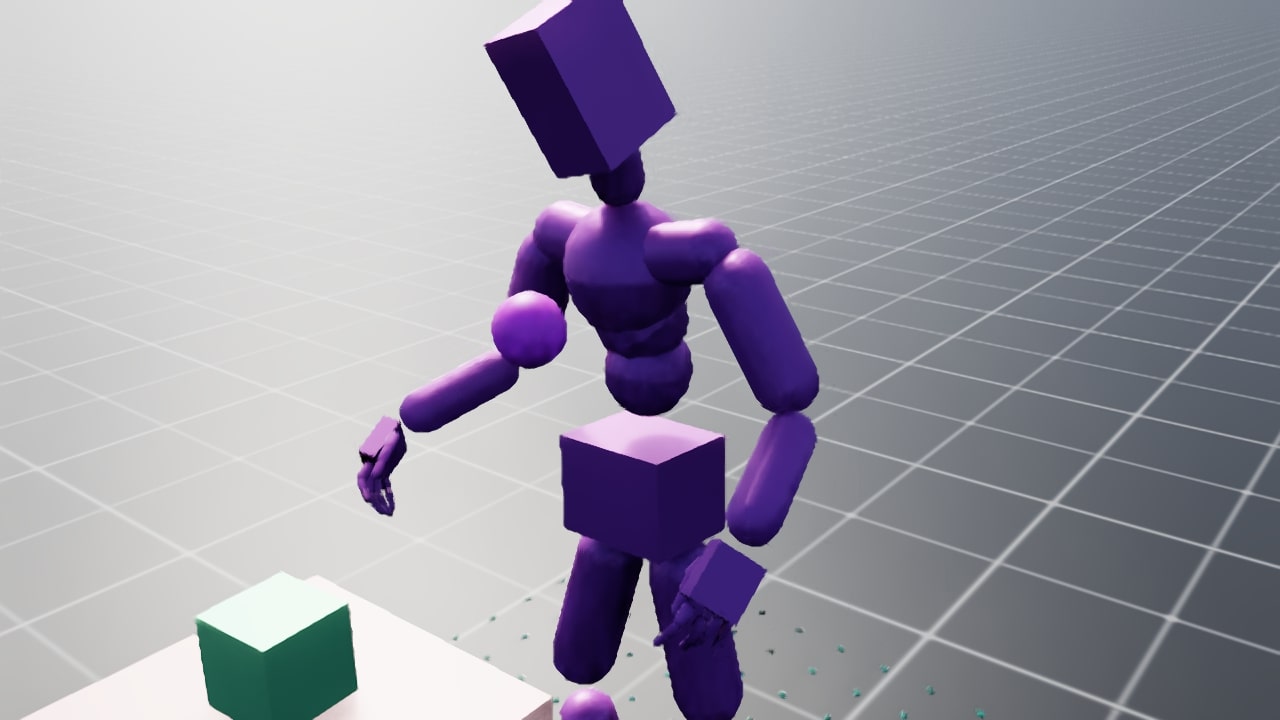}\hfill
         \includegraphics[trim={4cm 0cm 20cm 6.2cm},clip,width=0.164\textwidth]{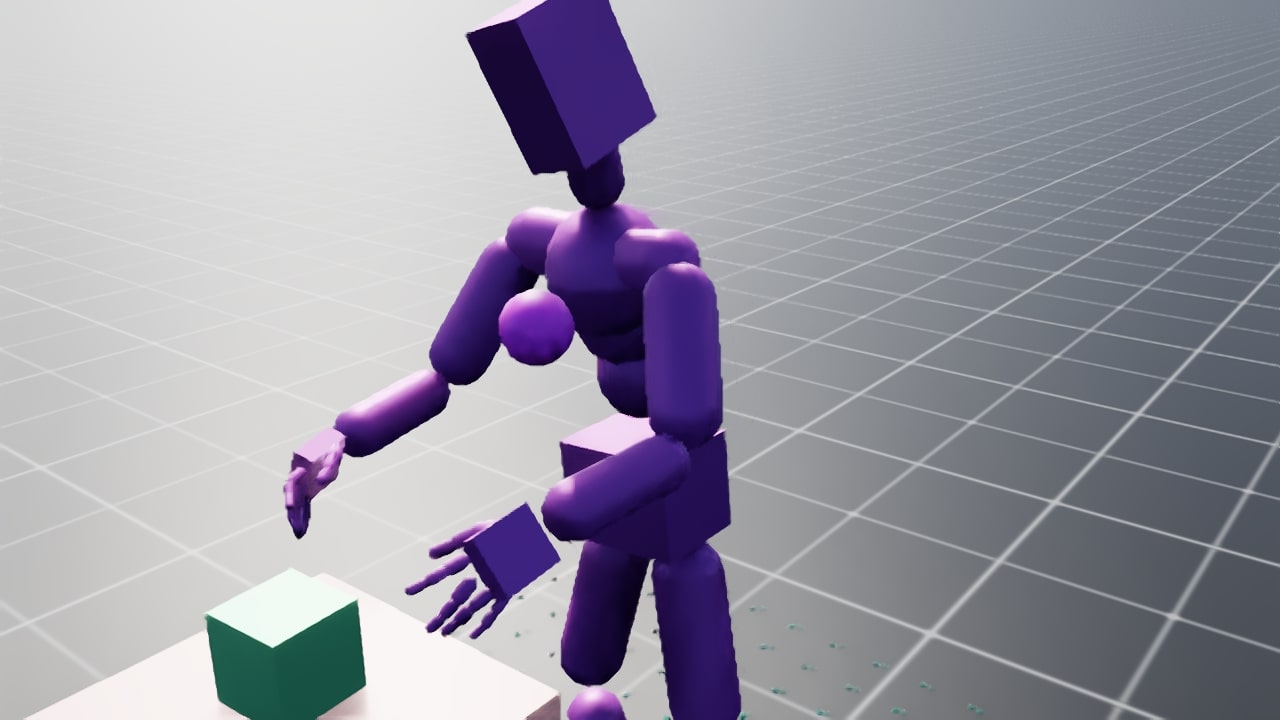}\hfill
         \includegraphics[trim={4cm 0cm 20cm 6.2cm},clip,width=0.164\textwidth]{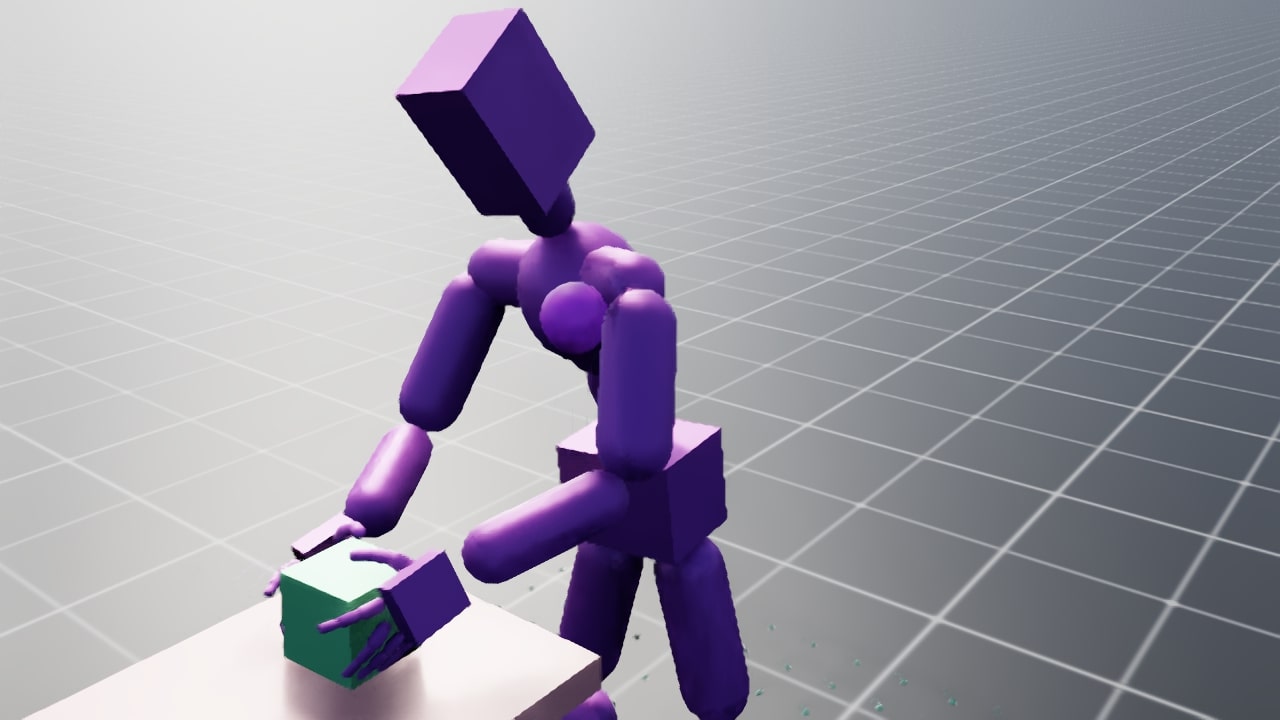}\hfill
         \includegraphics[trim={10cm 2cm 14cm 4.2cm},clip,width=0.164\textwidth]{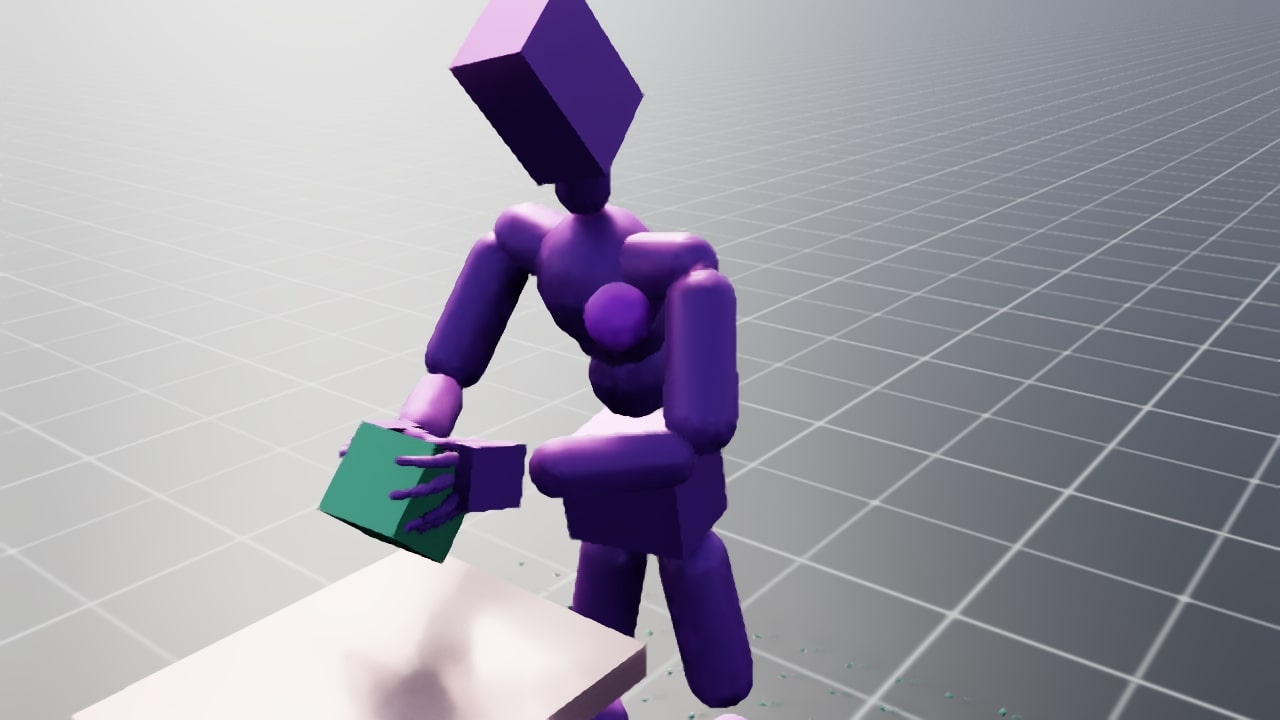}\hfill
         \includegraphics[trim={10cm 3cm 14cm 3.2cm},clip,width=0.164\textwidth]{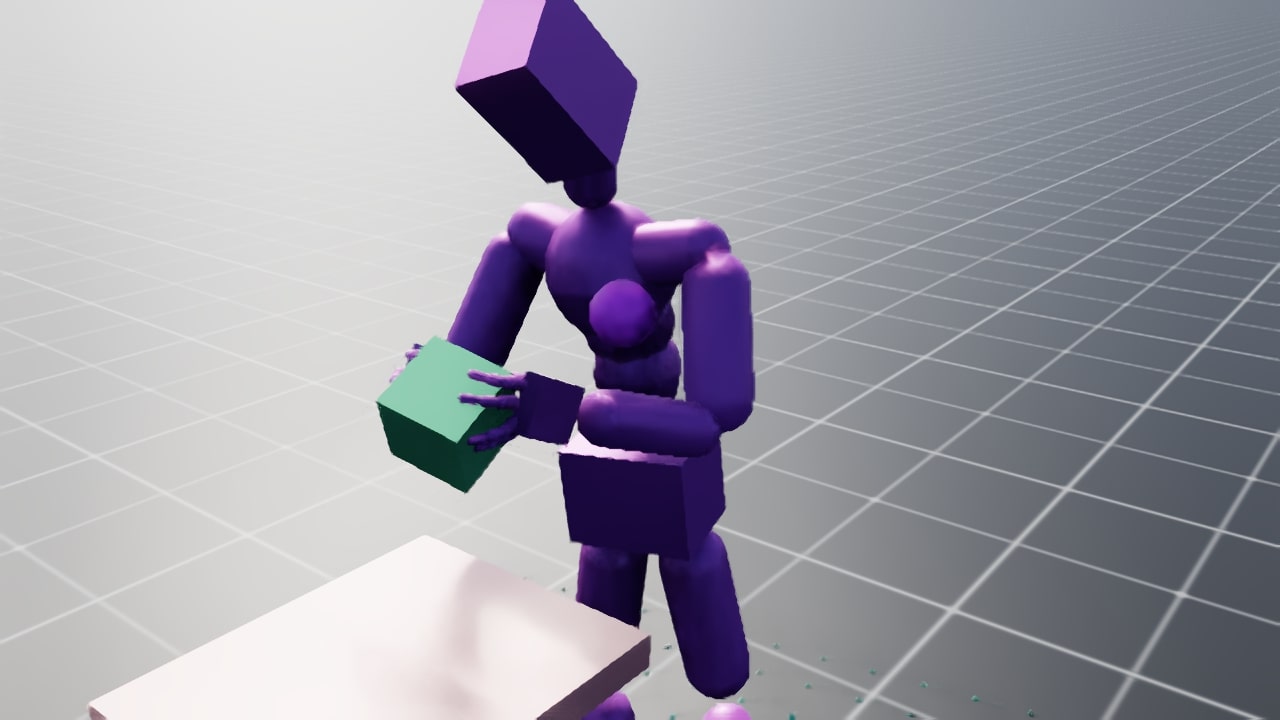}\hfill
         \includegraphics[trim={9cm 2cm 15cm 4.2cm},clip,width=0.164\textwidth]{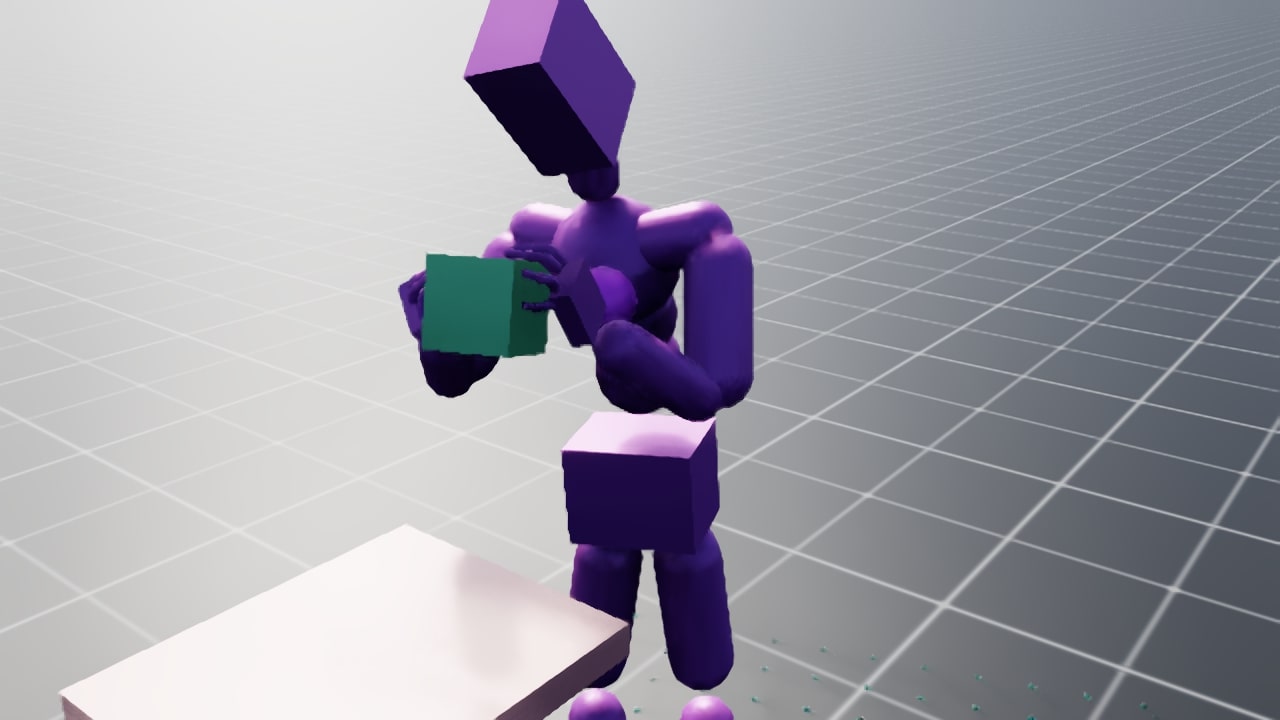}
         \caption{\textbf{Bi-manual:} \generator\ uses both hands when tasked with lifting a large object.}
         \label{fig: cube large manual}
    \end{subfigure}
    \caption{\textbf{\generator\ -- full-body motion from sparse object goals:} \generator\ is conditioned on when and where to transport the object.}
    \label{fig: object goals}
\end{figure*}

\begin{figure*}[t]
     \centering
     \begin{subfigure}[b]{\textwidth}
         \centering
         \includegraphics[trim={20cm 10cm 34cm 3.9cm},clip,width=0.164\textwidth]{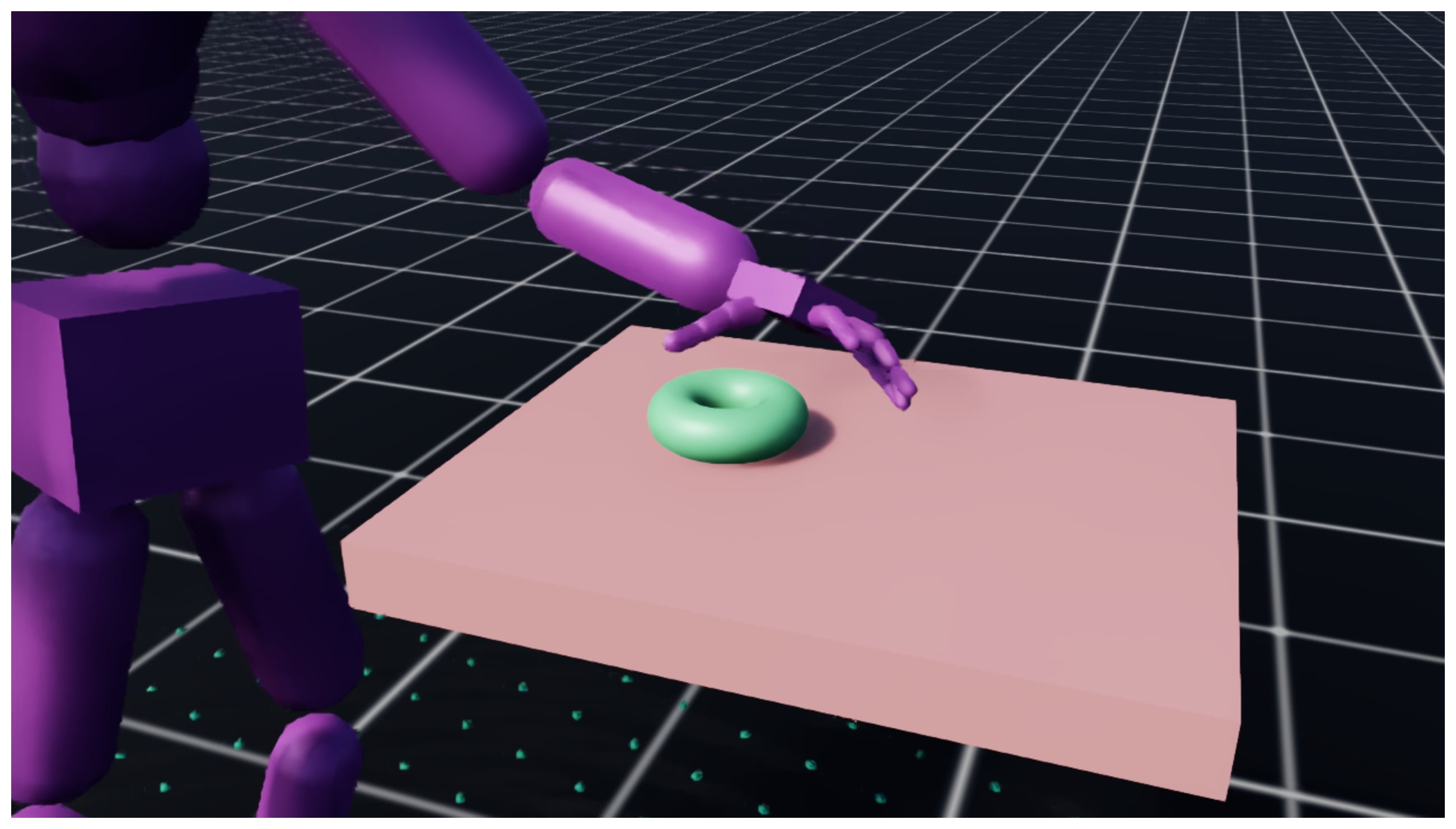}\hfill
         \includegraphics[trim={20cm 10cm 34cm 3.9cm},clip,width=0.164\textwidth]{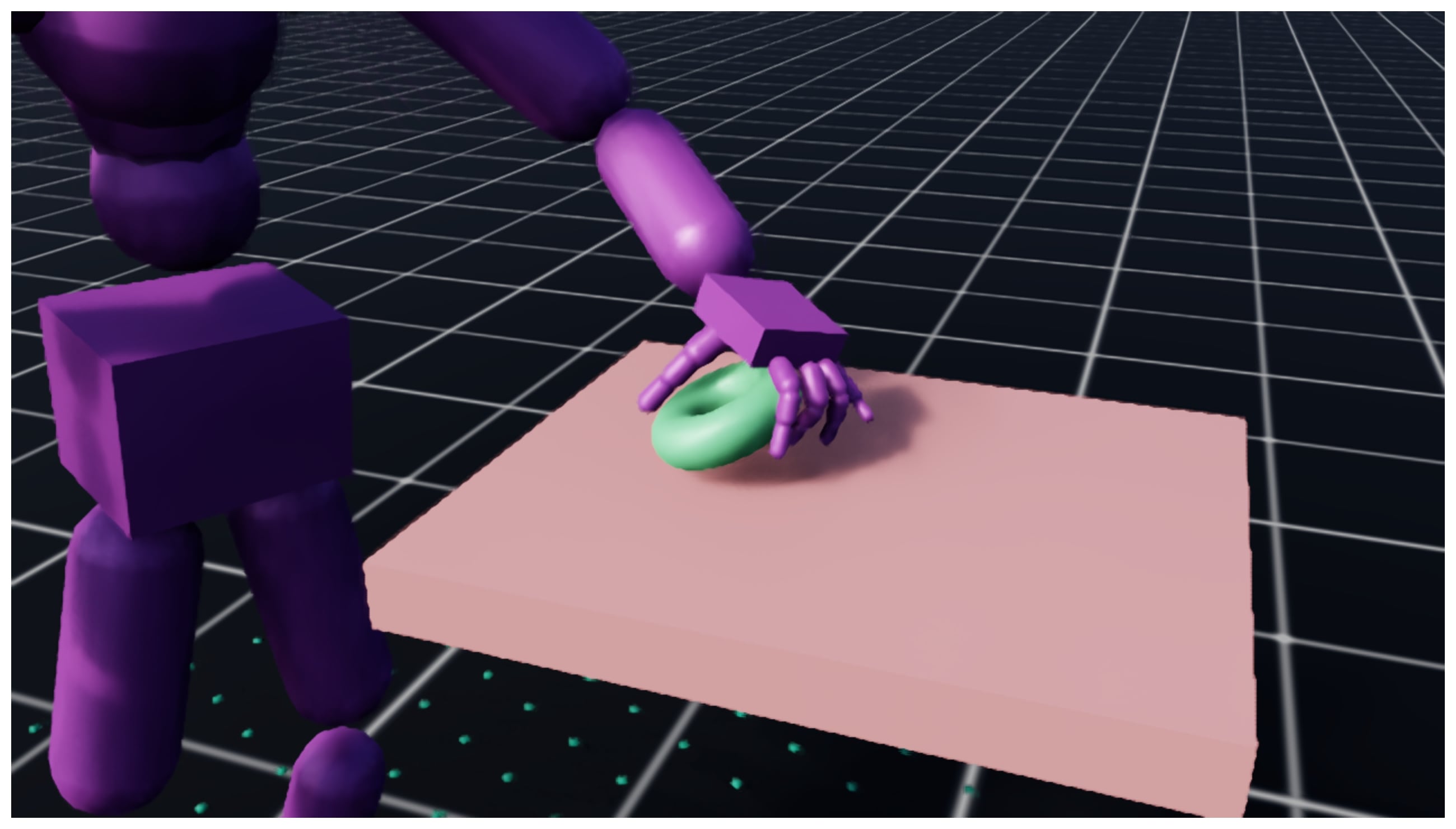}\hfill
         \includegraphics[trim={24cm 10cm 30cm 3.9cm},clip,width=0.164\textwidth]{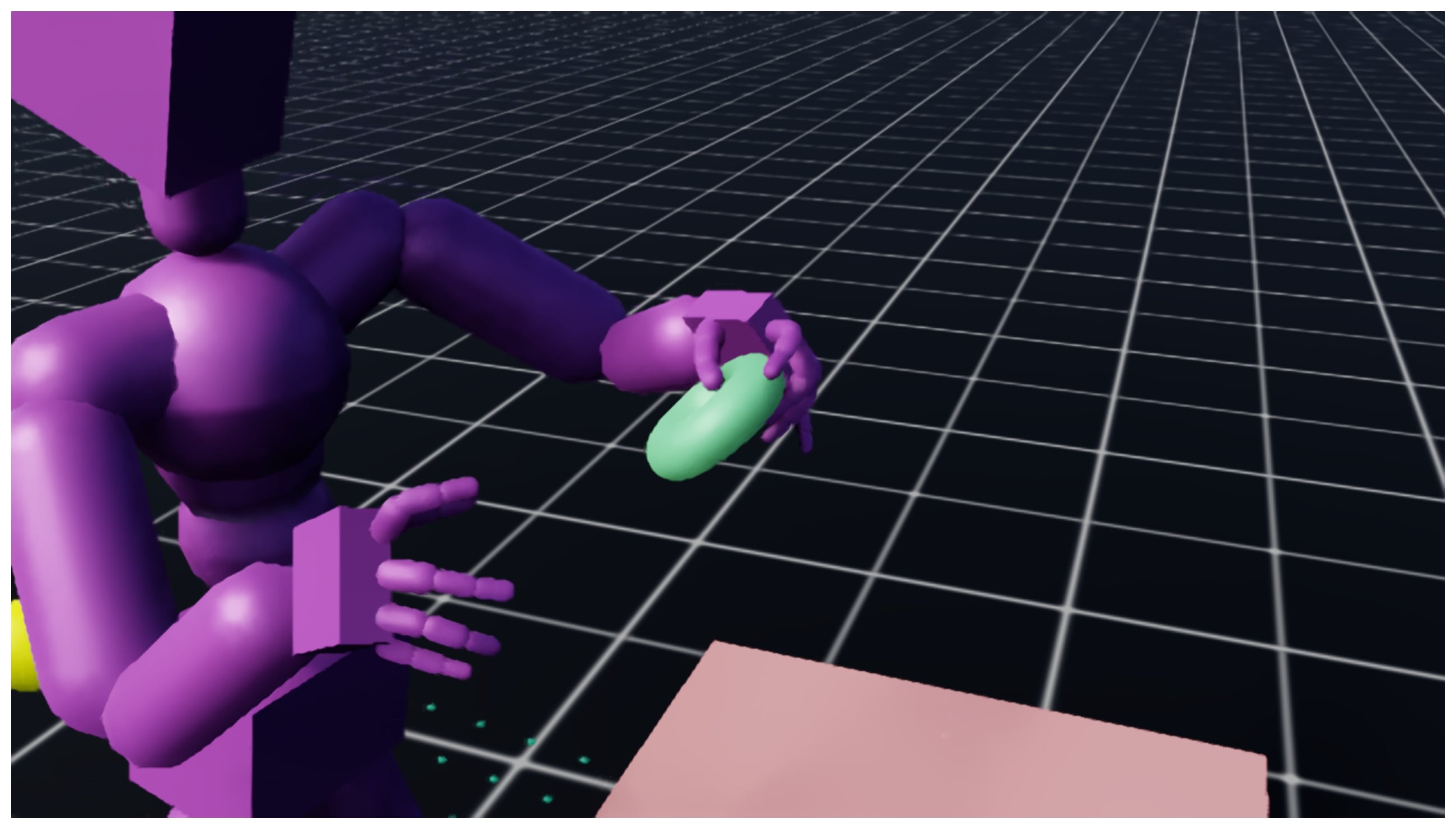}\hfill
         \includegraphics[trim={24cm 10cm 30cm 3.9cm},clip,width=0.164\textwidth]{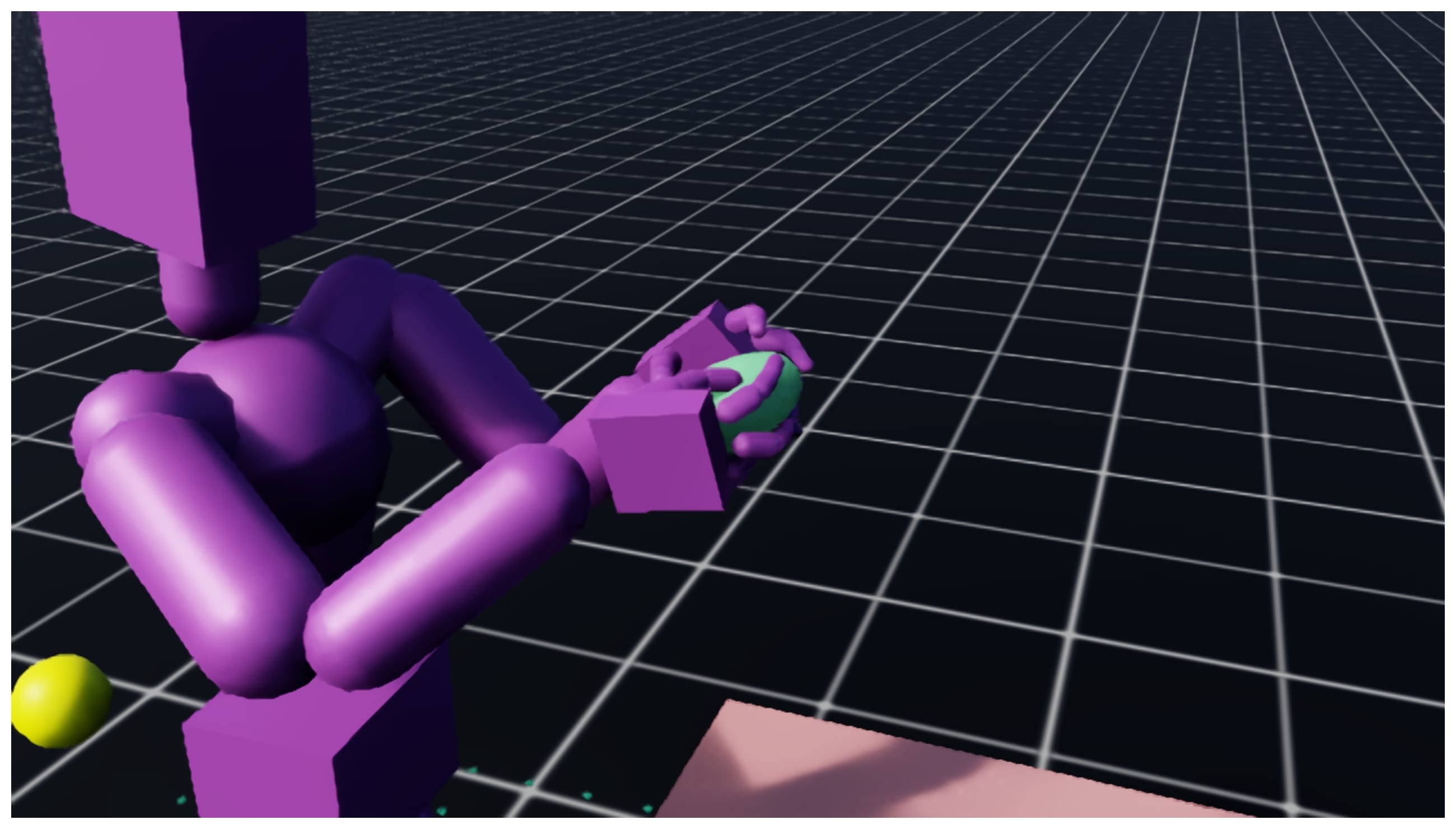}\hfill
         \includegraphics[trim={24cm 10cm 30cm 3.9cm},clip,width=0.164\textwidth]{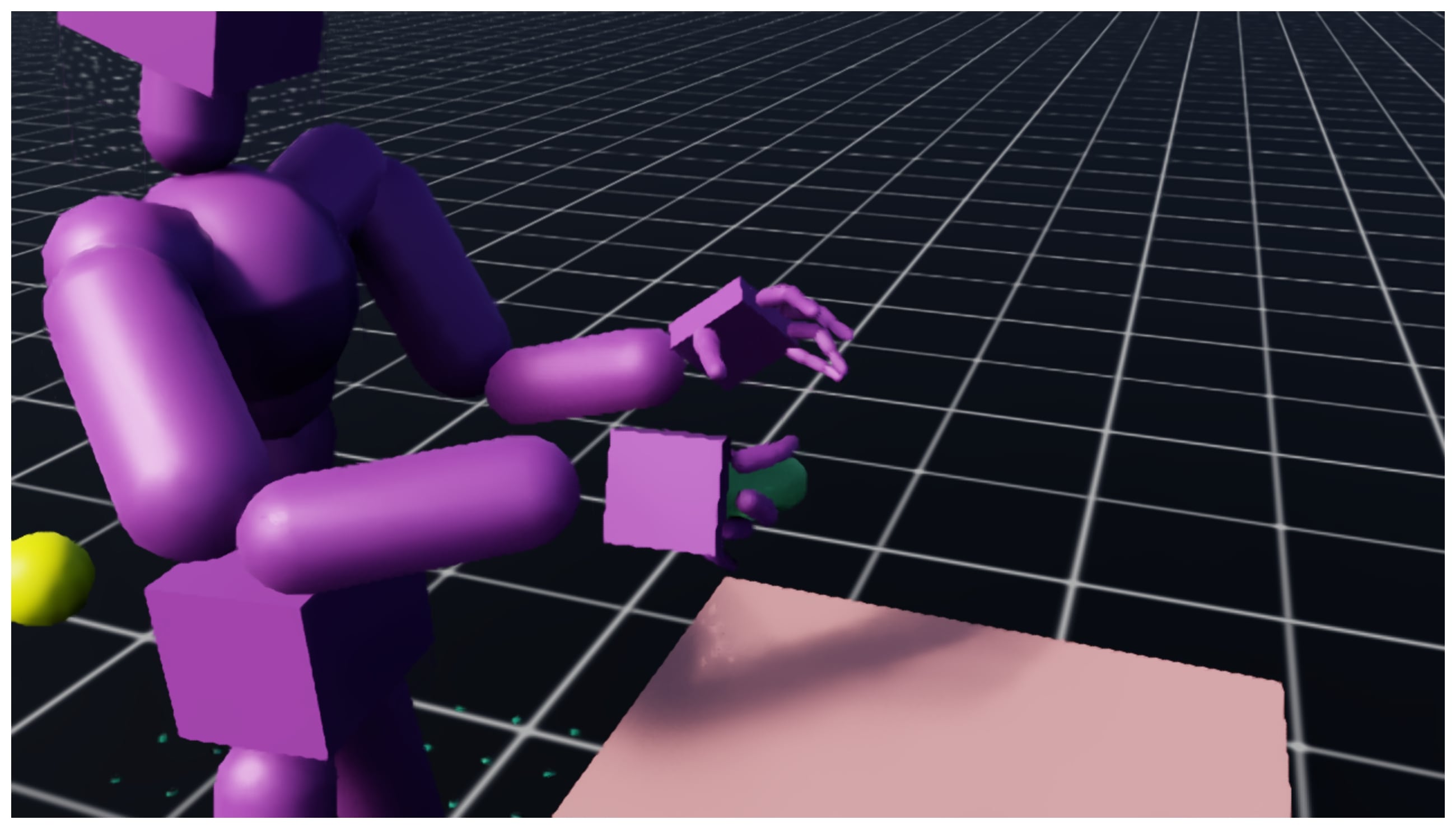}\hfill
         \includegraphics[trim={24cm 10cm 30cm 3.9cm},clip,width=0.164\textwidth]{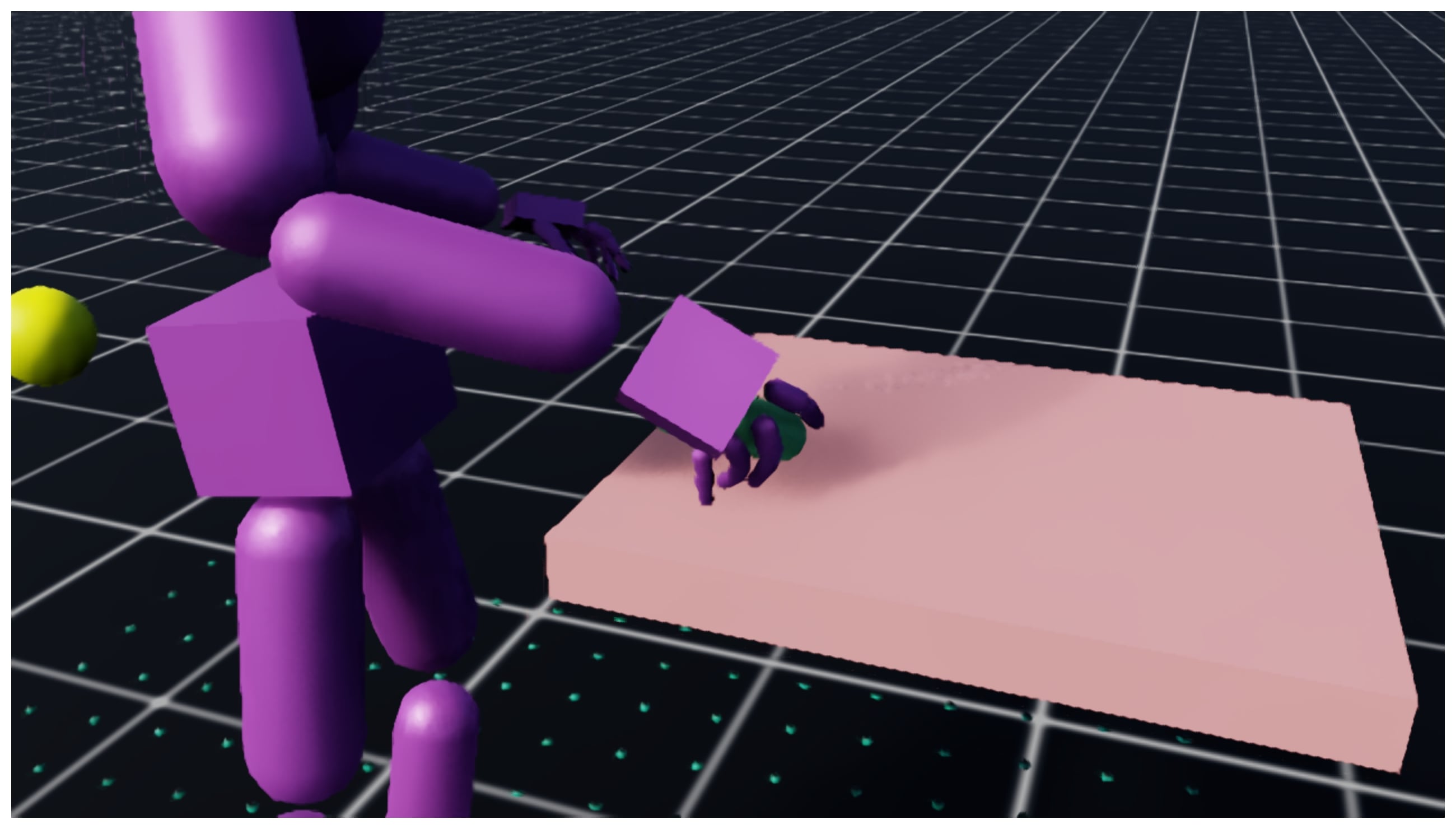}
         \caption{\textbf{Inspecting a large torus:} The agent picks up the object with its left hand, holds it with both hands while inspecting, transfers to the right hand, and puts it down.}
         \label{fig: torus large generative}
    \end{subfigure}\\
    \begin{subfigure}[b]{\textwidth}
         \centering
         \includegraphics[trim={16cm 7cm 38cm 6.9cm},clip,width=0.164\textwidth]{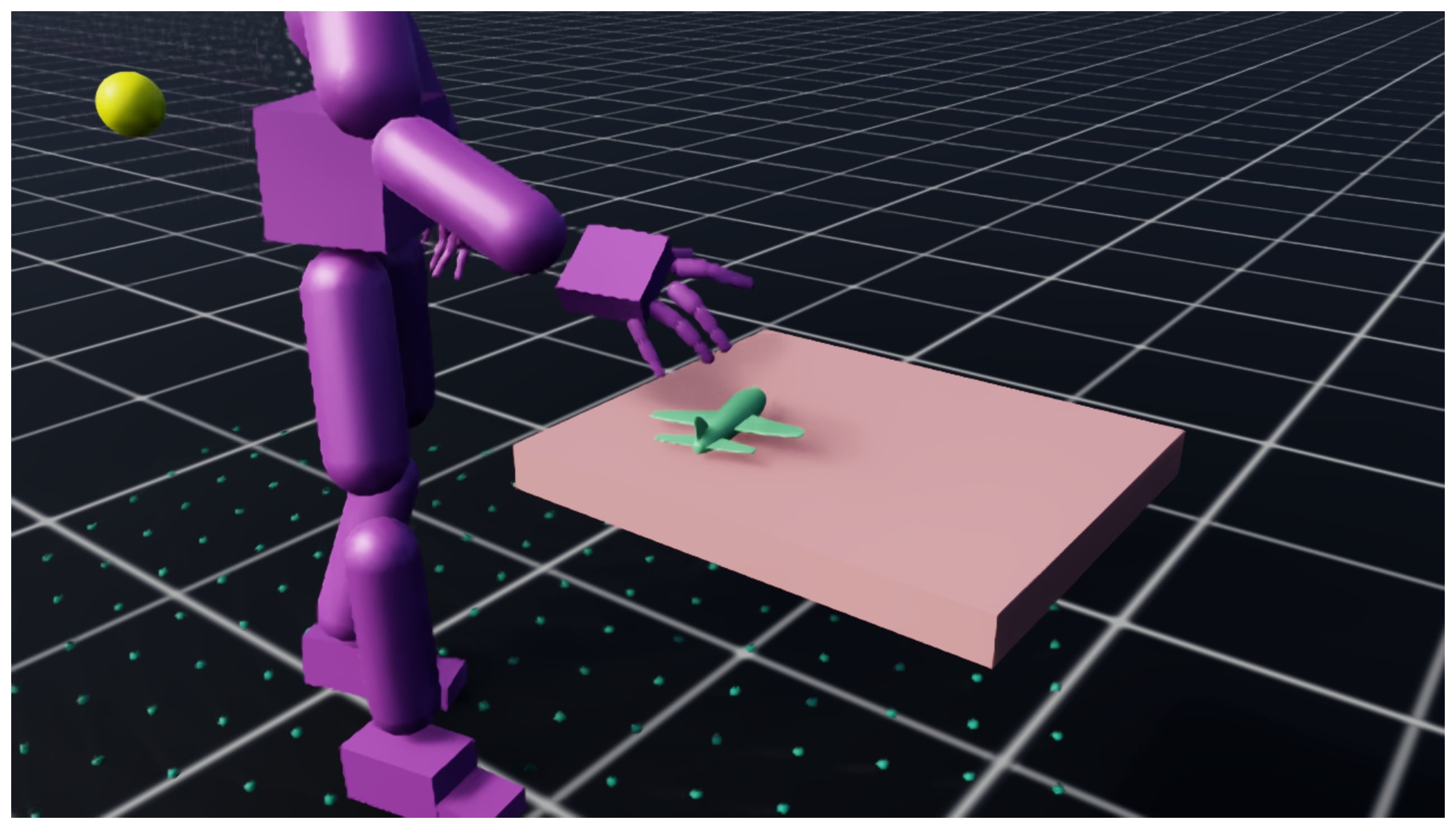}\hfill
         \includegraphics[trim={20cm 10cm 34cm 3.9cm},clip,width=0.164\textwidth]{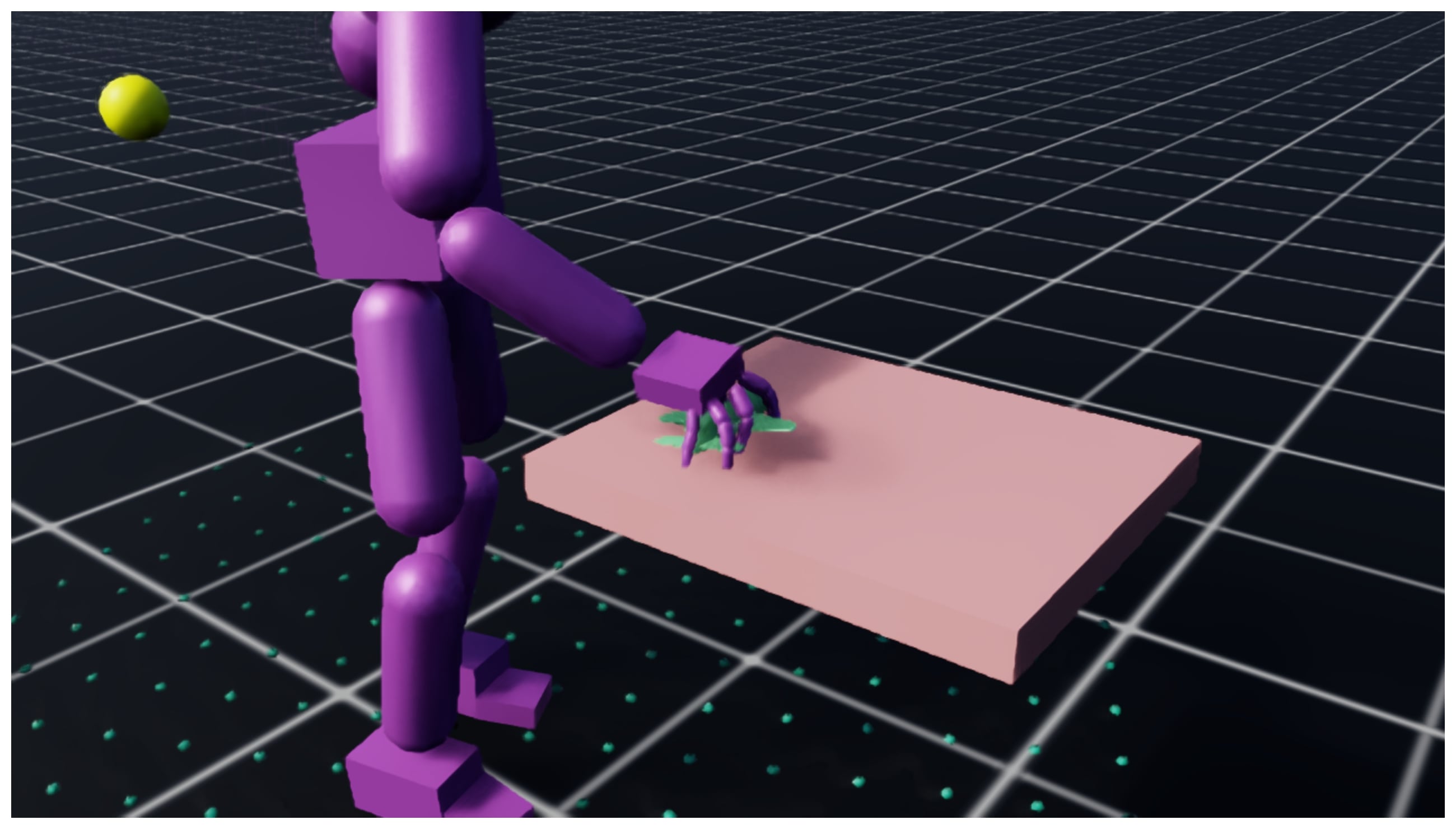}\hfill
         \includegraphics[trim={22cm 5cm 32cm 8.9cm},clip,width=0.164\textwidth]{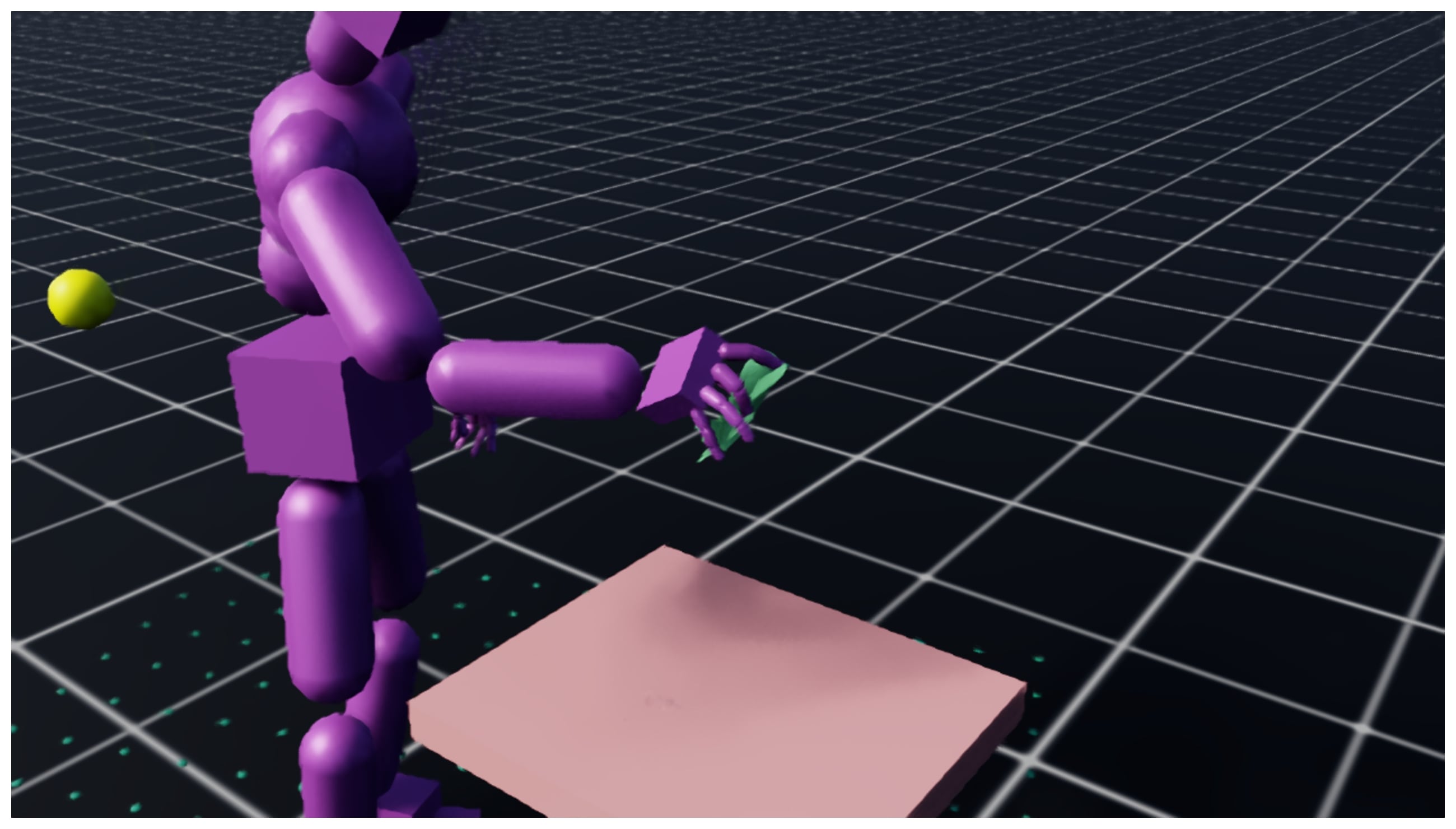}\hfill
         \includegraphics[trim={22cm 10cm 32cm 3.9cm},clip,width=0.164\textwidth]{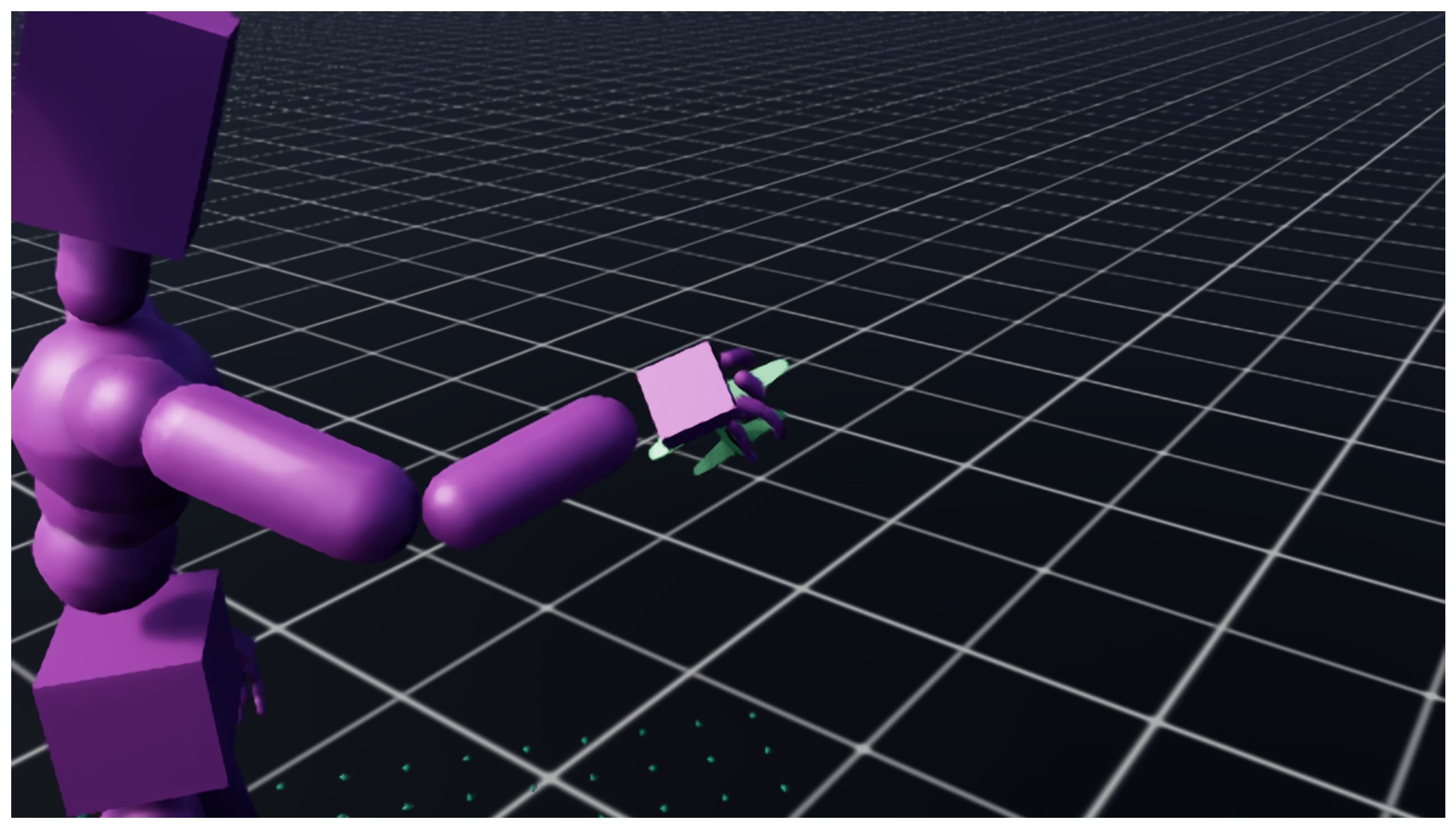}\hfill
         \includegraphics[trim={20cm 10cm 34cm 3.9cm},clip,width=0.164\textwidth]{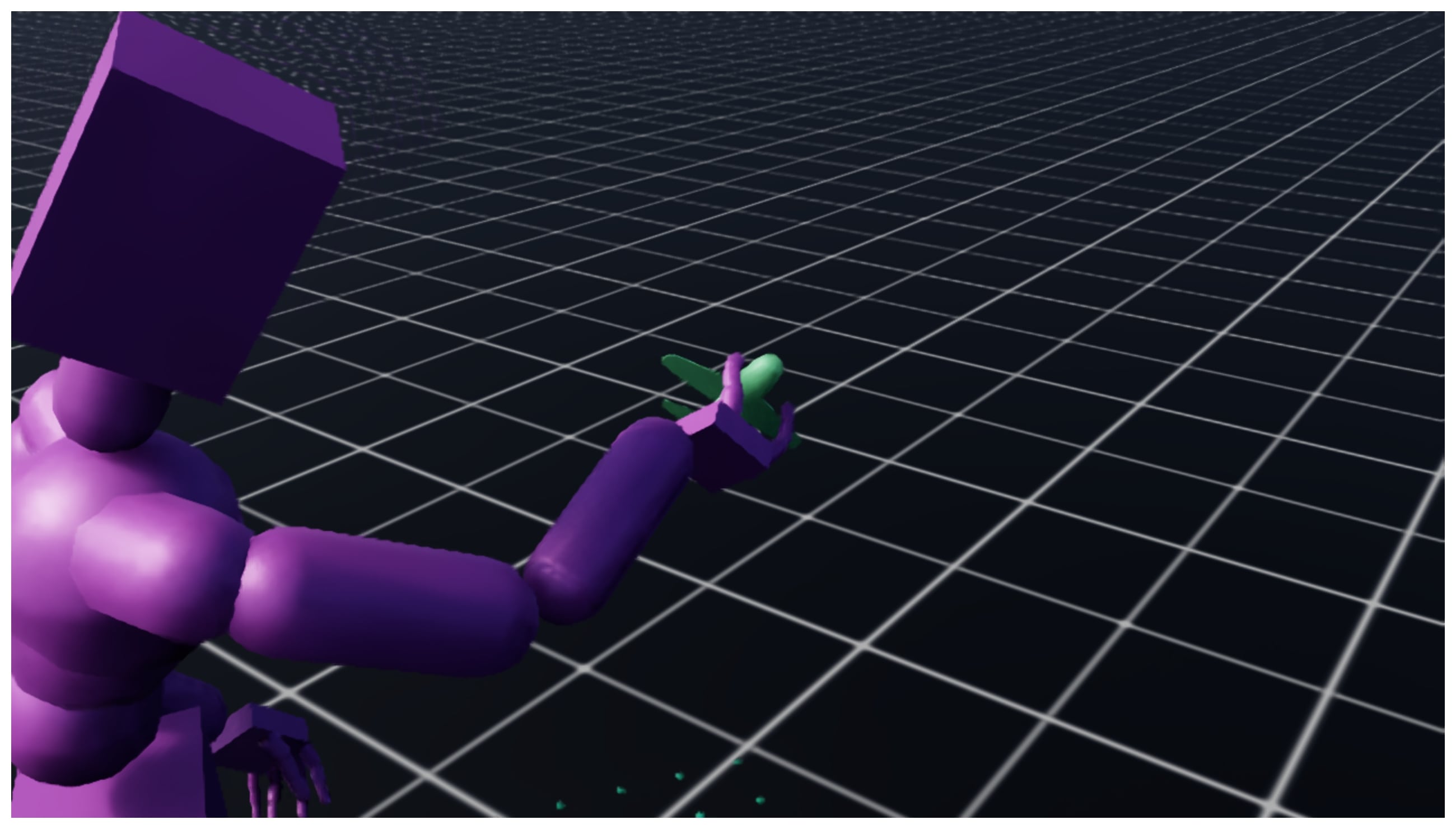}\hfill
         \includegraphics[trim={18cm 10cm 36cm 3.9cm},clip,width=0.164\textwidth]{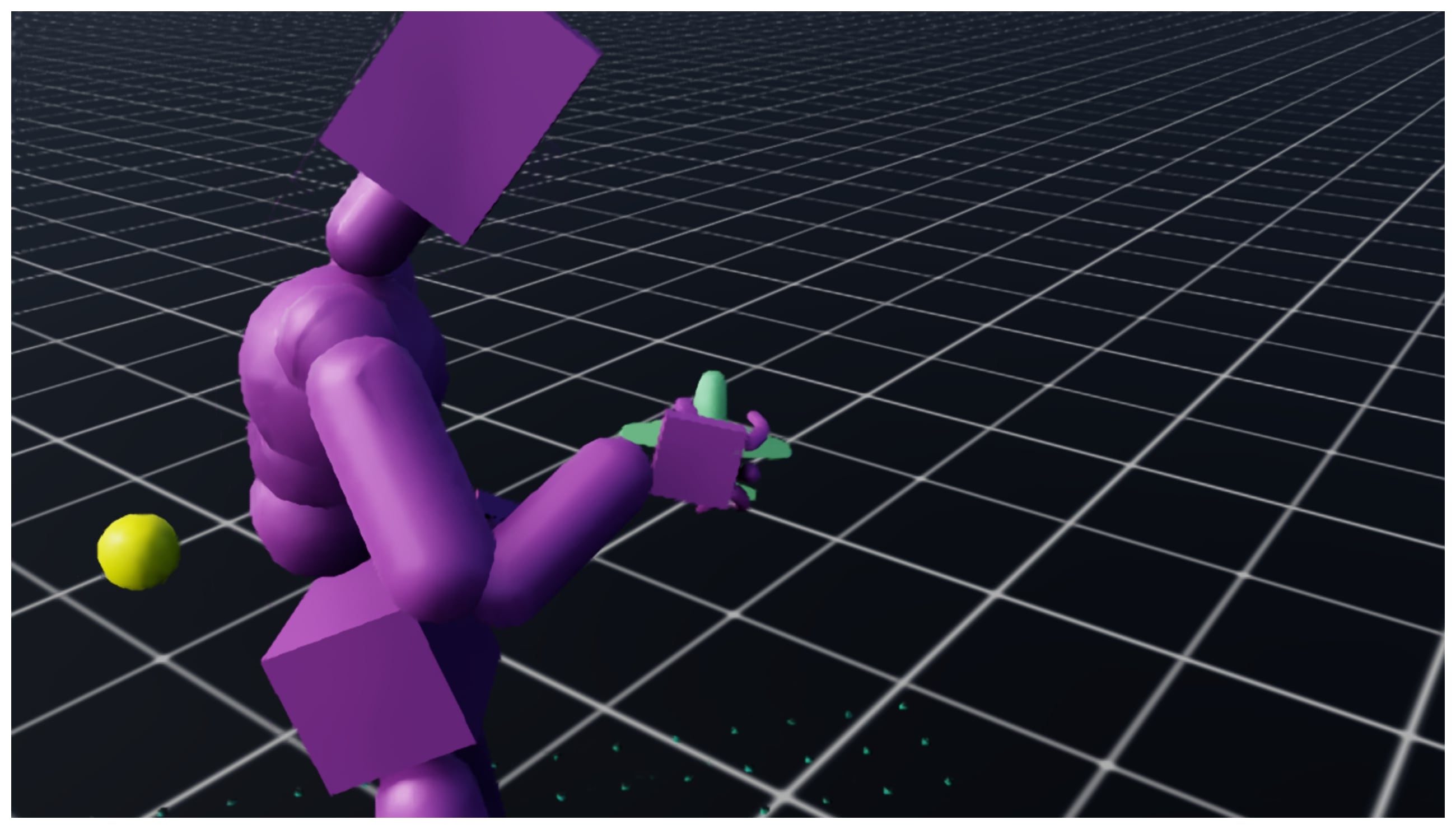}
         \caption{\textbf{Flying an airplane:} Observing only the object position and BPS \cite{prokudin2019efficient}, the agent ``flys'' the toy airplane through the air.}
         \label{fig: airplane generative}
    \end{subfigure}
    \caption{\textbf{\generator\ -- generative interaction:} When only conditioned on the pelvis position, \generator\ generates novel interactions.}
    \label{fig: purely generative}
\end{figure*}

\paragraph{Qualitative analysis.} \edited{In \cref{fig: teleop,fig: object goals,fig: purely generative}, we qualitatively demonstrate \generator’s versatility and the nuanced behaviors it can produce. As shown in \cref{fig: teleop}, when given goals for wrist position and rotation, \generator\ achieves stable and accurate pose tracking. Interestingly, even without explicit instructions, the policy often infers when to make or break contact with objects to facilitate the desired poses, highlighting its learned understanding of physical interaction. In \cref{fig: object goals}, we show that \generator\ can accomplish sparse object goals specified several seconds into the future. For example, transferring an object to a target location by generating the necessary full-body motion and even hand-to-hand object transfer when needed. Finally, \cref{fig: purely generative} highlights purely generative behavior when the policy runs with minimal or no explicit goals. In such cases, \generator\ reproduces natural interaction patterns from the GRAB training distribution, such as picking up a toy airplane and ``flying'' it through the air, showcasing its ability to generate contextually relevant, human-like actions without direct prompting.}

% \paragraph{Qualitative analysis.} We qualitatively showcase \generator's versatility and the nuanced behaviors it can generate. \cref{fig: teleop}: When provided with goals for wrist position and rotation, \generator\ demonstrates stable and accurate pose tracking. Notably, even without explicit instructions to do so, the policy often infers the appropriate moments to make or break contact with an object to facilitate the target poses, highlighting its learned understanding of physical interaction. \cref{fig: object goals}: \generator\ successfully achieves sparse object goals specified several seconds into the future. For example, it can be tasked to transfer an object to a target location, and the policy will generate the necessary full-body motion, including transferring the object between hands, to complete the objective. \cref{fig: purely generative}: When deployed with minimal or no explicit goal constraints, \generator\ exhibits interesting emergent behaviors that reflect the underlying distribution of the GRAB training data. An example of this is the policy picking up a toy airplane and ``flying'' it through the air, a common interaction pattern present in the dataset for that specific object. This demonstrates its capacity to generate contextually relevant, albeit unprompted, human-like interactions.

These qualitative examples underscore \generator's ability to not only follow specified goals but also to generate rich, physically grounded, and contextually appropriate behaviors learned from the human demonstration data.

\section{Limitations}

While our framework demonstrates significant progress in versatile full-body-manipulation, we identify several limitations that offer opportunities for future research:

\paragraph{Fidelity of the Unified Tracker (\tracker):} We observed that training \tracker\ to track the full diversity of the GRAB dataset, while successful, sometimes results in slightly lower reconstruction fidelity for individual complex sequences compared to a tracker specialized (i.e., overfit) to only that single motion. This suggests that the current policy representation or capacity might be a bottleneck when learning a very broad repertoire of intricate skills. Future work could explore more expressive policy architectures or adaptive mechanisms within the tracker.

\paragraph{Reconstruction Coverage:} Despite our efforts in data processing and robust reward design, \tracker\ does not reconstruct all motions from the processed GRAB dataset. Certain interactions remain challenging for the current system to reproduce faithfully within the physics simulation, and the performance of \generator\ on new and unseen tasks leaves room for improvement.

\paragraph{Granularity of Control in \generator:} Currently, \generator\ infers a significant amount of the interaction behavior (e.g., specific grasp points on an object, precise timing of contact engagement) based on the sparse goals and its learned priors from \tracker. It does not allow for explicit user control over how an interaction should be performed at a fine-grained level, such as specifying exact contact locations on the object surface.

Addressing these limitations could lead to more capable, precise, and controllable virtual humans, further bridging the gap between high-level creative intent and realistic, interactive animation.

\section{Conclusions}

In this work, we presented a novel two-stage framework for versatile, physics-based full-body-manipulation, enabling digital humanoids to perform complex interactions with objects in response to intuitive, high-level goals. Our approach bridges the gap between the need for precise physical execution and the desire for flexible, versatile control. The resulting \generator\ successfully synthesizes diverse and physically plausible full-body behaviors from sparse, underspecified goals, effectively navigating the multi-modal solution spaces inherent in versatile control.

Our experiments demonstrate that this framework can produce sophisticated manipulation behaviors, such as precise object placements, hand-to-hand transfers, and dynamic interaction with tools, all driven by simple goal specifications. The ability to control both the character and the object through a unified interface opens up exciting possibilities for animators and creators of interactive virtual experiences, simplifying the creation of nuanced and believable character actions.

\bibliographystyle{ACM-Reference-Format}
\bibliography{sample-bibliography}

@String{Computer = "{IEEE} Computer" }

@String{Macmillan = "Macmillan" }

@String{Springer = "Springer-Verlag" }

@article{tessler2024maskedmimic,
  title={Maskedmimic: Unified physics-based character control through masked motion inpainting},
  author={Tessler, Chen and Guo, Yunrong and Nabati, Ofir and Chechik, Gal and Peng, Xue Bin},
  journal={ACM Transactions on Graphics (TOG)},
  volume={43},
  number={6},
  pages={1--21},
  year={2024},
  publisher={ACM New York, NY, USA}
}

@article{luo2024omnigrasp,
  title={Omnigrasp: Grasping diverse objects with simulated humanoids},
  author={Luo, Zhengyi and Cao, Jinkun and Christen, Sammy and Winkler, Alexander and Kitani, Kris and Xu, Weipeng},
  journal={Advances in Neural Information Processing Systems},
  volume={37},
  pages={2161--2184},
  year={2024}
}

@inproceedings{taheri2020grab,
  title={GRAB: A dataset of whole-body human grasping of objects},
  author={Taheri, Omid and Ghorbani, Nima and Black, Michael J and Tzionas, Dimitrios},
  booktitle={Computer Vision--ECCV 2020: 16th European Conference, Glasgow, UK, August 23--28, 2020, Proceedings, Part IV 16},
  pages={581--600},
  year={2020},
  organization={Springer}
}

@article{luo2024precise,
  title={Precise and Dexterous Robotic Manipulation via Human-in-the-Loop Reinforcement Learning},
  author={Luo, Jianlan and Xu, Charles and Wu, Jeffrey and Levine, Sergey},
  journal={arXiv preprint arXiv:2410.21845},
  year={2024}
}

@article{iyer2024open,
  title={Open teach: A versatile teleoperation system for robotic manipulation},
  author={Iyer, Aadhithya and Peng, Zhuoran and Dai, Yinlong and Guzey, Irmak and Haldar, Siddhant and Chintala, Soumith and Pinto, Lerrel},
  journal={arXiv preprint arXiv:2403.07870},
  year={2024}
}

@article{mittal2023orbit,
   author={Mittal, Mayank and Yu, Calvin and Yu, Qinxi and Liu, Jingzhou and Rudin, Nikita and Hoeller, David and Yuan, Jia Lin and Singh, Ritvik and Guo, Yunrong and Mazhar, Hammad and Mandlekar, Ajay and Babich, Buck and State, Gavriel and Hutter, Marco and Garg, Animesh},
   journal={IEEE Robotics and Automation Letters},
   title={Orbit: A Unified Simulation Framework for Interactive Robot Learning Environments},
   year={2023},
   volume={8},
   number={6},
   pages={3740-3747},
   doi={10.1109/LRA.2023.3270034}
}

@book{puterman2014markov,
  title={Markov decision processes: discrete stochastic dynamic programming},
  author={Puterman, Martin L},
  year={2014},
  publisher={John Wiley \& Sons}
}

@inproceedings{SMPL-X:2019,
  title = {Expressive Body Capture: {3D} Hands, Face, and Body from a Single Image},
  author = {Pavlakos, Georgios and Choutas, Vasileios and Ghorbani, Nima and Bolkart, Timo and Osman, Ahmed A. A. and Tzionas, Dimitrios and Black, Michael J.},
  booktitle = {Proceedings IEEE Conf. on Computer Vision and Pattern Recognition (CVPR)},
  pages     = {10975--10985},
  year = {2019}
}

@inproceedings{zhou2019continuity,
  title={On the continuity of rotation representations in neural networks},
  author={Zhou, Yi and Barnes, Connelly and Lu, Jingwan and Yang, Jimei and Li, Hao},
  booktitle={Proceedings of the IEEE/CVF conference on computer vision and pattern recognition},
  pages={5745--5753},
  year={2019}
}

@inproceedings{luo2023perpetual,
  title={Perpetual humanoid control for real-time simulated avatars},
  author={Luo, Zhengyi and Cao, Jinkun and Kitani, Kris and Xu, Weipeng and others},
  booktitle={Proceedings of the IEEE/CVF International Conference on Computer Vision},
  pages={10895--10904},
  year={2023}
}

@inproceedings{prokudin2019efficient,
  title={Efficient learning on point clouds with basis point sets},
  author={Prokudin, Sergey and Lassner, Christoph and Romero, Javier},
  booktitle={Proceedings of the IEEE/CVF international conference on computer vision},
  pages={4332--4341},
  year={2019}
}

@inproceedings{xu2025intermimic,
  title = {{InterMimic}: Towards Universal Whole-Body Control for Physics-Based Human-Object Interactions},
  author = {Xu, Sirui and Ling, Hung Yu and Wang, Yu-Xiong and Gui, Liang-Yan},
  booktitle = {CVPR},
  year = {2025},
}

@article{park2019learning,
  title={Learning predict-and-simulate policies from unorganized human motion data},
  author={Park, Soohwan and Ryu, Hoseok and Lee, Seyoung and Lee, Sunmin and Lee, Jehee},
  journal={ACM Transactions on Graphics (TOG)},
  volume={38},
  number={6},
  pages={1--11},
  year={2019},
  publisher={ACM New York, NY, USA}
}

@article{won2020scalable,
  title={A scalable approach to control diverse behaviors for physically simulated characters},
  author={Won, Jungdam and Gopinath, Deepak and Hodgins, Jessica},
  journal={ACM Transactions on Graphics (TOG)},
  volume={39},
  number={4},
  pages={33--1},
  year={2020},
  publisher={ACM New York, NY, USA}
}

@article{peng2018deepmimic,
  title={Deepmimic: Example-guided deep reinforcement learning of physics-based character skills},
  author={Peng, Xue Bin and Abbeel, Pieter and Levine, Sergey and Van de Panne, Michiel},
  journal={ACM Transactions On Graphics (TOG)},
  volume={37},
  number={4},
  pages={1--14},
  year={2018},
  publisher={ACM New York, NY, USA}
}

@misc{ProtoMotions,
  title = {ProtoMotions: Physics-based Character Animation},
  author = {Tessler, Chen and Juravsky, Jordan and Guo, Yunrong and Jiang, Yifeng and Coumans, Erwin and Peng, Xue Bin},
  year = {2024},
  publisher = {GitHub},
  journal = {GitHub repository},
  howpublished = {\url{https://github.com/NVLabs/ProtoMotions/}},
}

@inproceedings{ross2011reduction,
  title={A reduction of imitation learning and structured prediction to no-regret online learning},
  author={Ross, St{\'e}phane and Gordon, Geoffrey and Bagnell, Drew},
  booktitle={Proceedings of the fourteenth international conference on artificial intelligence and statistics},
  pages={627--635},
  year={2011},
  organization={JMLR Workshop and Conference Proceedings}
}

@article{vaswani2017attention,
  title={Attention is all you need},
  author={Vaswani, Ashish and Shazeer, Noam and Parmar, Niki and Uszkoreit, Jakob and Jones, Llion and Gomez, Aidan N and Kaiser, {\L}ukasz and Polosukhin, Illia},
  journal={Advances in neural information processing systems},
  volume={30},
  year={2017}
}

@inproceedings{rempe2021humor,
  title={Humor: 3d human motion model for robust pose estimation},
  author={Rempe, Davis and Birdal, Tolga and Hertzmann, Aaron and Yang, Jimei and Sridhar, Srinath and Guibas, Leonidas J},
  booktitle={Proceedings of the IEEE/CVF international conference on computer vision},
  pages={11488--11499},
  year={2021}
}

@article{chi2023diffusion,
  title={Diffusion policy: Visuomotor policy learning via action diffusion},
  author={Chi, Cheng and Xu, Zhenjia and Feng, Siyuan and Cousineau, Eric and Du, Yilun and Burchfiel, Benjamin and Tedrake, Russ and Song, Shuran},
  journal={The International Journal of Robotics Research},
  pages={02783649241273668},
  year={2023},
  publisher={SAGE Publications Sage UK: London, England}
}

@article{ho2020denoising,
  title={Denoising diffusion probabilistic models},
  author={Ho, Jonathan and Jain, Ajay and Abbeel, Pieter},
  journal={Advances in neural information processing systems},
  volume={33},
  pages={6840--6851},
  year={2020}
}

@book{kahneman2011thinking,
  title={Thinking, fast and slow},
  author={Kahneman, Daniel},
  year={2011},
  publisher={macmillan}
}

@article{peng2022ase,
  title={Ase: Large-scale reusable adversarial skill embeddings for physically simulated characters},
  author={Peng, Xue Bin and Guo, Yunrong and Halper, Lina and Levine, Sergey and Fidler, Sanja},
  journal={ACM Transactions On Graphics (TOG)},
  volume={41},
  number={4},
  pages={1--17},
  year={2022},
  publisher={ACM New York, NY, USA}
}

@article{li2024virt,
  title={VIRT: Vision Instructed Transformer for Robotic Manipulation},
  author={Li, Zhuoling and Ren, Liangliang and Yang, Jinrong and Zhao, Yong and Wu, Xiaoyang and Xu, Zhenhua and Bai, Xiang and Zhao, Hengshuang},
  journal={arXiv preprint arXiv:2410.07169},
  year={2024}
}

@article{li2025hamster,
  title={Hamster: Hierarchical action models for open-world robot manipulation},
  author={Li, Yi and Deng, Yuquan and Zhang, Jesse and Jang, Joel and Memmel, Marius and Yu, Raymond and Garrett, Caelan Reed and Ramos, Fabio and Fox, Dieter and Li, Anqi and others},
  journal={arXiv preprint arXiv:2502.05485},
  year={2025}
}

@article{akkaya2019solving,
  title={Solving rubik's cube with a robot hand},
  author={Akkaya, Ilge and Andrychowicz, Marcin and Chociej, Maciek and Litwin, Mateusz and McGrew, Bob and Petron, Arthur and Paino, Alex and Plappert, Matthias and Powell, Glenn and Ribas, Raphael and others},
  journal={arXiv preprint arXiv:1910.07113},
  year={2019}
}

@article{yu2025skillmimic,
  title={SkillMimic-V2: Learning Robust and Generalizable Interaction Skills from Sparse and Noisy Demonstrations},
  author={Yu, Runyi and Wang, Yinhuai and Zhao, Qihan and Tsui, Hok Wai and Wang, Jingbo and Tan, Ping and Chen, Qifeng},
  journal={arXiv preprint arXiv:2505.02094},
  year={2025}
}

@article{lin2024twisting,
  publtype={informal},
  author={Toru Lin and Zhao-Heng Yin and Haozhi Qi and Pieter Abbeel and Jitendra Malik},
  title={Twisting Lids Off with Two Hands},
  year={2024},
  cdate={1704067200000},
  journal={CoRR},
  volume={abs/2403.02338},
  url={https://doi.org/10.48550/arXiv.2403.02338}
}

@article{wu2025uniphys,
  title={UniPhys: Unified Planner and Controller with Diffusion for Flexible Physics-Based Character Control},
  author={Wu, Yan and Karunratanakul, Korrawe and Luo, Zhengyi and Tang, Siyu},
  journal={arXiv preprint arXiv:2504.12540},
  year={2025}
}

@article{allshire2025visual,
  title={Visual Imitation Enables Contextual Humanoid Control},
  author={Allshire, Arthur and Choi, Hongsuk and Zhang, Junyi and McAllister, David and Zhang, Anthony and Kim, Chung Min and Darrell, Trevor and Abbeel, Pieter and Malik, Jitendra and Kanazawa, Angjoo},
  journal={arXiv preprint arXiv:2505.03729},
  year={2025}
}

@article{ze2025twist,
  title={TWIST: Teleoperated Whole-Body Imitation System},
  author={Ze, Yanjie and Chen, Zixuan and Ara{\~A}{\v{s}}jo, Jo{\~A}{\c{G}}o Pedro and Cao, Zi-ang and Peng, Xue Bin and Wu, Jiajun and Liu, C Karen},
  journal={arXiv preprint arXiv:2505.02833},
  year={2025}
}

@article{ji2024exbody2,
  title={Exbody2: Advanced expressive humanoid whole-body control},
  author={Ji, Mazeyu and Peng, Xuanbin and Liu, Fangchen and Li, Jialong and Yang, Ge and Cheng, Xuxin and Wang, Xiaolong},
  journal={arXiv preprint arXiv:2412.13196},
  year={2024}
}

@article{schulman2017proximal,
  title={Proximal policy optimization algorithms},
  author={Schulman, John and Wolski, Filip and Dhariwal, Prafulla and Radford, Alec and Klimov, Oleg},
  journal={arXiv preprint arXiv:1707.06347},
  year={2017}
}

@inproceedings{lu2025what,
  title={What Makes a Good Diffusion Planner for Decision Making?},
  author={Haofei Lu and Dongqi Han and Yifei Shen and Dongsheng Li},
  booktitle={The Thirteenth International Conference on Learning Representations},
  year={2025},
  url={https://openreview.net/forum?id=7BQkXXM8Fy}
}

@article{pignatelli2023survey,
  title={A survey of temporal credit assignment in deep reinforcement learning},
  author={Pignatelli, Eduardo and Ferret, Johan and Geist, Matthieu and Mesnard, Thomas and van Hasselt, Hado and Pietquin, Olivier and Toni, Laura},
  journal={arXiv preprint arXiv:2312.01072},
  year={2023}
}

\clearpage
\newpage

\clearpage
\newpage

\setcounter{page}{1}
% Appendix\appendix

\appendix

\section*{\centering\Huge Supplementary Material}

\section{Humanoid model and control}

Our humanoid character is based on the SMPL-X model structure \cite{SMPL-X:2019}, consistent with recent work \cite{luo2024omnigrasp,xu2025intermimic}. It possesses $N_a = 153$ actuated degrees of freedom (DoFs), corresponding to 51 joints, each with 3 DoFs. The policy's actions $a_t \in \mathbb{R}^{N_a}$ specify target positions for Proportional-Derivative (PD) controllers at each actuated DoF.

The humanoid's body links are represented using primitive geometries (capsules and boxes). This simplified geometric representation, particularly for the hands, can generate fewer contact points compared to real human hands with soft, deformable tissue. This difference can make certain fine-grained manipulation tasks harder to reconstruct. To better approximate the contact dynamics of human soft tissue interacting with objects using rigid body simulation, we use a global friction coefficient of $1.5$. This helps mitigate real-to-sim mismatch by enabling more stable grasps, such as single-handed grasping of larger objects. Furthermore, to ensure more natural and human-like finger movements, we constrain the joint limits of the hand DoFs to be within biomechanically plausible ranges, preventing uncanny configurations that can arise from the full rotational capacity of the original SMPL-X model.

In addition, we model all objects with a fixed mass of $1 [KG]$ and set the restitution at $0.7$.

\section{Retargeting}

The goal in this paper is not to reconstruct the entire GRAB dataset, but rather to learn full-body-manipulation skills. As such, we aim to utilize as much data as possible, and focus on a single humanoid morphology.

Transfering motion across morphologies isn't a trivial problem. Not only does the body-size ratio differ, but also the relative lengths of various body parts with respect to the character's height. The simplest approach for transferring motion between characters with similar (same structure) morphologies is by applying the DoF rotations from the original skeleton onto the new target skeleton. While simple, and this works nicely for learning locomotion, this does not apply to the object.

Unlike the humanoid's body parts, the object is not a fixed joint within the kinematic tree. Changing the humanoid shape results in a misalignment with the object. For example, a taller human lifting an object might lift it higher.

To tackle this, we combine an explicit object retargeting strategy with the robustness of reinforcement learning within our physics simulation. The object retargeting step, illustrated in Figure~\ref{fig: retargeting}, first corrects for the large translational discrepancies caused by the change in body shape, bringing the object approximately to the correct position relative to our character. From this more feasible starting point, the RL-trained policy, by interacting within the physics simulation, can then overcome smaller residual imperfections and learn to successfully reconstruct the nuances of the original human-object interaction.

When a reference motion (performed by a subject with a specific body shape, denoted ``original'') is applied to the target humanoid (which may have a different ``target'' shape), we adjust the object's trajectory to maintain interaction consistency. This simple approach preserves the relative positioning between the contacting hand(s) and the object. Leveraging the contact annotations, we determine the object translation offset $p$ by optimizing the following objective to best preserve the original contact relationships:
\begin{align*}
    p^* = \text{argmin}_p \sum_{j \in \text{ContactLinks}} \Bigg| &\left( \hat{c}^\text{original}_{j,t} - \hat{c}^\text{obj}_{j,t} \right) \\
    &- \left( \hat{c}^\text{target}_{j,t} - \left( \hat{c}^{\text{obj}}_{j,t} + p \right) \right) \Bigg|^2
    \label{eq:retargeting}
\end{align*}
where $\hat{c}^\text{obj}_{j,t}$ is the location joint $j$ is in contact with the object mesh at time $t$. This objective aims to find an object translation offset $p$ that maintains the same distance from bodies-in-contact to their corresponding contact coordinate on the object mesh. The retargeted object position is then $\hat{p}^\text{obj,retargeted}_t = \hat{p}^\text{obj}_t + p^*$.

Finally, as our goal is to learn manipulation behaviors and not necessarily successfully reconstruct the entire dataset, we identify and remove sequences exhibiting two primary issues:
\textbf{(1) Non-hand interactions.} Motions where crucial object interactions involve body parts other than the hands (e.g., the person places sunglasses on its face), which are beyond the scope of our hand-centric manipulation focus.
\textbf{(2) Retargeting instability.} Instances where the retargeting process might introduce discontinuities. We filter out motions if the retargeted object's center of mass ``jumps'' by more than 10cm between any two consecutive frames, as this often indicates an unstable or physically implausible retargeted interaction.

After this data processing and filtering pipeline, our training set comprises 1007 motion sequences, and our test set (derived from subject 10 in GRAB) contains 141 sequences. By closely tracking these detailed human demonstrations, $\pi_{\text{track}}$ learns to overcome remaining imperfections in the reference data and to master intricate and physically nuanced behaviors, such as coordinated bimanual operations, precise tool usage, and maintaining stable object control throughout extended manipulation sequences. We provide additional technical details in the supplementary material.

\section{Rewards}

Our reward is defined as a multiplicative reward with the following components:

Error on the human global translation: $r^{ht} = e^{-100 \cdot || \hat{p}_t - p_t ||}$.

Error on the human global rotation: $r^{hr} = e^{-2 \cdot < \hat{\theta}_t , \theta_t >}$ where $< \cdot , \cdot >$ is the quaternion difference.

Error on the human velocity: $r^{hv} = e^{-0.2 \cdot || \hat{v}_t - v_t ||}$

Error on the human angular velocity: $r^{hw} = e^{-0.02 \cdot || \hat{\omega}_t - \omega_t ||}$

Error on the human energy: $r^{pow} = e^{-0.00002 \cdot | \text{dof force}_t \cdot v_t |}$

Error on the finger contact forces: \\
$r^{penetration} = e^{-0.00001 \cdot \Pi_{j \in \text{contact bodies}} \text{contact force}_j}$

Contact rewards:
\begin{itemize}
    \item Pre-contact: \begin{itemize}
        \item Translation error: We compute the combined error across all future reference bodies in contact. \\ \begin{align*}
            \Pi_{j \in \text{in contact in the future}} e^{-100 (\hat{p}^j_t - {p}^j_t || + || ||\hat{p}^j_t - \hat{p}^\text{obj}_t || - ||{p}^j_t  - p^\text{obj}_t ||)}
        \end{align*}
        Which encourages future bodies in contact to not only maintain similar position to the reference data, but also with respect to the object. This helps prevent dynamic motion when close to contact, which may push the object far away. In addition, we use a normal cosine similarity component $\Pi_{j \in \text{in contact in the future}} (1 - \text{normal similarity}^j_t) / 2$ which compares the surface normal of the closest point for each joint in the reference with that in the simulation.
    \end{itemize}
    \item During contact we reward on the distance error for bodies that should be in contact. This encourages them to minimize the distance towards the object. A body part that is in contact is marked as distance 0. $\Pi_{j \in \text{should have contact}} e^{-2 \cdot d^j_t}$ where $d$ is the distance to the closest point on the object surface.
    \item Post-contact: \begin{itemize}
        \item Translation error: We compute the combined error across all previous reference bodies in contact. \\
        \begin{align*}
            \Pi_{j \in \text{in contact in the past}} e^{-100 (\hat{p}^j_t - {p}^j_t || + || ||\hat{p}^j_t - \hat{p}^\text{obj}_t || - ||{p}^j_t  - p^\text{obj}_t ||)}
        \end{align*}
        This encourages the agent to smoothly release the object and ensuring it remains in a similar position/rotation as the reference.
    \end{itemize}
\end{itemize}

Error on the object global translation: $r^{ht} = e^{-100 \cdot || \hat{p}_t - p_t ||}$.

Error on the object global rotation: $r^{hr} = e^{-1 \cdot < \hat{\theta}_t , \theta_t >}$ where $< \cdot , \cdot >$ is the quaternion difference.

We found that the reward and humanoid design were crucial to preventing penetrations, especially with small objects. A full hand grasp applies contact forces on all sides. These forces can be very high when the humanoid approaches with full body dynamics. This tends to result in penetration issues. These penetrations are not just visually unpleasing but also pose a problem in reconstructing the motions. For example, we observed issues such as a finger getting stuck inside an object. The object then follows the hand and the agent is unable to let go of it.

\begin{figure}
    \centering
    \includegraphics[width=0.5\linewidth]{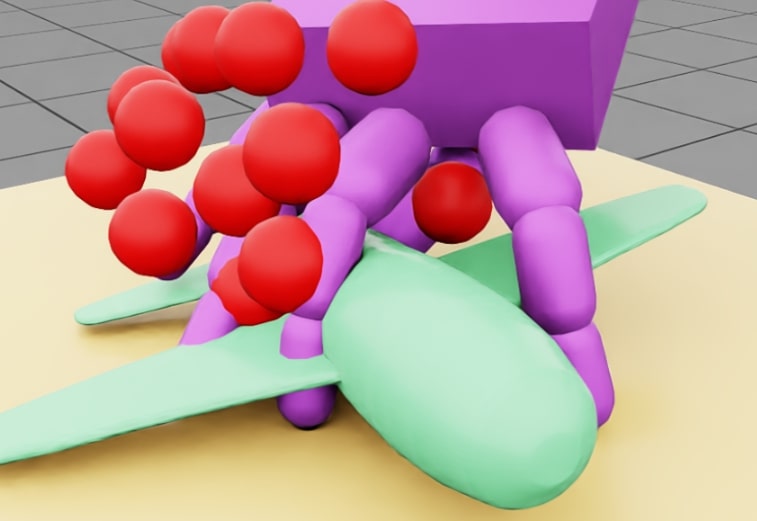}
    \caption{Penetation issues when improper reward and physics is used.}
    \label{fig: penetration}
\end{figure}

Reducing finger forces and ensuring the approach is smooth and contact forces remain within reasonable ranges -- these design choices help prevent penetrations such as shown in \cref{fig: penetration}.

% \bibliographystyle{ACM-Reference-Format}
% \bibliography{sample-bibliography}

\end{document}